\documentclass[twoside,11pt]{article}

\usepackage[abbrvbib, preprint]{jmlr2e} 

\usepackage{lmodern}
\usepackage[utf8]{inputenc}
\usepackage{mathtools}
\usepackage{microtype}
\usepackage{ifthen}
\usepackage{amssymb, amsmath}
\usepackage{booktabs}
\usepackage{multirow}

\usepackage{caption}
\usepackage{accents}
\usepackage{enumitem}

\usepackage[subrefformat=parens,labelformat=parens]{subcaption}
\usepackage[bibliography=common,appendix=append]{apxproof} 
\usepackage{tikz}
\usetikzlibrary{positioning, arrows.meta, shapes.geometric}
\tikzset{%
  dot/.style n args = {4}{name=#3, circle, draw, inner sep=1pt, minimum size=5pt, fill=black, label={[shift={(#1,#2)}]#4:$#3$}},
  lat/.style n args = {4}{name=#3, circle, draw, inner sep=1pt, minimum size=5pt, label={[shift={(#1,#2)}]#4:$#3$}},
  sb/.style n args = {4}{name=#3, circle, draw, inner sep=1pt, minimum size=7pt, label={[shift={(#1,#2)}]#4:$#3$}},
  dot5/.style n args = {5}{name=#3, circle, draw, inner sep=1pt, minimum size=5pt, fill=black, label={[shift={(#1,#2)}]#4:$#5$}},
  lat5/.style n args = {5}{name=#3, circle, draw, inner sep=1pt, minimum size=5pt, label={[shift={(#1,#2)}]#4:$#5$}},
  sq/.style n args = {4}{name=#3, rectangle, draw, inner sep=1pt, minimum size=5pt, fill=black, label={[shift={(#1,#2)}]#4:$#3$}},
  tr/.style n args = {4}{name=#3, regular polygon,regular polygon sides=4, draw, inner sep=1pt, minimum size=6pt, fill=gray, label={[shift={(#1,#2)}]#4:$#3$}},
  bordered/.style = {draw,outer sep=1, inner sep=2, minimum size=5pt},
  >={Latex[width=1.5mm,length=2mm]},
  every picture/.style={semithick}
}

\usepackage{algorithm}
\usepackage[indLines=false]{algpseudocodex}
\usepackage{ifthen}

\newcommand{\checkid}{\textsc{VerifyInputs}}
\newcommand{\simulate}{\textsc{SimulateInstance}}

\usepackage[table]{xcolor}
\definecolor{lightgray}{gray}{0.9}

\newcommand{\+}[1]{\ensuremath{\mathbf{#1}}}

\newcommand\independent{\protect\mathpalette{\protect\independenT}{\perp}} 
\def\independenT#1#2{\mathrel{\rlap{$#1#2$}\mkern2mu{#1#2}}}

\newcommand{\doo}{\mathrm{do}}
\newcommand{\cond}{\,\vert\,}
\newcommand{\Em}{\mathrm{em}}
\newcommand{\Pa}{\mathrm{pa}}
\newcommand{\PaV}{\mathrm{pa}}
\newcommand{\PaVU}{\mathrm{pa}^+}
\newcommand{\Ch}{\mathrm{ch}}

\newcommand{\rec}[1][]{%
  \ifthenelse{ \equal{#1}{} }
    {\mathrm{re}}
    {\mathrm{re}_{#1}}
}
\newcommand{\emi}[1][]{%
  \ifthenelse{ \equal{#1}{} }
    {\mathrm{em}}
    {\mathrm{em}_{#1}}
}

\newcommand{\An}[1][]{%
  \ifthenelse{ \equal{#1}{} }
    {\mathrm{an}}
    {\mathrm{an}_{#1}}
}
\newcommand{\De}[1][]{%
  \ifthenelse{ \equal{#1}{} }
    {\mathrm{de}}
    {\mathrm{de}_{#1}}
}

\newtheorem{defin}{Definition}
\newtheoremrep{lem}[defin]{Lemma}
\newtheoremrep{thm}[defin]{Theorem}

\newcommand{\sJ}{\mathcal{J}}
\newcommand{\sU}{\mathcal{U}}

\newcommand{\sM}{\mathcal{M}}
\newcommand{\sG}{\mathcal{G}}

\newcommand{\sI}{\mathcal{I}}

\newcommand{\sP}{\mathcal{P}}

\newcommand{\Ud}{\overline{\+U}}
\newcommand{\ud}{\overline{U}}
\newcommand{\Us}{\widetilde{\+U}}
\newcommand{\smallud}{\overline{\+u}}
\newcommand{\smallus}{\widetilde{\+u}}

\usepackage{lastpage}
\jmlrheading{27}{2026}{1-\pageref{LastPage}}{5/25; Revised
6/26}{8/26}{25-1140}{Otto Tabell, Santtu Tikka and Juha Karvanen}

\ShortHeadings{Clustering and Pruning in Causal Data Fusion}{Tabell, Tikka and Karvanen}

\begin{document}

\title{Clustering and Pruning in Causal Data Fusion}

\author{\name Otto Tabell \email otto.h.e.tabell@jyu.fi \\
        \name Santtu Tikka \email santtu.tikka@jyu.fi \\
        \name Juha Karvanen \email juha.t.karvanen@jyu.fi \\
        \addr Department of Mathematics and Statistics \\
        FI-40014 University of Jyvaskyla, Finland}

\editor{Jin Tian}

\maketitle

\begin{abstract}%
Data fusion---the process of combining observational and experimental data---can enable the identification of causal effects that would otherwise remain non-identifiable. Although identification algorithms have been developed for specific scenarios, do-calculus remains the only general-purpose tool for causal data fusion, particularly when variables are present in some data sources but not others. However, approaches based on do-calculus may encounter computational challenges as the number of variables increases and the causal graph grows in complexity. Consequently, there exists a need to reduce the size of such models while preserving the essential features. For this purpose, we propose pruning (removing unnecessary variables) and clustering (combining variables) as preprocessing operations for causal data fusion. We generalize earlier results on a single data source and derive conditions for applying pruning and clustering in the case of multiple data sources. We give sufficient conditions for inferring the identifiability or non-identifiability of a causal effect in a larger graph based on a smaller graph and show how to obtain the corresponding identifying functional for identifiable causal effects. Examples from epidemiology and social science demonstrate the use of the results.
\end{abstract}


\section{Introduction} \label{sec:intro}

The modern data landscape integrates data from multiple sources such as experiments, surveys, registries, and operational systems. In causal data fusion, the goal is to combine these data sources to identify a causal effect that cannot be identified from a single input distribution \citep{datafusion, shi2023data, colnet2024causal}. We call this the general causal effect identification problem. The potential application areas include, for instance, medicine \citep{dahabreh2023efficient,li2020causal}, ecology \citep{spake2022improving} and econometrics \citep{hunermund2023causal}.

The theoretical understanding of the general causal effect identification problem is still partial. While do-calculus \citep{pearl1995} is complete for causal effect identification with a single observational data source \citep{shpitser2006,huang2006}, a similar result has not been derived in the context of multiple data sources of unrestricted form. 

Algorithms for special cases of causal data fusion have been reviewed by \citet[Table 1]{tikka2021dosearch}. The more recent developments include the corrected g-identifiability algorithm with a positivity assumption \citep{kivva2022revisiting} and an algorithm for data fusion under systematic selection \citep{lee2024systematic}. All of these algorithms have important restrictions regarding their input distributions. For instance, g-identifiability \citep{lee2019gid,kivva2022revisiting} and g-transportability \citep{lee2020generalized} are constrained by the assumption that entire distributions are available, meaning that the union of observed and intervened variables must always be the set of all endogenous variables in the model. As an example, this means that besides direct application of do-calculus, there does not currently exist an identification algorithm that could determine the identifiability of $p(y \cond \doo(x))$ in the graph $X \rightarrow Z \rightarrow Y$ from the input distributions $p(x,z)$ and $p(z,y)$.

As a real-data example of these kinds of partially overlapping input distributions, \citet{karvanen2020epi} studied a causal data fusion problem where one data source contains information on salt-adding behavior and salt intake, while another data source contains information on salt intake and blood pressure. The goal was to estimate the causal effect of salt-adding behavior on blood pressure. Similar examples on partially overlapping data sources can be found in the Maelstrom Catalogue \citep{maelstrom_catalogue, maelstrom} that provides metadata on epidemiological studies. For instance, the catalogue lists 25 studies where some information on chronic obstructive pulmonary disease (COPD) has been measured. Out of these studies, 20 have data on physical measurements, 22 have data from biosamples, 11 have data on cognitive measures, and 10 have data from administrative databases; only two studies have data from all four categories. 

For more general cases with no constraints on the input distributions, Do-search \citep{tikka2021dosearch} has been suggested. Do-search is an algorithm that applies do-calculus and standard probability manipulations to determine the identifiability of a causal effect with given input distributions without constraints. However, as a consequence of the search-based nature of the algorithm, it becomes computationally expensive as the number of variables increases. The poor scalability of search-based identification algorithms also motivates the preprocessing efforts to reduce the size of the causal graph and thus decrease the number of variables. 

Importantly, if a causal graph is modified in any way, there is no guarantee that inferences made in the modified graph are transferable back to the original graph. Ideally, a method to reduce the size of causal graphs should not affect the identifiability of the causal effect of interest but such methods are scarce. We focus on two approaches for graph-size reduction: clustering and pruning.

In clustering, the aim is to represent a larger set of, typically related, vertices in a graph as a smaller set of vertices \citep{schaeffer2007clustering, malliaros2013clustering}. There are several examples of finding clusters and implementing clustering in causal graphs \citep{anand2023causal, zeng2025causal, niu2022learning}. However, due to the coarse graphical representations of clusters, these approaches may lead to a loss of identifiability, i.e., a causal effect may be identifiable in the clustered graph but not in the original graph. To remedy this issue, \citet{tikka2023clustering} introduced transit clusters which were proven to preserve the identifiability under certain conditions.

In addition, \citet{tikka2018pruning} introduced pruning, which refers to the process of completely removing vertices that are irrelevant for identification. Such vertices can, for instance, include variables that are descendants of the response variable or vertices that are connected to the causal graph through only one edge. Illustrations of pruning and clustering are presented in Figure~\ref{fig:illustrations}. The existing identifiability results regarding pruning and clustering have been presented in the context of a single observational data source, and are thus not directly applicable to causal data fusion.

\begin{figure}[!ht]
  \begin{center}
    \begin{subfigure}{0.36\textwidth}
      \centering
      \begin{tikzpicture}[scale=1.5]
        \node [dot5 = {-0.05}{0}{Z_5}{above}{Z_5}] at (-1,1) {};
        \node [dot5 = {-0.05}{0}{Z_6}{above}{Z_6}] at (2,1) {};
        \node [dot5 = {-0.05}{0}{Z_4}{above}{Z_4}] at (0,1) {};
        \node [dot5 = {0}{0}{Z_3}{below}{Z_3}] at (-1,0) {};
        \node [dot5 = {0}{0}{Z_2}{below}{Z_2}] at (0,0.5) {};
        \node [dot5 = {0.05}{0}{Z_1}{below}{Z_1}] at (0,-0.5) {};
        \node [dot5 = {0}{0}{X}{below}{X}] at (1, 0) {};
        \node [dot5 = {0}{0}{Y}{below}{Y}] at (2,0) {};
        \draw [->] (Z_4) -- (Z_3);
        \draw [->] (Z_5) -- (Z_3);
        \draw [->] (Z_4) to (Y); 
        \draw [->] (Z_3) -- (Z_1);
        \draw [->] (Z_3) -- (Z_2);
        \draw [->] (Z_1) -- (X);
        \draw [->] (Z_2) -- (X);
        \draw [->] (X) -- (Y);
        \draw [->] (Y) -- (Z_6);
      \end{tikzpicture}
      \caption{}
      \label{fig:size1}
    \end{subfigure}
    \begin{subfigure}{0.36\textwidth}
      \centering
      \begin{tikzpicture}[scale=1.5]
        \node [dot5 = {-0.05}{0}{Z_4}{above}{Z_4}] at (0,1) {};
        \node [dot5 = {0}{0}{Z_3}{below}{Z_3}] at (-1,0) {};
        \node [dot5 = {0}{0}{Z_2}{below}{Z_2}] at (0,0.5) {};
        \node [dot5 = {0.05}{0}{Z_1}{below}{Z_1}] at (0,-0.5) {};
        \node [dot5 = {0}{0}{X}{below}{X}] at (1, 0) {};
        \node [dot5 = {0}{0}{Y}{below}{Y}] at (2,0) {};
        \draw [->] (Z_4) -- (Z_3);
        \draw [->] (Z_4) to (Y); 
        \draw [->] (Z_3) -- (Z_1);
        \draw [->] (Z_3) -- (Z_2);
        \draw [->] (Z_1) -- (X);
        \draw [->] (Z_2) -- (X);
        \draw [->] (X) -- (Y);
      \end{tikzpicture}
      \caption{}
      \label{fig:size3}
    \end{subfigure}
    \begin{subfigure}{0.26\textwidth}
      \centering
      \begin{tikzpicture}[scale=1.5]
        \node [dot5 = {-0.05}{0}{Z_4}{above}{Z_4}] at (0.5,1) {};
        \node [dot5 = {0}{0}{T}{below}{T}] at (0,0) {};
        \node [dot5 = {0}{0}{X}{below}{X}] at (1, 0) {};
        \node [dot5 = {0}{0}{Y}{below}{Y}] at (2,0) {};
        \node [dot5 = {0}{0}{Y}{below}{Y}] at (2,0) {};
        \node (C) at (-0,-0.78) {\vphantom{C}};
        \draw [->] (Z_4) to (Y); 
        \draw [->] (Z_4) to (T); 
        \draw [->] (T) to (X); 
        \draw [->] (X) -- (Y);
      \end{tikzpicture}
      \caption{}
      \label{fig:size4}
    \end{subfigure}
  \end{center}
  \caption{Illustrations of size-reduction techniques for causal graphs: \subref{fig:size1} the original graph, \subref{fig:size3} a pruned graph where $Z_5$ and $Z_6$ have been deemed irrelevant for the identification of $p(y \cond \doo(x))$ and thus removed, \subref{fig:size4} a clustered graph obtained from the pruned graph where $T$ represents the set $\{Z_1,Z_2,Z_3\}$.}
  \label{fig:illustrations}
\end{figure}

In this paper, we derive sufficient conditions under which pruning and clustering can be applied within the general causal effect identification framework. We also show how identifying functionals obtained from a pruned or clustered graph can be used to derive identifying functionals for the original graph. Our results generalize the results by \citet{tikka2018pruning} and \citet{tikka2023clustering} to the case of multiple data sources. These generalizations are not straightforward because previous results on identification of causal effects regarding pruning and clustering rely on concepts that necessitate a clear distinction between observed and unobserved variables, such as c-components (districts), hedges, and the latent projection. These concepts are not well-defined when entire distributions are not available. 

By pruning and clustering, the computational load of algorithms used in the causal effect identification can be reduced, and the identifying functionals for the causal effects can be presented more concisely. Another perspective to clustering is top-down causal modeling where causal relations are first modeled between clusters of variables. This kind of approach is intuitive and commonly applied in practice \citep[and references therein]{tennant2021use}. Top-down causal modeling has also been applied with multiple data sources \citep{valkonen2024price,Youngmann2023Causal,karvanen2020epi}. The results derived in this paper can be used to determine when non-identifiability in a clustered graph implies non-identifiability in a graph without clusters, and when more detailed modeling of the internal structure of the clusters could be beneficial for the identification.

The paper is organized as follows. In Section~\ref{sec:causalmodel}, we introduce the notations and core definitions such as the structural causal model and the identifiability of causal effects. In Sections~\ref{sec:pruning} and \ref{sec:clustering}, we present our main results on causal effect identifiability with multiple data sources under pruning and clustering, respectively. In Section~\ref{sec:functionals}, we show that identifying functionals obtained from pruned or clustered causal graphs are applicable in the original causal graph. In Section~\ref{sec:simulation}, we perform a simulation study to assess the impact of pruning and clustering on running time performance of Do-search. In Section~\ref{sec:applications}, we illustrate our results via examples that originate from epidemiology and social science. Finally, Section~\ref{sec:conclusion} offers some concluding remarks. All identifying functionals of the paper were obtained using R Statistical Software version~4.5.1 \citep{r} and the \texttt{dosearch} package \citep{tikka2021dosearch, dosearch_package}. The R codes for the simulation study and the examples of the paper are available at \url{https://github.com/ottotabell/clustering-pruning}.

\section{Notations and Definitions}\label{sec:causalmodel}

We use upper case letters to denote random variables and vertices, and lower case letters to denote values. The shortcut notation $p(x)$ is used for the probability $P(X=x)$. We use bold letters to denote sets. We assume the reader is familiar with basic concepts of graph theory such as directed acyclic graphs (DAGs). We make no distinction between random variables and vertices and use the same symbols to refer to both. A DAG over a set of vertices $\+ V$ is denoted by $\sG(\+ V)$. We omit braces for sets consisting of only one element and write e.g., $\+V \setminus T$ as a shortcut notation for $\+V \setminus \{T\}$.

Let $\sG(\+ V)$ be a DAG and let $\+ W \subset \+ V$. The induced subgraph $\sG[\+ W]$ is a DAG obtained from $\sG$ by keeping all vertices in $\+ W$ and all edges between members of $\+ W$ in $\sG$. We use the notation $\sG[\overline{\+W}, \underline{\+Z}]$ to denote the DAG obtained from $\sG$ by removing all incoming edges of $\+W$ and outgoing edges of $\+Z$.

Following \citep{causality}, we define structural causal models as follows:

\begin{defin}[Structural causal model] \label{def:scm} 
A \emph{structural causal model} (SCM) $\sM$ is a tuple $(\+V, \+U, \+F, p)$, where 
\begin{itemize}
  \item $\+V$ and $\+U$ are disjoint sets of random variables. 
  \item $\+F$ is a set of functions such that each $f_i \in \+ F$ is a mapping from $\+U[V_i] \cup \PaV(V_i)$ to $V_i$, where $\+ U[V_i] \subset \+U$ and $\PaV(V_i) \subset \+V \setminus V_i$, such that $\+F$ forms a mapping from $\+U$ to $\+V$. In other words, each $V_i$ is defined by the structural equation $V_i=f_i(\PaV(V_i), \+U[V_i])$.
  \item $p$ is the joint probability distribution of $\+U$.
\end{itemize}
\end{defin}

We only consider recursive SCMs, i.e., models where the unique solution of $\+ F$ in terms of $\+ U$ is obtained by recursively substituting each $V_j \in \PaV(V_i)$ in each function $f_i$ with their corresponding function $f_j$. It is common to characterize members of $\+ V$ as ``observed'' or ``endogenous'', but because we consider scenarios with multiple data sources where the set of observations is not necessarily the same between the inputs, we will not use these terms. Similarly, members of $\+ U$ are commonly referred to as ``unobserved'', ``latent'' or ``exogenous'', which we will generally avoid. In other words, the only distinction between $\+ V$ and $\+ U$ is that the former is defined by the structural equations of $\+ F$ and the latter by the joint distribution $p(\+ u)$. Finally, we extend the notation $\+ U[V_i]$ to subsets $\+ W \subset \+ V$ disjunctively as $\+ U[\+ W] = \bigcup_{W_i \in \+ W} \+ U[W_i]$.

The DAG $\sG(\+ V \cup \+ U)$ induced by an SCM $\sM$ is called a \emph{causal graph} and it has a vertex for each member of $\+V \cup \+ U$, and there is a directed edge from each member of $\PaV(V_i) \cup \+ U[V_i]$ to $V_j$ for each $V_j$. A member of $\+U$ is called a \emph{dedicated error term} if it is present in exactly one structural equation, i.e., it has exactly one child in the corresponding causal graph \citep{karvanen2024counterfactuals}. The set of dedicated error terms is denoted as $\Ud$, and complementarily $\Us = \+U \setminus \Ud$ denotes members of $\+ U$ with more than one child. The dedicated error term for a variable $V_i\in\+V$ is denoted by $\ud[V_i]$. In all figures, dedicated error terms are omitted for clarity. Moreover, members of $\Us$ that have exactly two children are represented by dashed bi-directed edges in the figures. Starting from Section~\ref{sec:clustering} where we introduce clustering, we will also consider causal graphs where the dedicated error terms are explicitly omitted, i.e., DAGs of the form $\sG(\+ V \cup \Us)$, because it is natural to assume that the dedicated error terms are always clustered together with their children.

The sets of parents and ancestors of a vertex $V_i\in \+V \cup \+ U$ in a causal graph $\sG(\+V \cup \+ U)$ are denoted by $\Pa_\sG^+(V_i)$ and $\An_\sG^+(V_i)$, respectively. When we wish to include only those members of the aforementioned sets that are members of $\+ V$, we use notations $\Pa_\sG(V_i)$ and $\An_\sG(V_i)$, respectively. The sets of children and descendants are denoted by $\Ch_\sG(V_i)$ and $\De_\sG(V_i)$, respectively. All of these notations are also defined for sets of vertices by taking the union over the set elements, e.g., $\An_\sG(\+ W) = \bigcup_{W_i \in \+ W} \An_\sG(W_i)$.

The d-separation criterion can be used to infer conditional independence constraints of a joint distribution that is Markov-relative to a DAG $\sG$ \citep{pearl1995, causality}, such as the joint distribution induced by a recursive SCM.
\begin{defin}[d-separation]
Let $\sG(\+ V)$ be a DAG and let $X, Y \in \+ V$. A path $\sP$ between $X$ and $Y$ in $\sG$ is \emph{d-separated} by $\+ Z \subset (\+ V \setminus \{X, Y\})$ in $\sG$ if
\begin{itemize}
  \item $\sP$ contains a chain $A \rightarrow M \rightarrow B$ or a fork $A \leftarrow M \rightarrow B$ such that $M \in \+ Z$, or
  \item $\sP$ contains a collider $A \rightarrow M \leftarrow B$ such that $(M \cup \De_\sG(M)) \cap \+ Z = \emptyset$.
\end{itemize}
In addition, $X$ and $Y$ are d-separated by $\+ Z$ if $\+ Z$ d-separates every path between $X$ and $Y$. If $X$ and $Y$ are not d-separated by $\+ Z$, they are \emph{d-connected}. For sets, $\+ X$ and $\+ Y$ are d-separated by $\+ Z$ if $\+ Z$ d-separates every path between the members of $\+ X$ and $\+ Y$. 
\end{defin}
\noindent We denote the d-separation of $\+ X$ and $\+ Y$ given $\+ Z$ in a DAG $\sG$ as $(\+ X \independent \+ Y \cond \+ Z)_\sG$ and the implied conditional independence as $\+ X \independent \+ Y \cond \+ Z$.

Interventions by the do-operator \citep{causality} induce new causal models, called submodels.
\begin{defin}[Submodel] \label{def:submodel}
A \emph{submodel} $\sM_{\+x}$ of an SCM $\sM$ obtained from an intervention $\doo(\+ x)$ is an SCM $(\+ V, \+ U, \+ F', p)$ where $\+ F'$ is obtained from $\+ F$ by replacing $f_X \in \+ F$ for each $X \in \+ X$ with a constant function that outputs the corresponding value in $\+ x$.
\end{defin}
\noindent An \emph{interventional distribution} $p(\+y \cond \doo(\+ x))$ is the marginal distribution of $\+ Y$ in $\sM_{\+ x}$, which we also refer to as the \emph{causal effect} of $\+ X$ on $\+ Y$. 

In the general causal effect identification problem, the available information can be represented symbolically as a set of \emph{input distributions} $\sI = \{p(\+a_i \cond \doo(\+b_i),\+c_i)\}_{i = 1}^n$ where the sets $\+ A_i, \+ B_i,$ and $\+ C_i$ are disjoint and $\+ A_i \cup \+ B_i \cup \+ C_i \subset \+ V$ for each $i = 1,\ldots,n$, where $n$ is the number of input distributions. In this notation, $\+A_i$ represents the measured variables, $\+B_i$ represents the intervened variables, and $\+C_i$ represents the conditioning variables. For input distributions, the notation $p(\+a_i \cond \doo(\+b_i),\+c_i)$ implies that the probabilities are known for all possible values of $\+A_i$, $\+B_i$ and $\+C_i$.

For settings that involve multiple data sources, we define the identifiability of a causal effect as follows: 

\begin{defin}[Identifiability] \label{def:identifiability}
Let $\sG(\+V \cup \+U)$ be a causal graph and let $\+Y,\+X \subset \+V$ be disjoint. The causal effect $p(\+y \cond \doo(\+x))$ is \emph{identifiable} from $(\sG, \sI)$ if it is uniquely computable from $\sI$ in every SCM that induces $\sG$ and fulfills the positivity assumption $p(\+ v) > 0$.
\end{defin}
The positivity assumption $p(\+ v) > 0$ ensures that all input distributions and causal effects are always well-defined. The typical way to apply Definition~\ref{def:identifiability} is to demonstrate non-identifiability by constructing two models that agree on $\sI$ but disagree on $p(\+y \cond \doo(\+x))$. In many proofs of this paper, we construct two models, $\sM_1$ and $\sM_2$ for a graph $\sG$ based on models $\sM'_1$ and $\sM'_2$ that are known to exist in an altered graph $\sG'$ due to the non-identifiability of the causal effect in this altered graph or vice versa. 

Manual identification of causal effects relies on \emph{do-calculus} \citep{pearl1995}, which consists of three rules for manipulating interventional distributions:
\begin{enumerate}
\item{Insertion and deletion of observations: 
\[
  p(\+ y \cond \doo(\+ x), \+ z, \+ w) = p(\+ y \cond \doo(\+ x), \+ w), \text{ if } (\+ Y \independent \+Z \cond \+X, \+ W)_{\sG[\overline{\+ X}]}.
\]
}
\item{Exchanging actions and observations:
\[
  p(\+ y \cond \doo(\+x,\+ z), \+ w) = p(\+ y \cond \doo(\+ x), \+ z, \+ w), \text{ if } (\+ Y \independent \+Z \cond \+X, \+ W)_{\sG[\overline{\+ X},\underline{\+ Z}]}.
\]
}
\item{Insertion and deletion of actions:
\[
  p(\+ y \cond \doo(\+ x, \+ z), \+ w) = p(\+ y \cond \doo(\+ x), \+ w), \text{ if } (\+ Y \independent \+Z \cond \+X, \+ W)_{\sG[\overline{\+ X},\overline{Z(\+ W)}]},
\]
where $Z(\+ W) = \+Z \setminus \An(\+ W)_{\sG[\overline{\+ X}]}.$}
\end{enumerate}
When we consider the identifiability of causal effects outside the restricted settings of known identifiability algorithms, do-calculus remains the only tool available for demonstrating that a causal effect is identifiable. The positivity assumption of Definition~\ref{def:identifiability} also ensures the validity of the rules of do-calculus.

Clustering and pruning are operations that transform the causal graph and the input distribution in a specific way. We are interested in finding conditions under which these operations preserve identifiability and non-identifiability. We define this property formally as follows:
\begin{defin}[Identification invariance] \label{def:invariance} 
Let $\sG$ be a causal graph and let $\sI$ be a set of input distributions. An operation $\Omega$ is \emph{identification invariant} for a causal effect $p(\+y \cond \doo(\+x))$ if $p(\+y \cond \doo(\+x))$ is identifiable from $(\sG, \sI)$ if and only if $p'(\+y \cond \doo(\+x))$ is identifiable from $(\sG', \sI')$ where $(\sG', \sI') = \Omega(\sG, \sI)$.
\end{defin}
In the definition, $p$ refers to any distribution that is Markov-relative to $\sG$, and $p'$ refers to any distribution that is Markov-relative to  $\sG'$. We note that identification invariance establishes an equivalence class of causal graphs and input distributions in the sense that if we are able to determine whether a causal effect is identifiable from $(\sG', \sI')$, then we will also know whether it is identifiable from any $(\sG, \sI)$ such that $(\sG', \sI') = \Omega(\sG, \sI)$. In this paper, we focus on pruning and clustering which are examples of operations that produce coarser graphical representations of the system under study. We note that there are also operations that produce more refined graphical representations, the peripheral extension \citep{tikka2023clustering} being an example.

\section{Preprocessing by Pruning}\label{sec:pruning}

\begin{toappendix}
  \label{apx:pruningproof}
\end{toappendix}

It is sometimes possible to simplify the causal model by completely removing variables that are unnecessary for the identification of a causal effect of interest. This operation is called \emph{pruning} by \citet{tikka2018pruning} who studied pruning in the case of a single observational data source. Ideally, the pruning operation should be identification invariant.

We present three results regarding pruning in the case of multiple data sources. The first theorem states that non-ancestors of $\+Y$ can be removed without affecting the identifiability of $p(\+ y \cond \doo(\+ x))$, as long as none of the input distributions are conditioned or intervened by non-ancestors of $\+ Y$. The second theorem states conditions for removing variables that are d-separated from $\+ Y$ after an intervention on $\+ X$. The third theorem gives similar conditions for removing variables that are connected to the rest of the graph only through a single variable. The two latter theorems assume that non-ancestors of $\+ Y$ have already been pruned. Our results generalize those by \citet{tikka2018pruning} to multiple data sources. The proofs of this section are presented in Appendix~\ref{apx:pruningproof}.

First we define how pruning changes the input distributions. If the pruning reduces the set of variables from $\+V$ to $\+ W \subset \+ V$, the input distributions are restricted in the following way:
\begin{defin}[Restricted input distributions] \label{def:restrictedinput}
  Let $\sI = \{p(\+a_i \cond \doo(\+b_i),\+c_i)\}_{i = 1}^n$ be a set of input distributions and let $\+ W \subset \+ V$. The set of \emph{input distributions restricted to} $\+ W$ is
  \[
    \sI[\+ W] = \{ p(\+a_i \cap \+ w \cond \doo(\+b_i),\+c_i) : p(\+a_i \cond \doo(\+b_i),\+c_i) \in \sI, \,\+ A_i \cap \+ W \neq \emptyset, \,\+ B_i \cup \+ C_i \subseteq \+ W \}.
  \]
\end{defin}
\noindent Definition~\ref{def:restrictedinput} ensures that the restricted input distributions are well-defined and excludes inputs that are not obtainable from the original inputs $\sI$ directly through marginalization.

Definition~\ref{def:restrictedinput} may appear too limited at first glance because $\sI$ does not necessarily uniquely represent the available information. For example, if $\sI_1 = \{p(x), p(y \cond x)\}$, $\sI_2 = \{ p(x \cond y), p(y) \}$, and $\sI_3 = \{ p(x,y) \}$, then each of these three sets of input distributions represents the same information (the joint distribution $p(x,y)$) but according to Definition~\ref{def:restrictedinput} we have $\sI_1[\{Y\}]=\emptyset$ while $\sI_2[\{Y\}] = \sI_3[\{Y\}]=\{p(y)\}$. However, there are good reasons why Definition~\ref{def:restrictedinput} is given as it is. Assume that we would consider all distributions that can be derived from the original input distributions via standard probability calculus. The number of these distributions increases quickly when the number of inputs and the number of variables increase. Even more distributions can be derived if we consider conditional independence relations implied by the graph and apply the rules of do-calculus. Another argument for the current definition is that we wish pruning to be a genuine preprocessing step meaning that there is no reason to use the variables that are removed in pruning. For example, if we have datasets 1 and 2 that provide information on $p(x)$ and $p(y \cond x)$, respectively, we do not have direct access to data from $p(y)$ but we should, for instance, resample the rows of dataset 2 using the probabilities $\hat{p}(x)$ estimated from dataset 1 as weights. To summarize, an analyst wishing to extend the possibilities for pruning has to explicitly derive new input distributions before the pruning is done. The formulation of Definition~\ref{def:restrictedinput} is a design choice that deliberately favors simplicity. Alternative views are also possible, such as considering the input distributions from a purely symbolic perspective and allowing pruning whenever a joint distribution is derivable from the inputs. Such an approach would extend the applicability of pruning in theory while ignoring aspects related to practical data analysis. 
 
\begin{thmrep}[Pruning non-ancestors] \label{thm:Yancestors}
Let $\sG(\+ V \cup \+ U)$ be a causal graph. If $\+B_i \cup \+C_i \subset \An[\sG](\+ Y)$ for all $i = 1,\ldots,n$, then pruning $(\sG, \sI) \rightarrow (\sG', \sI')$, where $\sI' = \sI[\An[\sG](\+ Y) \cup \+ Y]$, $\sG' = \sG[\An[\sG]^+(\+ Y) \cup \+ Y]$ and $\+ X' = \+ X \cap \An[\sG](\+ Y)$, is identification invariant for $p(\+ y\cond\doo(\+ x))$, and
$p(\+ y\cond\doo(\+ x)) = p'(\+ y\cond\doo(\+ x'))$.
\end{thmrep}
\begin{proof}
Let $\+ W = \+ V \setminus (\An[\sG](\+ Y) \cup \+ Y)$. By applying rule 3 of do-calculus twice and then noting that a causal model that induces $\sG'$ has the same joint distribution over $\+ W$ as the submodel obtained from the intervention $\doo(\+W = \+ w)$, we have that
\[
  p(\+y \cond \doo(\+ x)) = p(\+ y \cond \doo(\+ x'))
                          = p(\+ y \cond \doo(\+ x', \+ w))
                          = p'(\+ y \cond \doo(\+ x')).
\]
This shows that the causal effects are the same in $\sG$ and $\sG'$. It remains to show that the identifiability of the causal effect is not affected by the removal of the non-ancestors. 

Assume first that $p'(\+y\cond\doo(\+x))$ is identifiable from $\sI'$ in $\sG'$. Then, members of $\sI'$ are directly identifiable from $\sI$ through marginalization, because of the assumption $\+B_i \cup \+C_i \subset \An[\sG](\+ Y)$ and because no member of $\+ V \setminus (\An[\sG](\+ Y) \cup \+ Y)$ appears in the structural equations of members of $\An[\sG](\+ Y) \cup \+ Y$. Because of this, and because the causal effects are the same, $p(\+y\cond\doo(\+x))$ is identifiable from $\sI$ in $\sG$ with the same identifying functional as in $\sG'$.

Assume then that $p'(\+y\cond\doo(\+x))$ is not identifiable. It follows that there exist two models $\sM_{1}'$ and $\sM_{2}'$ for which the input distributions $\sI'_1$ and $\sI'_2$ agree but $p_{1}'(\+y \cond \doo(\+x)) \neq p_{2}'(\+y \cond \doo(\+x))$ for some value assignment $\+x$. The goal is to construct two models $\sM_{1}$ and $\sM_{2}$ for $\sG$ for which the input distributions $\sI$ agree but $p_{1}(\+y \cond \doo(\+x)) \neq p_{2}(\+y\cond \doo(\+x))$ for some value assignment $\+x$. We construct these models as follows. The functions for the members of $\An[\sG](\+Y) \cup \+ Y$ are copied from $\sM_{1}'$ to $\sM_{1}$ and analogously from $\sM_{2}'$ into $\sM_{2}$. This guarantees that the causal effects differ between $\sM_{1}$ and $\sM_{2}$. The functions for the members of $\+V \setminus (\An[\sG](\+Y) \cup \+ Y)$ are defined the same way for both $\sM_{1}$ and $\sM_{2}$ so that variables $\+V \setminus (\An[\sG](\+Y) \cup \+ Y)$ are Bernoulli distributed random variables independent of all other variables in $\+ V \cup \+ U$ but their dedicated error terms. This guarantees that $\sI_1 = \sI_2$, and together with the difference in the causal effects, it applies that $p(\+y \cond\doo(\+ x))$ is not identifiable.
\end{proof}

Theorem~\ref{thm:Yancestors} allows us to only consider graphs $\sG = \sG[\An_\sG^+(\+ Y) \cup \+ Y]$ where the non-ancestors of the response variables $\+Y$ have been removed. In the remainder of the paper, we assume that the graphs have already been pruned this way. 

We highlight the nuances of the assumption $\+B_i \cup \+C_i \subset \An_\sG(\+ Y)$ via two examples. First, consider the graph of Figure~\ref{fig:ancnec1} where $Z$ is not an ancestor of $Y$. The variable $Z$ is necessary for the identification of $p(y \cond \doo(x))$ if the input distributions are $p(z)$ and $p(x,y \cond z)$. On the other hand, one might be tempted to represent the input distributions simply as $p(x,y,z)$, which would satisfy the assumption and thus seems to permit the pruning of $Z$. This is, however, a misguided conclusion because forming the input distribution $p(x,y)$ requires the use of $Z$. In the graph of Figure~\ref{fig:ancnec2}, $p(y \cond \doo(x))$ can be identified from the input distribution $p(x,y \cond z)$ directly as $p(y \cond \doo(x)) = p(y \cond x,z)$ because $Z$ and $Y$ are independent when $X$ is intervened. Here, the input distribution $p(x,y \cond z)$ is necessary for the identification but the actual value of $Z$ in the identifying functional $p(y \cond x,z)$ is irrelevant. 

\begin{figure}[!ht]
  \begin{center}
    \begin{subfigure}{0.25\textwidth}
      \centering
      \begin{tikzpicture}[scale=2]
        \node [dot = {0}{0}{X}{above}] at (0,0) {};
        \node [dot = {0}{0}{Y}{above}] at (1,0) {};
        \node [dot = {0}{0}{Z}{below}] at (0.5,-0.67) {};
        \draw [->] (X) -- (Y);
        \draw [->] (X) -- (Z);
        \draw [->] (Y) -- (Z);
      \end{tikzpicture}
      \caption{}
      \label{fig:ancnec1}
    \end{subfigure}
       \begin{subfigure}{0.25\textwidth}
      \centering
      \begin{tikzpicture}[scale=2]
        \node [dot = {0}{0}{X}{above}] at (0,0) {};
        \node [dot = {0}{0}{Y}{above}] at (1,0) {};
        \node [dot = {0}{0}{Z}{below}] at (0.5,-0.67) {};
        \draw [->] (X) -- (Y);
        \draw [->] (X) -- (Z);
      \end{tikzpicture}
      \caption{}
      \label{fig:ancnec2}
    \end{subfigure}
  \end{center}
  \caption{Causal graphs demonstrating the necessity of the assumption $\+B_i \cup \+C_i \subset \An_\sG(\+ Y)$ for pruning non-ancestors.}
  \label{fig:ancnec}
\end{figure}

The second pruning theorem considers the removal of variables that are independent of the response after an intervention.

\begin{thmrep}[Pruning non-ancestors after intervention] \label{thm:Xancestors} Let $\sG(\+ V \cup \+ U)$ be a causal graph such that $\sG = \sG[\An_\sG^+(\+ Y) \cup \+ Y]$ and let 
$\+ Z = \{Z \in \+ V \setminus \+ X\, \colon Z \independent \+Y \cond \+X \text{ in }\sG[\overline{\+X}] \}$, 
$\+V'= \+ V \setminus \+Z$ and $\+ U' =\+U[\+ V']$. If all of the following conditions hold
\begin{enumerate}[label=(\alph*)]
\item $\+Z \cap \De_\sG(\+X) = \emptyset$,
\item $\+B_i \cup \+C_i \subset \+ V'$ for all $i = 1,\ldots,n$, and 
\item for all members of $\+U[\+Z]$ it holds that if $U \in \+U[\+Z]$ is an ancestor of all variables in $\+X_S \subseteq (\+X \cap \Ch_\sG(\+Z \cup \+U[\+Z]))$, then there exists $U_S \in \+U$ that is a parent of all members of $\+X_S$, 
\end{enumerate}
then pruning $(\sG, \sI) \rightarrow (\sG', \sI')$, where $\sI' = \sI[\+ V']$ and $\sG' = \sG[\+ V' \cup \+U']$,  is identification invariant for $p(\+ y\cond \doo(\+ x))$, and
$p(\+ y\cond\doo(\+ x)) = p'(\+ y\cond\doo(\+ x))$. 
\end{thmrep}
\begin{proof}
By assumption, $\sG$ only contains $\+Y$ and the ancestors of $\+ Y$. The definition of $\+ Z$ allows us to use rule 3 of do-calculus, and thus we have $p(\+ y\cond \doo(\+ x)) = p(\+ y \cond \doo(\+x, \+z))$. Applying the truncated factorization formula \citep{causality} we can write
\begin{align*}
p(\+ y \cond \doo(\+ x)) &= p(\+y\cond \doo(\+x,\+z)) \\
&= \sum_{\+ u} \sum_{\+v \setminus (\+y \cup \+x \cup \+z) } \prod_{\+v \setminus (\+x \cup \+z)} p(v_i \cond \Pa(v_i), \+ u[v_i]) \prod_{\+ u} p(u_i) \\
&= \sum_{\+u \setminus \+u'}\sum_{\+u'} \sum_{\+v \setminus (\+y \cup \+x \cup \+z) } \prod_{\+v \setminus (\+x \cup \+z)} p(v_i \cond \Pa(v_i),\+u[v_i]) \prod_{\+u'} p(u_i) \prod_{\+u \setminus \+u'} p(u_i).
\end{align*}
As no children of $\+U \setminus \+U'$ are present in the expression, we can sum $\+U \setminus \+U'$ out and obtain
\begin{align*}
p(\+y \cond \doo(\+x))  
&= \sum_{\+u'} \sum_{\+v \setminus (\+y \cup \+x \cup \+z)} \prod_{\+v \setminus (\+x \cup \+z)} p(v_i \cond \Pa(v_i), \+u[v_i]) \prod_{\+u'} p(u_i) \\
& = \sum_{\+u'} \sum_{\+v' \setminus (\+y \cup \+x)} \prod_{\+v' \setminus \+x} p'(v_i \cond \Pa(v_i),\+u[v_i]) \prod_{\+u'} p'(u_i) \\ 
&= p'(\+y \cond \doo(\+x)). 
\end{align*}
In other words, pruning does not change the causal effect $p(\+y \cond \doo(\+ x))$. It remains to show that the identifiability of the causal effect is preserved.

Assume first that causal effect $p(\+y \cond \doo(\+x))$ is not identifiable from $\sI$ in $\sG$. It follows that there exist two models $\sM_1$ and $\sM_2$ for which $p_1(\+y\cond \doo(\+x)) \neq p_2(\+y\cond \doo(\+x))$ but the input distributions $\sI_1$  and $\sI_2$ are the same. We may divide $\+U[\+ Z]$ into partition $\sU$ in such a way that all members of $\+U[\+ Z]$ that have the same set of descendants in $\+X \cap \Ch_\sG(\+Z \cup \+U[\+Z])$ belong to the same class, i.e. $U_i$ and $U_j$ belong to the same class if $U_i,U_j \in \+U[\+Z]$ and $\De_\sG(U_i) \cap (\+X \cap \Ch_\sG(\+Z \cup \+U[\+Z])) = \De_\sG(U_j) \cap (\+X \cap \Ch_\sG(\+Z \cup \+U[\+Z]))$. The class whose children in the set $\+X \cap \Ch_\sG(\+Z \cup \+U[\+Z])$ are $\+X_S$ is denoted by $\sU[\+X_S]$. It follows from condition (c) that there exists a surjective mapping $g$ from $\sU$ to $\+U$ so that $U_S = g(\sU[\+X_S])$ is a parent of all members of $\+X_S$. This is fulfilled, for instance, if there exists $U \in \+U$ that is a parent of all members of $\+X \cap \Ch_\sG(\+Z \cup \+U[\+Z])$. 
Models $\sM'_1$ and $\sM'_2$ are obtained from $\sM_1$ and $\sM_2$, respectively, by the following process which is presented only for $\sM'_1$:
\begin{enumerate}
\item The structural equations for variables $\+V \setminus (\+X \cup \+Z)$ are copied from $\sM_1$ to $\sM'_1$. The assumption $\sG = \sG[\An_\sG^+(\+ Y) \cup \+ Y]$, the definition of $\+Z$, and condition (a) guarantee that these structural equations do not contain members of $\+Z$.
\item The variable $U_S$ is redefined to be a vector-valued variable that contains all members of $\+U[\+Z]$ that belong to any class $\sU[\+X_S]$ for which $U_S=g(\sU[\+X_S])$. The distribution of $U_S$ in $\sM'_1$ is defined to be equal to the joint distribution of the members of all classes $\sU[\+X_S]$ in $\sM_1$ for which $U_S=g(\sU[\+X_S])$. 
\item The structural equations for variables in $\+X$ are copied from $\sM_1$ into $\sM'_1$ and then redefined by the following procedure:
\begin{enumerate}[label*=\arabic*.]
\item If $X \in \+X$ is a function of a variable $Z \in \+Z$, variable $Z$ is replaced by its function $f_Z(\Pa(Z),\+U[Z])$. From condition (a) it follows that $\Pa(Z)$ cannot contain members of $\+X$. 
\item Step 3.1 is repeated until the structural equation does not contain any members of $\+Z$. 
\item If $X$ is a function of a variable $U_j$ and $U_j$ belongs to class $\sU[\+X_S]$, then $U_j$ is replaced by $U_S=g(\sU[\+X_S])$ and $X$ becomes a function of the component of $U_S$ that corresponds to $U_j$. 
\end{enumerate}
\end{enumerate}

As a result of these steps, the distribution of $\+X$ is equal between $\sM'_1$ and $\sM_1$, and between $\sM'_2$ and $\sM_2$. It follows from this and the first step of the procedure that the input distributions are the same between $\sM'_1$ and $\sM'_2$. Condition (b) guarantees that the input distributions $\sI'$ are well-defined. Since it was shown above that $p_1(\+y\cond \doo(\+x))=p'_1(\+y\cond \doo(\+x))$ and $p_2(\+y\cond \doo(\+x))=p'_2(\+y\cond \doo(\+x))$, the causal effects also differ between $\sM'_1$ and $\sM'_2$. 

Assume then that causal effect $p'(\+y\cond \doo(\+x))$ is not identifiable from $\sI'$ in $\sG'$. It follows that there exist models $\sM'_1$ and $\sM'_2$ for which $p'_1(\+y\cond \doo(\+x)) \neq p'_2(\+y\cond \doo(\+x))$ but the input distributions $\sI'_1$  and $\sI'_2$ are the same. We construct models $\sM_1$ and $\sM_2$ from $\sM'_1$ and $\sM'_2$, respectively, by defining variables $\+Z$ as Bernoulli variables independent of other variables except for their dedicated error terms. Now $p_1(\+y\cond \doo(\+x)) \neq p_2(\+y\cond \doo(\+x))$ but the input distributions $\sI_1$ and $\sI_2$ are the same.
\end{proof}

In Theorem~\ref{thm:Xancestors}, the set of pruned variables, $\+Z$, is defined to be d-separated from the response $\+Y$ when $\+X$ is intervened. The conditions (a), (b), and (c) are needed to ensure that the essential properties of the graph do not change when $\+Z$ is pruned. Condition (a) ensures that the members of $\+Z$ cannot be on a directed path between members of $\+X$ because then the removal of $\+Z$ could change the relationships between members of $\+X$. Condition (b) says that $\+Z$ cannot contain variables that are a part of intervention or conditioning in some input distribution. Condition (c) ensures that the d-separation properties between the members of $\+X$ do not change when $\+Z$ and $\+U[\+Z]$ are removed.  

Our final pruning result relates to a scenario where a set of vertices is connected to the rest of the causal graph via a single vertex.
\begin{thmrep}[Pruning isolated vertices] \label{thm:isolated}
  Let $\sG(\+ V \cup \+ U)$ be a causal graph such that $\sG = \sG[\An_\sG^+(\+ Y) \cup \+ Y]$ and let $W \in \+ V$. If there exists a set $\+ Z \subset \+ V$ such that $\+ Z \cap (\+ X \cup \+ Y) = \emptyset$, $\+ Z$ is connected to $\+ V \setminus \+ Z$ only through $W$, and $\+ B_i \cup \+ C_i \subset \+ V \setminus \+ Z$ for all $i = 1,\ldots,n$, 
then pruning $(\sG, \sI) \rightarrow (\sG', \sI')$, where $\sI' = \sI[\+ V']$, $\sG' = \sG[\+ V' \cup \+ U']$ and $\+ V' = \+ V \setminus \+ Z$ and $\+ U' = \+ U[\+ V']$, is identification invariant for $p(\+ y \cond \doo(\+ x))$, and
$p(\+ y\cond\doo(\+ x)) = p'(\+ y\cond\doo(\+ x))$. 
\end{thmrep}
\begin{proof}
Applying the truncated factorization formula, we obtain
\[
  p(\+ y \cond \doo(\+ x)) = \sum_{\+ u} \sum_{\+ v \setminus (\+ y \cup \+ x)} p(w \cond \Pa(w), \+ u[w]) \prod_{\+ v \setminus (\+ x \cup w)} p(v_i \cond \Pa(v_i), \+ u[v_i]) \prod_{\+ u} p(u_i).
\]
By definition, members of $\+ Z$ and $\+ U[\+ Z]$ are connected to $\+ V \setminus (\+ Z \cup W)$ only through $W$. By assumption, $\sG$ does not contain any non-ancestors of $\+Y$, which means that all members of $\+ Z$ must be ancestors of $W$. Thus, we can complete the marginalization over $\+ Z$ and $\+ U[\+ Z] \setminus \+ U[W]$ to obtain
\[
  \begin{aligned}
  &p(\+ y \cond \doo(\+ x)) \\
  &= \sum_{(\+ u \setminus \+ u[\+ z]) \cup \+ u[w]} \sum_{\+ v \setminus (\+ y \cup \+ x \cup \+ z)} p(w | \Pa(w) \setminus \+ z, \+ u[w]) \!\! \prod_{\+ v \setminus (\+ x \cup \+ z \cup w)} p(v_i | \Pa(v_i), \+ u[v_i]) \!\! \prod_{(\+ u \setminus \+ u[\+ z]) \cup \+ u[w]} p(u_i) \\
  &= \sum_{\+ u'} \sum_{\+ v' \setminus (\+ y \cup \+ x)} p'(w \cond \Pa'(w), \+ u'[w]) \prod_{\+ v' \setminus (\+ x \cup w)} p'(v_i \cond \Pa'(v_i), \+ u'[v_i]) \prod_{\+ u'} p'(u_i) \\
  &= \sum_{\+ u'} \sum_{\+ v' \setminus (\+ y \cup \+ x)} \prod_{\+ v' \setminus \+ x} p'(v_i \cond \Pa'(v_i), \+ u'[v_i]) \prod_{\+ u'} p'(u_i) \\
  &= p'(\+ y \cond \doo(\+ x))
  \end{aligned}
\]
Thus, the causal effect is unaltered by the pruning. It remains to show that the identifiability of the causal effect is preserved. 

Assume first that $p(\+ y \cond \doo(\+ x))$ is not identifiable from $\sI$ in $\sG$. It follows that there exist models $\sM_1$ and $\sM_2$ for which $p_1(\+y\cond \doo(\+x)) \neq p_2(\+y\cond \doo(\+x))$ but the input distributions $\sI_1$ and $\sI_2$ are the same. We construct $\sM'_1$ and $\sM'_2$ from $\sM_1$ and $\sM_2$, respectively. First, the structural equations for variables in $\+ V \setminus (\+ Z \cup W)$ are copied from $\sM_1$ into $\sM'_1$ and from $\sM_2$ into $\sM'_2$. Next, the structural equation for $W$ in $\sM'_1$ is defined by first copying the function of $W$ in $\sM_1$ and then replacing every instance of members of $\Pa_\sG(W) \cap \+ Z$ by its corresponding function recursively until the function of $W$ is expressed in terms of $\+ U$ only. The same process is carried out for $\sM'_2$. This ensures that $\sI'_1 = \sI'_2$, but $p'_1(\+y\cond \doo(\+x)) \neq p'_2(\+y\cond \doo(\+x))$.

Assume then that $p'(\+y\cond \doo(\+x))$ is not identifiable from $\sI'$ in $\sG'$. It follows that there exist models $\sM'_1$ and $\sM'_2$ for which $p'_1(\+y\cond \doo(\+x)) \neq p'_2(\+y\cond \doo(\+x))$ but the input distributions $\sI'_1$ and $\sI'_2$ are the same. We construct models $\sM_1$ and $\sM_2$ from $\sM'_1$ and $\sM'_2$, respectively, by defining variables $\+ Z$ as Bernoulli variables independent of other variables except for their dedicated error terms in both models. The functions for $W$ are constructed so that the value of $W$ does not depend on the parents of $W$ in $\+ Z$. Now $p_1(\+y\cond \doo(\+x)) \neq p_2(\+y\cond \doo(\+x))$ but the input distributions $\sI_1$ and $\sI_2$ are the same.
\end{proof}

In some instances, it is possible to identify a causal effect using only a subset of the input distributions. In such cases, the conditions of Theorems~\ref{thm:Yancestors}--\ref{thm:isolated} need to hold only for the members of this subset, thus expanding the applicability of these theorems. We will explain this feature in greater detail in Section~\ref{sec:functionals}.

To illustrate the results of this section, we consider the graph of Figure~\ref{fig:graph2g} with input distributions of $p(y, z_3, z_4, z_5 \cond \doo(w_1), w_2)$, $p(w_1, z_1, z_2, z_5, z_7 \cond \doo(x_1,x_2), w_2)$, and $p(w_2, z_6, z_7)$, and aim to identify $p(y \cond \doo(x_1,x_2))$. First, Theorem~\ref{thm:Yancestors} is applied to remove $Z_4$ and $Z_5$ because they are not ancestors of $Y$. We can see that $Z_1$, $Z_2$, and $Z_3$ are d-separated from $Y$ when the incoming edges to $X_1$ and $X_2$ are removed. As $\+Z=\{Z_1,Z_2,Z_3\}$ fulfills the conditions of Theorem~\ref{thm:Xancestors}, we can prune this set next. Note that the unobserved confounder between $X_1$ and $X_2$ is necessary for the application of Theorem~\ref{thm:Xancestors} because $X_1$ and $X_2$ have common ancestors in $\+U[\+Z]$. Finally, we apply Theorem~\ref{thm:isolated} to remove $Z_6$ and $Z_7$. Figure~\ref{fig:graph2gprime} presents the pruned version of the graph.

\begin{figure}[!ht]
  \begin{center}
    \begin{subfigure}{0.6\textwidth}
      \centering
      \begin{tikzpicture}[xscale=2.3,yscale=3]
        \node [dot = {0.15}{-0.1}{X_1}{below}] at (-1,0) {};
        \node [dot = {0}{0}{X_2}{below}] at (-1,-0.5) {};
        \node [dot = {0}{0}{Y}{below}] at (1,0) {};
        \node [dot = {0.2}{0}{W_1}{above}] at (0,0) {};
        \node [dot = {0}{0}{W_2}{below}] at (0,-0.5) {};
        \node [dot = {0}{0}{Z_7}{below}] at (1,-0.5) {};
        \node [dot = {0}{0}{Z_6}{below}] at (0.5, -0.5) {};
        \node [dot = {0}{0}{Z_5}{above}] at (-0.25,0.5) {};
        \node [dot = {0}{0}{Z_4}{above}] at (-1,0.5) {};
        \node [dot = {0}{0}{Z_3}{above}] at (-1.75,0.5 ) {};
        \node [dot = {0}{0}{Z_2}{below}] at (-2.25,0) {};
        \node [dot = {0}{0}{Z_1}{below}] at (-1.75,-0.5) {};
        \draw [->] (X_1) -- (W_1) ;
        \draw [->] (X_2) -- (W_1) ;
        \draw [->] (W_1) -- (Y) ;
        \draw [->] (W_2) -- (Y) ;
        \draw [->] (W_2) -- (X_1) ;
        \draw [->] (W_2) -- (W_1) ;
        \draw [->] (W_1) -- (Z_5) ;
        \draw [->] (X_1) -- (Z_4) ;
        \draw [->] (Z_7) -- (Z_6) ;
        \draw [->] (Z_6) -- (W_2) ;
        \draw [->] (Z_5) -- (Z_4) ;
        \draw [->] (Z_3) -- (Z_4) ;
        \draw [->] (Z_2) -- (Z_3) ;
        \draw [->] (Z_2) -- (Z_1) ;
        \draw [->] (Z_1) -- (X_1) ;
        \draw [->] (Z_1) -- (X_2) ;
        \draw [->] (Z_3) -- (X_1) ;
        \draw [<->, dashed] (X_1) to [bend left=35] (Y) ;
        \draw [<->, dashed] (X_1) to [bend left=30] (W_1) ;
        \draw [<->, dashed] (X_1) to [bend right=30] (X_2) ;
      \end{tikzpicture}
      \caption{}
      \label{fig:graph2g}
    \end{subfigure}
    \hfill
    \begin{subfigure}{0.35\textwidth}
      \begin{tikzpicture}[xscale=2.3,yscale=3]
        \node [dot = {0.1}{-0.1}{X_1}{below}] at (0,0) {};
         \node [dot = {0}{0}{X_2}{below}] at (0,-0.5) {};
        \node [dot = {0}{0}{Y}{below}] at (2,0) {};
        \node [dot = {0.1}{0}{W_1}{above}] at (1,0) {};
        \node [dot = {0}{0}{W_2}{below}] at (1,-0.5) {};
        \draw [->] (X_1) -- (W_1) ;
        \draw [->] (X_2) -- (W_1) ;
        \draw [->] (W_1) -- (Y) ;
        \draw [->] (W_2) -- (Y) ;
        \draw [->] (W_2) -- (X) ;
        \draw [->] (W_2) -- (W_1) ;
        \draw [<->, dashed] (X_1) to [bend left=35] (Y) ;
        \draw [<->, dashed] (X_1) to [bend left=30] (W_1) ;
        \draw [<->, dashed] (X_1) to [bend right=50] (X_2) ;
      \end{tikzpicture}
      \caption{}
      \label{fig:graph2gprime}
    \end{subfigure}
  \end{center}
  \caption{Graphs related to the example on pruning with multiple data sources: \subref{fig:graph2g} the original graph, \subref{fig:graph2gprime} the pruned graph obtained by applying Theorems~\ref{thm:Yancestors}, \ref{thm:Xancestors}, and \ref{thm:isolated}.} 
  \label{fig:graph2example}
\end{figure}

Applying do-calculus to the pruned graph and the corresponding input distributions $p(y \cond \doo(w_1), w_2)$, $p(w_1 \cond \doo(x_1,x_2), w_2)$ and $p(w_2)$, we obtain the identifying functional
\[
p(y \cond \doo(x_1,x_2)) = \sum_{w_1,w_2} p(y \cond \doo(w_1),w_2)p(w_1 \cond \doo(x_1,x_2),w_2)p(w_2).
\]
In this example, all three data sources are necessary, as the causal effect is not identifiable if any of the input distributions is unavailable. It is easy to see that $p(y \cond \doo(x_1,x_2))$ cannot be identified if $Y$, $X_1$, or $X_2$ is not present in any input distribution, and the necessity of $p(w_2)$ follows from the fact that the other two input distributions are conditioned on $W_2$ and cannot be made independent of $W_2$.

\section{Clustering Variables in Causal Graphs} \label{sec:clustering}

Clustering is a method of preprocessing, in which a group of vertices in a causal graph is instead represented by a smaller set of vertices. In particular, we are interested in clusters that retain identifiability and non-identifiability of causal effects between the original graph and the clustered graph. Clustering can be especially useful when the causal graph consists of multiple vertices with similar ``roles'' regarding the causal effect to be identified, such as variables related to the environment or demographics.

\subsection{Basic Concepts of Clustering}

We define clustering as an operation that takes a causal graph $\sG$ and a set of vertices $\+ T$ as inputs and produces a new graph where $\+ T$ is instead represented by a single vertex $T$. In addition, as we are clustering DAGs induced by causal models, a variable being clustered must always be clustered together with its dedicated error term. For this purpose, we will explicitly consider graphs where the dedicated error terms have been omitted, i.e., causal graphs over $\+ V \cup \Us$.

\begin{defin}[Clustering] \label{def:clustering}
Let $\sG(\+ V \cup \Us)$ be a causal graph and let $\+ T \subset \+ V \cup \Us$. \emph{Clustering} of $\+ T$ in $\sG$ induces a graph $\sG'(\+ V' \cup \Us')$ where $\+ V' = (\+ V \setminus \+ T) \cup T$ and $\Us' = (\Us \setminus \+ T)$, that is obtained from $\sG$ by replacing $\+ T$ with a new vertex $T$ whose parents and children in $\sG'$ are $\Pa_\sG^+(\+ T) \setminus \+ T$ and $\Ch_\sG(\+ T) \setminus \+ T$, respectively.
\end{defin}
We denote the graph $\sG'$ induced by clustering $\+ T$ into $T$ in $\sG$ by $\sG[\+ T \to T]$. Importantly, we always consider $T$ as a member of $\+ V'$ in the causal model whose induced causal graph is $\sG[\+ T \to T]$ regardless of whether $\+ T$ contains members of $\Us$. We also note that the above definition is very general, and the resulting graph may not be a DAG. For these reasons, it is important to focus on a restricted class of clusters.

We concentrate on transit clusters \citep{tikka2023clustering} which can be intuitively explained as structures where information is transmitted from the receivers of the cluster to the emitters of the cluster. We define receivers and emitters as follows.

\begin{defin}[Receiver] For a set of vertices $\+ T \subset \+ V \cup \Us$ in a causal graph $\sG(\+ V \cup \Us)$, the set of \emph{receivers} is the set
\[
  \rec[\sG](\+ T) = \{R \in \+ T \mid \Pa_\sG^+(R) \cap ((\+ V \cup \Us) \setminus \+ T) \neq \emptyset\}.
\]
\end{defin}

\begin{defin}[Emitter] For a set of vertices $\+ T \subset  \+V \cup \Us$  in a causal graph $\sG(\+V \cup \Us)$, the set of $\emph{emitters}$ is the set
\[
  \emi[\sG](\+ T) = \{E \in \+ T \mid \Ch_\sG(E) \cap ((\+ V \cup \Us) \setminus \+ T) \neq \emptyset\}.
\]
\end{defin}
In other words, a receiver is a vertex in $\+ T$ that has parents outside of $\+T$, while an emitter is a vertex in $\+ T$ that has children outside of $\+T$. Transit clusters can now be defined as follows.
\begin{defin}[Transit cluster] \label{def:transitcluster}
A non-empty set $\+ T \subset \+ V \cup \Us$ is a \emph{transit cluster} in a connected causal graph $\sG(\+ V\cup\Us)$ if the following conditions hold
\begin{enumerate}[label=(\alph*)]
  \item \label{itm:parents} $\Pa_\sG^+(R_i) \setminus \+ T = \Pa_\sG^+(R_j) \setminus \+ T$ for all pairs $R_i,R_j \in \rec[\sG](\+ T)$, 
  \item \label{itm:children} $\Ch_\sG(E_i) \setminus \+ T = \Ch_\sG(E_j) \setminus \+ T$ for all pairs $E_i,E_j \in \emi[\sG](\+ T)$, 
  \item \label{itm:no_ronsy_allowed} For all vertices $T_i \in \+ T$, there exists a receiver $R$ or an emitter $E$ such that $T_i$ and $R$ or $T_i$ and $E$ are connected via an undirected path in a subgraph of $\sG$ where all incoming edges to receivers of $\+ T$ and all outgoing edges from emitters of $\+ T$ are removed.
  \item \label{itm:forallr} If $\emi[\sG](\+ T) \neq \emptyset$ then for all $R \in \rec[\sG](\+ T)$ there exists $E \in \emi[\sG](\+ T)$ such that $E \in \De_\sG(R)$,
  \item \label{itm:foralle} If $\rec[\sG](\+ T) \neq \emptyset$ then for all $E \in \emi[\sG](\+ T)$ there exists $R \in \rec[\sG](\+ T)$ such that $R \in \An_\sG(E)$.
\end{enumerate}
\end{defin}
Conditions~(a) and (b) determine that all receivers in $\+T$ must have the same parents outside of $\+T$, and all emitters must have the same children outside of $\+T$. Condition~(c) ensures that within $\+T$, each vertex is connected to a receiver or an emitter. Conditions~(d) and (e) state that if the receivers and emitters are non-empty sets, then each receiver is connected to an emitter via a directed path, and each emitter is connected to a receiver via a directed path. Importantly, $\sG[\+T \to T]$ is a DAG when $\sG$ is a DAG and $\+ T$ is a transit cluster. 

As an illustration of transit clusters, consider the graph in Figure~\ref{fig:clustg}. The set $\+S = \{R, S_1, S_2, E_1, E_2\}$ fulfills the conditions of Definition~\ref{def:transitcluster}, and is thus a transit cluster. The set $\+ T = \{T_1, T_2\}$ also fulfills these conditions. The graph obtained by clustering both $\+ S$ and $\+ T$ is presented in Figure~\ref{fig:clustgdot}.

\begin{figure}
  \begin{center}
    \begin{subfigure}{0.55\textwidth}
      \centering
      \begin{tikzpicture}[scale=2]
        \node [dot = {0}{0}{E_2}{below}] at (0,0) {};
        \node [dot = {0}{0}{E_1}{above}] at (0,0.5) {};
        \node [dot = {0}{0}{T_2}{above}] at (0,1.5) {};
        \node [dot = {0}{0}{T_1}{above}] at (0,1) {};
        \node [dot = {0}{0}{S_2}{below}] at (-1,0.) {};
        \node [dot = {0}{0}{S_1}{above}] at (-1,0.5) {};
        \node [dot = {0}{0}{R}{below}] at (-2,0.25) {};
        \node [dot = {0}{0}{X}{above}] at (-1,1) {};
        \node [dot = {0}{0}{W_1}{below}] at (1,0.25) {};
        \node [dot = {0.15}{0}{W_2}{above}] at (1.5,0.25) {};
        \node [dot = {0}{0}{Y}{below}] at (2,0.25) {};
        \draw [->] (E_1) to [bend left=35.5] (W_2);
        \draw [->] (E_2) to [bend right=35.5] (W_2);
        \draw [->] (E_1) -- (W_1) ;
        \draw [->] (E_2) -- (W_1) ;
        \draw [->] (W_1) -- (W_2) ;
        \draw [->] (W_2) -- (Y) ;
        \draw [->] (S_2) -- (E_2) ;
        \draw [->] (S_2) -- (E_1) ;
        \draw [->] (S_1) -- (E_1) ;
        \draw [->] (S_1) -- (S_2) ;
        \draw [->] (R) -- (S_2) ;
        \draw [->] (R) -- (S_1) ;
        \draw [->] (X) -- (R) ;
        \draw [->] (T_1) -- (X) ;
        \draw [->] (T_2) -- (X) ;
        \draw [->] (T_1) -- (W_1) ;
        \draw [->] (T_2) -- (W_1) ;
      \end{tikzpicture}
      \caption{}
      \label{fig:clustg}
    \end{subfigure}
    \hfill
    \begin{subfigure}{0.40\textwidth}
      \centering
      \begin{tikzpicture}[xscale=2.25,yscale=1.5]
        \node [dot = {0}{0}{S}{below left}] at (0,0) {};
        \node [dot = {0}{0}{W_2}{above}] at (1.33,0) {};
        \node [dot = {0}{0}{X}{above}] at (0,1) {};
        \node [dot = {0}{0}{T}{above}] at (0.66,1) {};
        \node [dot = {-0}{0}{W_1}{above right}] at (0.66,0) {};
        \node [dot = {-0}{0}{Y}{below}] at (2,0) {};
        \draw [->] (S) -- (W_1) ;
        \draw [->] (W_2) -- (Y) ;
        \draw [->] (X) -- (S) ;
        \draw [->] (W_1) -- (W_2) ;
        \draw [->] (T) -- (W_1) ;
        \draw [->] (T) -- (X) ;
        \draw [->] (S) to [bend right=45] (W_2);
      \end{tikzpicture}
      \caption{}
      \label{fig:clustgdot}
    \end{subfigure}
  \end{center}
  \caption{Graphs for the example on clustering with multiple data sources: \subref{fig:clustg} the original graph, \subref{fig:clustgdot} the clustered graph where the sets $\{R, S_1, S_2, E_1, E_2\}$ and $\{T_1, T_2\}$ have been clustered as $S$ and $T$, respectively.}
\end{figure}

Clustering a causal graph not only has an impact on the graph itself but also on the available input distributions. We define clustered input distributions for identifiability considerations as follows. 

\begin{defin}[Clustered input distributions] \label{def:clusteredinput}
Let $\sI = \{p(\+ a_i \cond \doo(\+ b_i), \+ c_i)\}_{i=1}^n$ be a set of input distributions and let $\+ T \subset \+ V \cup \Us$ be a transit cluster such that $\Em_\sG(\+T)\cap \Us = \emptyset$ in a causal graph $\sG(\+ V \cup \Us)$. If for each $i = 1,\ldots,n$, exactly one of the following conditions holds
\begin{enumerate}[label=(\alph*)]
  \item $\emi_\sG(\+T) \subseteq \+A_i$ and $\+ T \cap (\+ B_i \cup \+ C_i) = \emptyset$, 
  \item $\emi_\sG(\+T) \subseteq \+B_i$ and $\+ T \cap (\+ A_i \cup \+ C_i) = \emptyset$, 
  \item $\emi_\sG(\+T) \subseteq \+C_i$ and $\+ T \cap (\+ A_i \cup \+ B_i) = \emptyset$, 
  \item $\+T \cap (\+A_i \cup \+B_i \cup \+C_i) = \emptyset$,
\end{enumerate}
and at least one input distribution satisfies one of the conditions (a)--(c)
then $\sI$ is \emph{compatible} with $\+ T$ and the set of \emph{clustered input distributions} with respect to $\+ T$ is $\sI[\+ T \to T] = \{ p(\+ a'_i \cond \doo(\+ b_i'), \+ c'_i) \}_{i = 1}^n$ where 
\[
  \+ W_i' = \begin{cases}
    \+ W_i & \textrm{if } \+ W_i \cap \+T = \emptyset, \\
    (\+ W_i \cup T) \setminus \+T & \textrm{otherwise},
  \end{cases}, \text{ for } i = 1\,\ldots,n \text{ and } \+ W_i = \+ A_i, \+ B_i, \+ C_i,
\]
and $T$ is a new variable that represents $\+ T$ after clustering.
\end{defin}

Conditions (a)--(c) can be summarized as a requirement that the cluster is not split among the sets $\+ A_i$, $\+ B_i$, and $\+ C_i$, and that at least the emitters of the cluster are present in one of these sets. We exclude the case where $\+T$ is not present in any input distribution by requiring that one of the conditions (a)--(c) holds for at least one input distribution.

If the emitters are not entirely present in $\+A_i$, $\+B_i$, or $\+C_i$, condition (d) states that no other member of $\+ T$ can belong to $\+A_i$, $\+B_i$, or $\+C_i$. If exactly one of the conditions holds for each input distribution, we can replace the members of $\+ T$ in each input distribution by $T$. We highlight that considering only the emitters of $\+ T$, instead of $\+T\cap\+V$, is sufficient to represent a set of input distributions as a set of clustered input distributions. Intuitively, this can be understood by noting that information flows out from a transit cluster only through the emitters. This sufficiency is further demonstrated by the results of Section~\ref{subsec:clusterid}. We note that there exists a one-to-one mapping between the clustered input distributions and their original counterparts. This property is important for using identifying functionals obtained from the clustered causal graph in the original graph, as we discuss in Section~\ref{sec:functionals}.

Importantly, Definition~\ref{def:clusteredinput} rules out scenarios where only some, but not all emitters are observed in a data source. For example, consider the causal graph of Figure~\ref{fig:nouemiex1} where $U \in \Us$. If we ignore compatibility and consider $\{Z,U\}$ as a transit cluster, we can obtain the graph of Figure~\ref{fig:nouemiex2}, where by Definition~\ref{def:clustering}, $T \in \+ V'$. It is clear that $p(y \cond \doo(x))$ is not identifiable from $p(x,y,z)$ in the original graph but is identifiable from $p(x,y,t)$ in the clustered graph. It is crucial that scenarios like this are ruled out. 

\begin{figure}[!ht]
  \begin{center}
    \begin{subfigure}{0.25\textwidth}
      \centering
      \begin{tikzpicture}[scale=2]
        \node [dot = {0}{0}{X}{below}] at (0,0) {};
        \node [dot = {0}{0}{Y}{below}] at (1,0) {};
        \node [lat = {0}{0}{U}{above}] at (1,0.67) {};
        \node [dot = {0}{0}{Z}{above}] at (0,0.67) {};
        \draw [->] (X) -- (Y);
        \draw [->] (Z) -- (Y);
        \draw [->] (Z) -- (X);
        \draw [->,dashed] (U) -- (X);
        \draw [->,dashed] (U) -- (Y);
      \end{tikzpicture}
      \caption{}
      \label{fig:nouemiex1}
    \end{subfigure}
       \begin{subfigure}{0.25\textwidth}
      \centering
      \begin{tikzpicture}[scale=2]
        \node [dot = {0}{0}{X}{below}] at (0,0) {};
        \node [dot = {0}{0}{Y}{below}] at (1,0) {};
        \node [dot = {0}{0}{T}{above}] at (0.5,0.67) {};
        \draw [->] (X) -- (Y);
        \draw [->] (T) -- (X);
        \draw [->] (T) -- (Y);
      \end{tikzpicture}
      \caption{}
      \label{fig:nouemiex2}
    \end{subfigure}
  \end{center}
  \caption{Causal graphs demonstrating the importance of the compatibility of the input distributions for the identification invariance of clustering.}
  \label{fig:nouemi}
\end{figure}

We present the following two lemmas that characterize transit clusters and are needed in later results that concern identification invariance. The first lemma states that a transit cluster remains a transit cluster after the removal of incoming or outgoing edges of a set of vertices.

\begin{lemrep} \label{thm:transitcluster_edgeremoval}
A transit cluster $\+T$ in $\sG$ is also a transit cluster in $\sG[\overline{\+ Z},\underline{\+ W}]$ where  $\+Z$ and $\+W$ are disjoint, possibly empty, subsets of $\+V$ and  $\+T$ is either a subset of $\+Z$, a subset of $\+W$, or disjoint from both $\+Z$ and $\+W$.
\end{lemrep}
\begin{proof}
By definition, $\+T$ is either a subset of $\+Z$, a subset of $\+W$, or disjoint from both $\+Z$ and $\+W$. If $\+T \subseteq \+Z$ then $\+T$ does not have parents in $\sG[\overline{\+ Z},\underline{\+ W}]$ and the set of receivers of $\+ T$ is empty. If $\+T \subseteq \+W$ then $\+T$ does not have children in $\sG[\overline{\+ Z},\underline{\+ W}]$ and the set of emitters of $\+ T$ is empty. In both cases, $\+T$ is a transit cluster in $\sG[\overline{\+ Z},\underline{\+ W}]$ after these operations. Finally, we consider the case where $\+T$ is disjoint from both $\+Z$ and $\+W$. The transit cluster remains unchanged if $\+Z$ does not contain children of $\+T$ and $\+W$ does not contain parents of $\+T$. If $\+Z$ contains a child of $\+T$, all edges between $\+T$ and this child are removed. If $\+W$ contains a parent of $\+T$, all edges between this parent and $\+T$ are removed. In these cases, there are no other changes and $\+T$ remains as a transit cluster after these operations. We conclude that the parents and the children of $\+T$ may change when edges are removed but in all cases $\+T$ is a transit cluster in $\sG[\overline{\+ Z},\underline{\+ W}]$.
\end{proof}

The second lemma concerns d-separation in clustered and unclustered graphs.
\begin{lemrep} \label{thm:transitcluster_dseparation}
Let $\+T$ be a transit cluster in $\sG(\+V \cup \Us)$ and let $\sG'(\+V' \cup \Us ') = \sG[\+T \rightarrow T]$. Let $\+X'$, $\+Y'$ and $\+W'$ be disjoint subsets of $\+V'$ and define $\+X$, $\+Y$ and $\+W$ to be the corresponding subsets of $\+V$ when $T$ is replaced by $\+T$. Then $\+X'$ and $\+Y'$ are d-separated by $\+W'$ in $\sG'$ if and only if $\+X$ and $\+Y$ are d-separated by $\+W$ in $\sG$. 
\end{lemrep}
\begin{proof}
We choose arbitrary $V_1 \in \+X'$ and arbitrary $V_2 \in \+Y'$ and consider a path $s$ between $V_1$ and $V_2$ in $\sG$. First, we study the case where $s$ does not contain members of $\+T$. This means that the vertices of the path remain unchanged after clustering $\+T$ and we may denote $s' = s$ where $s'$ is the corresponding path in $\sG'_*$. If $T \notin \+W'$, clustering $\+T$ as $T$ does not affect whether $s$ is open or not. If $T \in \+W'$, the properties of the transit cluster ensure that $s$ is blocked by $\emi_\sG(\+T)$ in $\sG_*$ if and only if it is blocked by $T$ in $\sG'_*$.

Next, we study the more complicated case where $s$ contains members of $\+T$. First, we assume that $V_1 \neq T$, and $V_2 \neq T$. In this case, there may not be a corresponding path in $\sG'_*$ because path $s$ can go through $\+T$ multiple times in $\sG_*$ which is not possible in $\sG'_*$. To address this challenge, we show that there exists in $\sG_*$ a path $s^*$ from $V_1$ to $V_2$ that goes through $\+T$ only once. To construct the path $s^*$, we start by writing the path $s$ as follows
\[
s = V_1,s_1,T_1,s_0,T_2,s_2,V_2,
\]
where $s_0$, $s_1$, and $s_2$ are paths, $T_1$ is the first member of $\+T$ on path $s$ when starting from $V_1$, and $T_2$ is the first member of $\+T$ on path $s$ when starting from $V_2$. In other words, $s_1$ and $s_2$ do not contain members of $\+T$ but $s_0$ may contain members and non-members of $\+T$. We define path $s^*$ as follows 
\[
s^* = 
\begin{cases}
V_1,s_1,T_1,s^*_0,T_2,s_2,V_2 & \begin{aligned} &\textrm{if } (T_1 \in \rec_{\sG_*}(\+T) \textrm{ and } T_2 \in \emi_{\sG_*}(\+T)) \textrm{ or } \\
& \quad (T_1 \in \emi_{\sG_*}(\+T) \textrm{ and } T_2 \in \rec_{\sG_*}(\+T)), \end{aligned} \\
V_1,s_1,T_1,s_2,V_2 & \textrm{if } T_1,T_2 \in \rec_{\sG_*}(\+T) \textrm{ or } T_1,T_2 \in \emi_{\sG_*}(\+T),
\end{cases}
\]
where the path $s^*_0$ contains only members of $\+T$. The existence of $s^*_0$ follows from conditions~\ref{itm:forallr} and \ref{itm:foralle} of Definition~\ref{def:transitcluster}. In the case $T_1,T_2 \in \rec_{\sG_*}(\+T)$ or $T_1,T_2 \in \emi_{\sG_*}(\+T)$, the existence of edge from $T_1$ to path $s_2$ follows from conditions~\ref{itm:parents} and \ref{itm:children} of Definition~\ref{def:transitcluster}. If path $s^*$ is blocked, then path $s$ is blocked as well because the vertices of $s^*$ that are outside $\+T$ are also in $s$. This means we may consider only paths like $s^*$ that go through $\+T$ only once. 

For each path $s^*$ in $\sG_*$, there exists the corresponding path $s'$ in $\sG'_*$ defined as follows
\[
s' = V_1,s_1,T,s_2,V_2.
\]

The vertices other than $T$ and members of $\+T$ are the same on the paths $s^*$ and $s'$. If $s'$ is blocked by a vertex that is not $T$, $s^*$ is blocked by the same vertex. If $s'$ is blocked by $T$ and $T$ is a collider then $s^*$ is blocked by a receiver that is a collider in $\sG_*$. If $s'$ is blocked by $T$ when $T$ is not a collider then $s^*$ is blocked by $\emi_\sG(\+T)$ and by Definition~\ref{def:clusteredinput}, no emitter can be a member of $\Us$. We conclude that if $s'$ is blocked in $\sG'_*$ then $s^*$ is blocked in $\sG$. If all paths between $V_1$ and $V_2$ given $\+W'$ are blocked in $\sG'_*$ then all paths between $V_1$ and $V_2$ given $\+W$ are blocked in $\sG_*$. By the properties of a transit cluster, all paths between $V_1$ and $V_2$ are blocked also if conditioning set $\+W$ contains other members of $\+T$ in addition to $\emi_\sG(\+T)$. 

Finally, we consider the case $V_1=T$. Let $T_1 \in \+T$ be the first vertex on path $s$ that can be now written as
\[
s = T_1,s_0,T_2,s_2,V_2,
\] 
where $s_0$ and $s_2$ are paths, and $T_2$ is the first member of $\+T$ on path $s$ when starting from $V_2$. The conditions~\ref{itm:forallr} and \ref{itm:foralle} of Definition~\ref{def:transitcluster} allow us to define path $s^*$ as 
\[
s^* = T_1,s^*_0,T_2,s_2,V_2, 
\] 
where the path $s^*_0$ contains only members of $\+T$. Noting that $\+T \cap \+W = \emptyset$, we conclude $s^*$ is blocked by $\+W$ in $\sG$ if and only if  $s' = T,s_2,V_2$ is blocked by $\+W'=\+W$ in $\sG'$. The case $V_2=T$ is analogous.

In all cases, $V_1$ and $V_2$  are d-separated given $\+W'$ in $\sG'$ if and only if $V_1$ and $V_2$  are d-separated given $\+W$ in $\sG$. As $V_1$, $V_2$ and $\+W'$ were chosen arbitrarily, the same conclusion applies to all choices. The claim follows because sets $\+X$ and $\+Y$ are by definition d-separated if and only if all paths between them are d-separated. 
\end{proof}

In Lemma~\ref{thm:transitcluster_dseparation}, the sets $\+X$, $\+Y$ and $\+W$ are obtained from disjoint sets $\+X'$, $\+Y'$ and $\+W'$, respectively. This rules out cases where the transit cluster is divided between the sets $\+X$, $\+Y$ and $\+W$ because $T$ can be a member of only one of the sets $\+X'$, $\+Y'$ and $\+W'$ in $\sG'$. Together Lemmas~\ref{thm:transitcluster_edgeremoval} and \ref{thm:transitcluster_dseparation} imply that the rules of do-calculus are applicable in a clustered graph $\sG'$ if and only if they are applicable in the unclustered graph $\sG$. More precisely, whenever a transit cluster $\+ T$ is not split among the sets $\+ X, \+ Y, \+ Z$ and $\+ W$ that appear in the rules, and whenever $\+ T$ is present for a rule that applies in $\sG$, then the rule applies with $\+ T$ replaced by $T$ in $\sG'$ and vice versa. If $\+ T$ is not present, then the applicability of the rule is always preserved between the graphs.

\subsection{Identification Based on Clustered Causal Graphs} \label{subsec:clusterid}

\begin{toappendix}
  \label{apx:clusteringproof}
\end{toappendix}

We present conditions for determining the identification invariance of clustering with multiple data sources. We consider the identification invariance separately based on the identifiability of the causal effect in the clustered graph. Theorems~\ref{thm:invariance_id} and \ref{thm:invariance_nonid} consider identifiable and non-identifiable scenarios, respectively. In cases where the causal effect is not identifiable in the clustered graph, the input distributions must satisfy additional conditions. These conditions are verified by Algorithm~\ref{alg:idinvariant}. The proofs of this section are presented in Appendix~\ref{apx:clusteringproof}.

\begin{algorithm}[!ht]
  \begin{algorithmic}[1]
  \Function{VerifyInputs}{$\sG', \sI', T$}
    \If{$\PaVU_{\sG'}(T) = \emptyset$} \Return \textbf{TRUE} \label{line:noparents} \EndIf
    \ForAll{$\sI_i' \in \sI'$} \label{line:forall}
      \If{$T \in \+A_i' \cup \+B_i' \cup \+C_i'$} \label{line:abc}
        \If{$\+C_i' \cap \De_{\sG'}(T) \neq \emptyset$} \textbf{return FALSE}\label{line:cdesc1} \EndIf
        \State $\+D \gets \De_{\sG'}(T) \cap \+A_i'$
        \If{$T \in \+A_i'$ \textbf{and} $(\+D\independent T \cond \+B_i', \+C_i', \+A_i'\setminus(\+D \cup T))_{\sG'[\overline{\+B_i'}, \underline{T}]}$} \textbf{continue} \label{line:aparents} \EndIf
        \If{$T\in \+B_i'$} \textbf{continue} \label{line:bparents} \EndIf 
        \If{$T\in \+C_i'$ \textbf{and} $(\+A'_i\independent T \cond \+B_i', \+C_i' \setminus T)_{\sG'[\overline{\+B_i'}, \underline{T}]}$} \textbf{continue} \label{line:cparents} \EndIf
      \Else \label{line:missingt}
        \If{$\exists \sI_j' \in \sI'$ s.t. $T \in \+A_j'$  \textbf{and} $\+B_j' = \emptyset$  \textbf{and} $\+C_j' = \emptyset$} \label{line:emptybc}
          \If{$(\+A_i' \independent T \cond \+B_i', \+C_i')_{\sG'[\overline{\+B_i'}, \underline{T}]}$ \textbf{and} $(T \independent \+C_i' \cond \+B_i')_{\sG'[ \overline{\+B_i'}]}$ \textbf{and} $(T \independent \+B_i')_{\sG'[\overline{\+B_i'}]}$} \textbf{continue} \EndIf \label{line:docalcrule2}
        \EndIf
        \If{$(\+A_i' \independent T \cond \+B_i, \+C_i)_{\sG'[\overline{\+B_i'}, \overline{T(\+C_i')}]}$} \label{line:docalcrule3} \textbf{continue}
        \EndIf
        \If{$\exists \sI_k'\in\sI'$ s.t. $\+A_i'\subseteq \+A_k'$  \textbf{and} $\+B_k' = \+B_i' \cup T$  \textbf{and} $\+C_k' = \+C_i'$} \textbf{continue} \label{line:anotherinput}
        \EndIf
      \EndIf
        \State \Return \textbf{FALSE}
    \EndFor
    \State \Return \textbf{TRUE} \label{line:return_true}
  \EndFunction
  \end{algorithmic}
  \caption{Verify the clustered input distributions to assess identification invariance of clustering $\+ T$ for a non-identifiable causal effect $p(\+y\cond \doo(\+x))$, where $\+X \cap \+T = \emptyset$, $\+Y \cap \+T = \emptyset$. The inputs are a clustered causal graph $\sG' = \sG[\+ T \to T]$, a set of clustered input distributions $\sI' = \sI[\+ T \to T]$, and the cluster vertex $T$ that represents the cluster $\+T$.}
  \label{alg:idinvariant}
\end{algorithm}

\begin{thmrep} \label{thm:invariance_id}
  Let $\sG(\+V\cup\Us)$ be a causal graph where $\+T \subset \+V \cup \Us$ is a transit cluster and let $\sI$ be a set of input distributions compatible with $\+ T$. Let $\sI' = \sI[\+ T \to T]$ in $\sG' = \sG[\+ T \to T]$. If $p(\+y \cond \doo(\+x))$ is identifiable from $\sI'$ in $\sG'$, then it is identifiable from $\sI$ in $\sG$, where $\+X \cup \+Y \subset \+V$, $\+Y \cap \+X = \emptyset$, $\+X \cap \+T = \emptyset$, $\+Y \cap \+T = \emptyset$.
\end{thmrep}
\begin{proof}
Assume that $p(\+y \cond \doo(\+x))$ is not identifiable in $\sG$ from $\sI$. By Definition~\ref{def:identifiability}, non-identifiability implies that there exist two models $\sM_1$ and $\sM_2$ for $\sG$ for which the causal effects differ, but the input distributions agree. To show the non-identifiability in $\sG'$, we need to find two models $\sM'_1$ and $\sM'_2$ with the same properties. We aim to construct these models $\sM'_1$ and $\sM'_2$ based on $\sM_1$ and $\sM_2$. We show only how $\sM'_1$ is constructed from $\sM_1$ because the construction is done similarly for $\sM'_1$ and $\sM'_2$. According to Definition~\ref{def:clustering}, we have  $\+ V' = (\+ V \setminus \+ T) \cup T$ and $\+ U' = (\+ U \setminus \+ T) \cup U[T]$. We define $U[T]$ as a vector-valued random variable whose components are $\+U[\+T]$ and set $p'(u[T])$ in $\sM'_1$ to be equal to $p(\+ t \cap \+ u)$ in $\sM_1$. Other members of $\+U'$ are distributed as they are distributed in $\sM_1$. Since $\+T$ is a transit cluster, we may factorize the distribution $p'(\+u')$ as $p'(\+u')=p(\+ u \setminus \+ t)p'(u[T])$.
  
We define every variable in $\+V'$, apart from $T$ and $\Ch_{\sG'}(T)$, using the same structural equations as in $\sM_1$, i.e., for each $V_j \notin \Ch_{\sG'}(T) \cup T$, we set $f'_j = f_j$. The random vector $T=(T_1,\ldots,T_m)$ contains as many components as there are emitters in $\+T$. The structural equations for the components of $T$ are constructed by the following iterative procedure: 
\begin{enumerate}
\item Copy the structural equation of emitter $E_j$ as the structural equation for component~$T_j$.
\item If $T_j$ is a function of a variable $Z \in (\+T \cap \+V)$, variable $Z$ is replaced by the function $f_Z(\Pa(Z),\+U[Z])$. 
\item Step 2 is repeated until the structural equation does not contain any members of $\+T \cap \+V$.
\item Each member of  $\+U[\+T]$ in the structural equation is replaced by the corresponding component of $U[T]$.
\end{enumerate}
As a result of the procedure, each component of $T$ becomes a function of $U[T]$ and the parents of $\+T$ that are not members of $\+ T$. The structural equations for $\Ch_{\sG'}(T)$ are otherwise the same as the structural equations for $\Ch_{\sG}(\+T)$ but each emitter $E_j$ is replaced by the corresponding component $T_j$ of $T$. By this construction, we have $p'(T)=p(\emi_\sG(\+T))$ and $p'(\Ch_{\sG'}(T)) = p(\Ch_\sG(\+T))$. As the structural equations were copied directly from $\sM_1$ to $\sM'_1$, the construction does not change the functional relationships. This implies that the causal effects differ between $\sM'_1$ and $\sM'_2$ while all input distributions are equal. Thus the claim follows by contraposition.
\end{proof}

\begin{thmrep} \label{thm:invariance_nonid}
  Let $\sG(\+V\cup\Us)$ be a causal graph where $\+T \subset \+V \cup \Us$ is a transit cluster and let $\sI$ be a set of input distributions compatible with $\+ T$. Let $\sI' = \sI[\+ T \to T]$ in $\sG' = \sG[\+ T \to T]$. 
  If $p(\+y \cond \doo(\+x))$ is not identifiable from $\sI'$ in $\sG'$ and if \checkid{}$(\sG', \sI', T)$ returns \textbf{\em{TRUE}}, then $p(\+y \cond \doo(\+x))$ is not identifiable from $\sI$ in $\sG$, where $\+X \cup \+Y \subset \+V$, $\+Y \cap \+X = \emptyset$, $\+X \cap \+T = \emptyset$, $\+Y \cap \+T = \emptyset$.
\end{thmrep}
\begin{proof}
By Definition~\ref{def:identifiability}, non-identifiability implies that there exist two models $\sM_1'$ and $\sM_2'$ for $\sG'$ for which the causal effects differ but the input distributions agree. Let $\sG(\+ V \cup \Us)$ be an arbitrarily chosen causal graph such that $\+T$ is a transit cluster in $\sG$ and $\sG' = \sG[\+T \rightarrow T]$.  
To ensure the positivity of the joint distribution, $\+U[\+T]$ must contain a dedicated error term for each member of $\+T \cap \+V$. 
  
To show non-identifiability in $\sG$, we need to find two models $\sM_1$ and $\sM_2$ for which the causal effects differ but the input distributions agree. We construct these models based on $\sM_1'$ and $\sM_2'$. As the proof for non-identifiability is long, it will be divided into several parts. First, we will show how models $\sM_1$ and $\sM_2$ are constructed. Next, we will show that the input distributions are the same for $\sM_1$ and $\sM_2$. Finally, we will show that the causal effects are different between $\sM_1$ and $\sM_2$. 
  
\emph{Constructing the models:} By Definition~\ref{def:identifiability}, non-identifiability implies that we have two SCMs, $\sM_1' =(\+V', \+U',\+F_1',p_1')$ and $\sM_2' =(\+V', \+U',\+F_2',p_2')$, for $\sG'$ for which the causal effects differ but the input distributions are the same. We construct two models $\sM_1$ and $\sM_2$ for $\sG$ so that for each input distribution $p_1(\+a_i \cond \doo(\+b_i), \+c_i) = p_2(\+a_i \cond \doo(\+b_i), \+c_i)$ and $p_1(\+y\cond \doo(\+x)) \neq p_2(\+y\cond \doo(\+x))$. We show in detail how $\sM_1=(\+V, \+U,\+F_1,p_1)$ is constructed based on $\sM_1'$. Model $\sM_2$ is defined analogously based on $\sM_2'$ instead. Without loss of generality, we assume that in $\sM_1'$ and $\sM_2'$, all variables in $\+V'$ are scalar-valued, including $T$. 
We let $\+V=(\+V' \setminus T) \cup \+T$ and $\+U=(\+U'\setminus \ud[T]) \cup \Ud[\+T] \cup \Us[\+T]$. We let each variable in $\Ud[\+T] \cup \Us[\+T]$ be independent of the other variables in $\+U$ and uniformly distributed on the interval $(0,1)$. We define $p_1(\+u)=p_1'(\+u') p_1(\smallud[\+ t])p_1(\smallus[\+ t])$. It remains to construct the functions $\+F_1$ that define the structural equations for the variables in $\+V$. From Definition~\ref{def:transitcluster} it follows $\Ch_{\sG'}(T) \setminus T = \Ch_\sG(\+T)\setminus\+T = \Ch_\sG(\emi_\sG(\+T)) \setminus \+T$. We define every variable, apart from $\Ch_\sG(\+T) \cup \+T$, using the same structural equations as in $\sM_1'$, i.e., for each $V_j \notin \Ch_\sG(\+T) \cup \+T$, we let $f_j = f_j'$. This is possible because these vertices have the same incoming edges before and after the clustering and therefore share the same set of parents in both cases. 
  
Next, we define $f_j$ for each $V_j \subset \+ T \cap \+ V$. If $\rec_\sG(\+T)\neq \emptyset$, we select a receiver $R_1 \in \rec_\sG(\+T)$ that does not have other receivers of $\+ T$ as its parents. Definition~\ref{def:transitcluster} guarantees that if also $\emi_\sG(\+T) \neq \emptyset$, there must be a directed path from $R_1$ to some emitter $E_1 \in \emi_\sG(\+T)$. If $\rec_\sG(\+T) = \emptyset$, we choose any $T_j \in \An_\sG(E_1) \cap \+T$, for which $\Pa_\sG(T_j) = \emptyset$ as $R_1$, and if $\emi_\sG(\+T) = \emptyset$ we choose any $T_j \in \De_\sG(R_1) \cap \+T$ as $E_1$. Without loss of generality, we may assume that the length of this directed path is greater than one, as the claim follows directly from Theorem~\ref{thm:singlelayerid} otherwise. If the cluster has no receivers, then we can follow similar reasoning as in the proof of Theorem~\ref{thm:singlelayerid}, and simply select any emitter of $\+ T$ and copy the structural equation of $T$ as the structural equation of this emitter, from which the equality of input distributions and the difference of causal effects follows directly.

The directed path to the emitter is written as $R_1 \rightarrow T_1 \rightarrow \ldots \rightarrow T_k \rightarrow E_1$. Now, let the dedicated error term for $R_1$ be $\ud[T]$, i.e. $\ud[R_1]=\ud[T]$. Let $R_1$ have the same structural equation as $T$ had in $\sM_1'$, while other members of $\+ U$ that are parents of $R_1$ are set to not affect $R_1$, i.e. $R_1 = f_{R_1}(\Pa(R_1), {\+U}[R_1]) = f_{R_1}(\Pa(T), \ud[T]) = f_{T}(\Pa(T), \ud[T]) = T$. To define the structural equations for the rest of the vertices in the directed path, we apply the following inheritance function
\[
h(X, U) = \begin{cases} 
  X & \text{if } U \leq \gamma, \\
  g\left(\frac{U}{1 - \gamma}\right) & \text{if } U > \gamma,
\end{cases}
\]
where $\gamma \in (0,1)$ is a fixed probability parameter and $g$ is either the quantile function of the marginal distribution of $T$ in $\sM_1'$ if there exists $\sI_i' \in \sI'$, such that $T \in \+A_i'$ and $\+B_i' \cup \+C_i' = \emptyset$. Otherwise, if such an input distribution does not exist, $g$ is the quantile function of any arbitrary distribution over the domain of $T$ in both models. If $T$ was vector-valued, we would define $g$ as a multivariate quantile function.
  
Let $\ud[T_1],\ldots, \ud[T_k]$ and $\ud[E_1]$ be the dedicated error terms that are the parents of $T_1,\ldots,T_k$ and $E_1$, respectively. Now, we define $T_1 = h(R_1, \ud[T_1])$, $T_{\ell} = h(T_{\ell-1}, \ud[T_{\ell}])$, $2 \leq \ell \leq k$, and $E_1 = h(T_k, \ud[E_1])$. In other words, there is a probability of $\gamma$ that a vertex of the chain inherits the value from its parent vertex in the chain. If the value is not inherited, then the value is drawn from a distribution that has the same domain as $T$. This assures that every vertex in the chain has the same domain of values as $T$ and any interventional distribution $p(\+y \cond \doo(T=t))$ in $\sG'$ can have a well-defined counterpart in $\sG$. 

We define the members of $\+ T$ that are not on the path $R_1 \rightarrow T_1 \rightarrow \cdots \rightarrow T_k \rightarrow E_1$ to be Bernoulli-distributed random variables with a probability parameter $\theta$ that are independent of all other variables. The remaining undefined variables in $\sM_1$ are the members of $\Ch_\sG(\+T) \setminus \+T$. In $\sM_1'$, each $V_j \in \Ch_{\sG'}(T)$ is defined as a function of $T$ and the other parents $\PaVU_{\sG'}(V_j)\setminus T$. In $\sM_1$, we let each $V_j$ have the same function, but with $T$ replaced by $E_1$, i.e., $f_{V_j'}(E_1=t, \Pa(V_j)\setminus E_1)=f_{V_j'}(T=t, \Pa(V_j')\setminus T)$. By this maneuver, each $V_j \in \Ch_\sG(\+T) \setminus \+T$ now has the same set of possible values in $\sM_1$ and $\sM_1'$ and for all $V_j \in \Ch_\sG(\+T) \setminus \+T$ it holds 
\begin{equation} \label{eq:descendantsdo}
  p_1(v_j \cond \doo(E_1=t,\+b),\+c) = p'_1(v_j \cond \doo(T=t,\+b),\+c),
\end{equation} 
where $\+B$ and $\+C$ are disjoint subsets of $\+V'$. As the structural equations are the same in $\sM_1$ and $\sM_1'$ for variables not in $\Ch_\sG(\+T) \cup \+T$, we can extend this conclusion to all $V_j \in \+ V \setminus \+ T$.
  
\emph{Equality of Input Distributions:} In this part, we show that having constructed the models as was explained in the previous part, the input distributions are the same in $\sM_1$ and $\sM_2$. In other words, we must show that for each $i = 1,\ldots,n$ it holds that
\[
  p_1(\+a_i\cond \doo(\+b_i), \+c_i) = p_2(\+a_i\cond \doo(\+b_i), \+c_i).
\]
We divide the proof into four parts: (I) when $T \in \+A'_i$, (II) when $T \in \+B'_i$, (III) when $T \in \+C'_i$, and (IV) when $T$ is not a member of $\+A'_i$, $\+B'_i$ or $\+C'_i$. 
  
(I) Let $T \in \+A'_i$. Because $T \in \+ A'_i$, by Definition~\ref{def:clusteredinput} it must be the case that $\emi_\sG(\+ T) \subseteq \+ A_i$. Let us denote $\+D = \De_\sG'(T) \cap \+A'_i$, $\+Q = \+A_i'\setminus(\+D\cup T)$, and $\+T^* = \+T\cap \+A_i$. Now, the clustered input distribution can be factorized in the following way:
\begin{equation} \label{eq:clusteredfactorization}
  \begin{aligned}
    p'_1(\+a'_i \cond \doo(\+b'_i), \+c'_i) &= p'_1(t, \+q, \+d  \cond \doo(\+b'_i), \+c'_i) \\
    &= p'_1(\+d\cond \doo(\+b'_i), \+c'_i, t,\+q)p'_1(t\cond \doo(\+b_i), \+c_i, \+q)p'_1(\+q \cond \doo(\+b'_i), \+c'_i). 
  \end{aligned}
\end{equation}
As $p'_1(\+a'_i \cond \doo(\+b'_i), \+c'_i) = p'_2(\+a'_i \cond \doo(\+b'_i), \+c'_i)$, 
it holds that 
\begin{align} 
p_1'(t\cond \doo(\+b_i), \+c_i, \+q) &= p_2'(t \cond \doo(\+b_i), \+c_i, \+q), \label{eq:Tequal} \\ 
p_1'(\+q \cond \doo(\+b_i), \+c_i) &= p_2'(\+q \cond \doo(\+b_i), \+c_i), \textrm{ and} \label{eq:Qequal} \\
p'_1(\+d\cond \doo(\+b_i), \+c_i, t,\+q) &= p'_2(\+d\cond \doo(\+b_i), \+c_i, t,\+q).
\end{align} 
When the condition  $(\+D \independent T \cond \+B_i', \+C_i', \+Q)_{\sG'[\overline{\+B_i'}, \underline{T}]}$ on line~\ref{line:aparents} holds,  we can use the second rule of do-calculus to obtain:
\begin{equation} \label{eq:Wequal2}      
  p'_1(\+d\cond \doo(\+b_i), \+c_i, t,\+q) = p'_1(\+d\cond \doo(\+b_i, t), \+c_i,\+q). 
\end{equation}
We can write similarly for $M_2'$. For $\sM_1$, using the chain rule, we can write
\begin{align*}
p_1(\+a_i \cond \doo(\+b_i), \+c_i) &= p_1(\+t^*, \+q, \+d \cond \doo(\+b_i), \+c_i) \\
&= p_1(\+d\cond \doo(\+b_i), \+c_i, \+t^*, \+q) \\
&\quad\times p_1(\+t^*\cond \doo(\+b_i), \+c_i, \+q)p_1(\+q\cond \doo(\+b_i), \+c_i).
\end{align*}
  
By Lemmas~\ref{thm:transitcluster_edgeremoval} and \ref{thm:transitcluster_dseparation}, $(\+D \independent T \cond \+B_i', \+C_i', \+Q)_{\sG'[\overline{\+B_i'}, \underline{T}]}$ implies $(\+D \independent \+T^* \cond \+B_i, \+C_i, \+Q)_{\sG[\overline{\+B_i}, \underline{\+T\cap \+A_i}]}$. Again, using the second rule of do-calculus we obtain
\begin{equation} \label{eq:originalfactorization}
  p_1(\+a_i \cond \doo(\+b_i), \+c_i) = p_1(\+d\cond \doo(\+b_i, \+t^*), \+c_i, \+q) 
  p_1(\+t^*\cond \doo(\+b_i), \+c_i, \+q) 
  p_1(\+q\cond \doo(\+b_i), \+c_i). 
\end{equation}
Now, we can derive $p_2(\+a_i \cond \doo(\+b_i), \+c_i)$ similarly as $p_1(\+a_i \cond \doo(\+b_i), \+c_i)$ was derived, which yields similar formulas for the input distributions but with $p_1$ replaced by $p_2$. We will proceed by showing that each term in equation~\eqref{eq:originalfactorization} is equal to the corresponding term in the factorization for $p_2(\+a_i \cond \doo(\+b_i), \+c_i)$.
  
The condition of line~\ref{line:cdesc1} assures that $\+C_i$ does not contain any descendants of the cluster. This means that we can use the rule 3 of do-calculus to conclude the following equalities 
\[
  \begin{aligned}
    p'_1(\+q\cond \doo(\+b_i), \+c_i) &= p'_1(\+q\cond \doo(\+b_i, T=t), \+c_i), \\
    p'_2(\+q\cond \doo(\+b_i), \+c_i) &= p'_2(\+q\cond \doo(\+b_i, T=t), \+c_i), \\
    p_1(\+q\cond \doo(\+b_i), \+c_i) &= p_1(\+q\cond \doo(\+b_i, E_1=e_1), \+c_i), \\
    p_2(\+q\cond \doo(\+b_i), \+c_i) &= p_2(\+q\cond \doo(\+b_i, E_1=e_1), \+c_i).
  \end{aligned}
\]
Now, equations~\eqref{eq:descendantsdo} and \eqref{eq:Qequal} assure that 
\[
\begin{aligned}
 p_1(\+q\cond \doo(\+b_i, E_1=e_1), \+c_i) &= p'_1(\+q \cond \doo(\+b_i, T=e_1), \+c_i) \\
&= p'_2(\+q\cond \doo(\+b_i, T=e_1), \+c_i) \\
&= p_2(\+q\cond \doo(\+b_i, E_1=e_1), \+c_i).
\end{aligned}
\]
Similarly, we can use equations~\eqref{eq:descendantsdo} and \eqref{eq:Wequal2} to have

\begin{equation} \label{eq:wequal3}  
\begin{aligned}
 p_1(\+d\cond \doo(\+b_i, E_1=e_1), \+c_i,\+q) &= p'_1(\+d\cond \doo(\+b_i, T=e_1), \+c_i,\+q) \\
&= p'_2(\+d\cond \doo(\+b_i, T=e_1), \+c_i,\+q) \\
&= p_2(\+d\cond \doo(\+b_i, E_1=e_1), \+c_i, \+q).
\end{aligned}
\end{equation} 
For the vertices on the path $R_1 \rightarrow T_1 \rightarrow \dots \rightarrow T_k \rightarrow E_1$, we can write
\begin{equation} \label{eq:pathfactor}
  \begin{aligned}
  &p(r_1,t_1,\ldots,t_k,e_1\cond \doo(\+ b_i), \+ c_i, \+ q) \\
  &\quad = p_1(r_1\cond \doo(\+b_i), \+c_i, \+q)p(t_1 \cond \doo(\+b_i), \+c_i, \+q, r_1)\cdots p(e_1\cond \doo(\+b_i), \+c_i, \+q, t_k).
  \end{aligned}
\end{equation}
In the model construction, the structural equation of $R_1$ is the same as the structural equation of $T$. This, together with equation~\eqref{eq:Tequal} assures that
\begin{equation} \label{eq:r1equal}
  p_1(r_1 \cond \doo(\+b_i), \+c_i, \+q) = p_2(r_1 \cond \doo(\+b_i), \+c_i, \+q).
\end{equation}
Now, per the inheritance function $h$, $T_1$ obtains the value $t$ by either inheriting from $R_1$ or by having it drawn from $p^g(t)$, meaning that 
\[
  p_1(T_1 = t \cond \doo(\+ b_i), \+ c_i, \+q, R_1 = r) = \begin{cases} 
    \gamma + (1 - \gamma) p_1^g(t) & \text{if } r = t, \\
    p_1^g(t) & \text{otherwise}.
  \end{cases}
\]
By equation~\eqref{eq:r1equal}, $p_1(R_1=t\cond \doo(\+b_i), \+c_i, \+q) = p_2(R_1=t\cond \doo(\+b_i), \+c_i, \+q)$. The equality $p_1^g(t) = p_2^g(t)$ is guaranteed by construction. Therefore
\[
  p_1(t_1 \cond \doo(\+b_i), \+c_i, \+q, r_1) = p_2(t_1 \cond \doo(\+b_i), \+c_i, \+q, r_1).
\]
By induction, the equality holds for the remaining factors in equation~\eqref{eq:pathfactor} as well. Finally, the model construction assures that all vertices outside the path $R_1 \rightarrow T_1 \rightarrow \dots \rightarrow T_k \rightarrow E_1$ are Bernoulli distributed with the same probability parameter between the two models. This, together with the equality of all the factors in equation~\eqref{eq:pathfactor} implies that
\[
  p_1(\+t^* \cond \doo(\+b_i), \+c_i, \+q) = p_2(\+t^* \cond \doo(\+b_i), \+c_i, \+q). 
\]
We have shown that all factors in the input distributions remain the same between $\sM_1$ and $\sM_2$, and therefore $p_1(\+a_i \cond \doo(\+b_i), \+c_i)=p_2(\+a_i \cond \doo(\+b_i), \+c_i)$ in case (I).

(II) Let us assume that $T \in \+B_i'$, and denote $\+T^*=\+T\cap\+B_i$ and $\+W=\+B_i'\setminus T$. Now, for $M_1'$ we can write
\begin{equation} \label{eq:bequal}
  \begin{aligned}
      p'_1(\+a_i' \cond \doo(\+b_i'), \+c_i') = p'_1(\+a_i' \cond \doo(\+w, T=t), \+c_i').
  \end{aligned} 
\end{equation}
Similarly, for $M_1$, we have
\begin{align*}
  p_1(\+a_i \cond \doo(\+b_i), \+c_i) = p_1(\+a_i \cond \doo(\+w, E_1=e_1), \+c_i).
\end{align*} 
Using equations~\eqref{eq:descendantsdo} and \eqref{eq:bequal}, we have  
\[
\begin{aligned}
 p_1(\+a_i\cond \doo(\+w_i, E_1=e_1), \+c_i) &= p'_1(\+a_i \cond \doo(\+w, T=e_1), \+c_i) \\
&= p'_2(\+a_i\cond \doo(\+w, T=e_1), \+c_i) \\
&= p_2(\+a_i\cond \doo(\+w, E_1=e_1), \+c_i).
\end{aligned}
\]
  
(III) Let us assume that $T \in \+C_i'$, and denote $\+T^*=\+C_i\cap\+T$ and  $\+Q=\+C_i'\setminus T$. Now, we can write:
\[
  p'_1(\+a_i' \cond \doo(\+b_i'), \+c_i') = p'_1(\+a_i' \cond \doo(\+b_i'), \+q, T=t).
\]
When the condition $(\+A_i'\independent T \cond \+B_i', \+C_i'\setminus T)_{\sG^{'}[\overline{\+B_i'}, \underline{T}]}$ on line~\ref{line:cparents} holds, we can now use the second rule of do-calculus to obtain 
\[
  p'_1(\+a_i' \cond \doo(\+b_i'), \+q, T=t) = p'_1(\+a_i' \cond \doo(\+b_i', T=t), \+q).
\]
From Lemma~\ref{thm:transitcluster_edgeremoval} and Lemma~\ref{thm:transitcluster_dseparation}, it follows that $(\+A_i\independent (\+T\cap\+C_i) \cond \+B_i, \+C_i\setminus \+T)_{\sG_[\overline{\+B_i}, \underline{\+T\cap \+C_i}]}$. For $\sM_1$, we obtain
\[
  p_1(\+a_i \cond \doo(\+b_i), \+c_i) = p_1(\+a_i \cond \doo(\+b_i, T=t), \+q).
\]
We now have the same factorizations for $p'_1(\+a_i' \cond \doo(\+b_i'), \+c_i')$ and $p_1(\+a_i \cond \doo(\+b_i), \+c_i)$ as in part (II). We can therefore use the same steps to conclude
\[
  p_1(\+a_i \cond \doo(\+b_i), \+c_i) = p_2(\+a_i \cond \doo(\+b_i), \+c_i).
\]

(IV) Finally, we assume that $T \notin (\+ A_i' \cup \+ B_i' \cup \+ C_i')$. For $\sM_1'$ we can write
\[
  p_1'(\+a_i'\cond \doo(\+b_i'), \+c_i') = \sum_t p_1'(\+a_i'\cond \doo(\+b_i'), \+c_i', t)p'_1(t\cond \doo(\+b_i'), \+c_i'),
\]
and correspondingly for $\sM_1$
\[
  p_1(\+a_i\cond \doo(\+b_i), \+c_i) = \sum_{e_1} p_1(\+a_i\cond \doo(\+b_i), \+c_i, e_1)p_1(e_1\cond \doo(\+b_i), \+c_i).
\]

As emitter $E_1$ obtains the value $t$ by either inheriting it through the path $R_1 \rightarrow T_1 \rightarrow \cdots \rightarrow T_k \rightarrow E_1$ or by having it drawn by $g$ we can write
\begin{align*}
  p_1(E_1=t \cond \doo(\+ b_i), \+ c_i) &= \gamma^{k+1} p'_1(T=t\cond \doo(\+b_i), \+c_i) + (1 - \gamma^{k+1}) p_1^g(t).
\end{align*}

It then follows
\begin{align*}
  p_1(\+a_i\cond \doo(\+b_i), \+c_i) &= \gamma^{k+1} \sum_{t}p_1(\+a_i\cond \doo(\+b_i), \+c_i, E_1=t) p'_1(T=t\cond \doo(\+b_i), \+c_i) \\
  &\quad + (1 - \gamma^{k+1}) \sum_{t}p_1(\+a_i\cond \doo(\+b_i), \+c_i, E_1 = t) p_1^g(t), 
\end{align*}
In the case of inheritance, which has probability $\gamma^{k+1}$, the value of $E_1$ in $\sM_1$ is the same as the value of $T$ in $\sM'_1$ and $p_1(\+a_1 \cond \doo(\+b_i),\+c_i,E_1=t)=p'_1(\+a_1 \cond \doo(\+b_i),\+c_i,T=t)$.
In the case of non-inheritance, which has probability $1 - \gamma^{k+1}$, the value of $E_1$ is generated either from $p'_1(t)$ or from a distribution over the domain of $T$ that is the same between the models, i.e., the value of $E_1$ is set by a stochastic intervention. This allows us to write  
\begin{align*}
  &p_1(\+a_i\cond \doo(\+b_i), \+c_i) \\
  &\quad = \gamma^{k+1} \sum_{t}p'_1(\+a_i\cond \doo(\+b_i), \+c_i, T=t) p'_1(T=t\cond \doo(\+b_i), \+c_i) \\
  &\qquad + (1 - \gamma^{k+1}) \sum_{t}p'_1(\+a_i\cond \doo(\+b_i, T=t), \+c_i) p_1^g(T=t) \\
  &\quad = \gamma^{k+1} p'_1(\+a_i\cond \doo(\+b_i), \+c_i) +  (1 - \gamma^{k+1})\sum_{t}p'_1(\+a_i\cond \doo(\+b_i, T=t), \+c_i) p_1^g(T=t).
\end{align*}
Now it holds $p'_1(\+a_i\cond \doo(\+b_i), \+c_i)=p'_2(\+a_i\cond \doo(\+b_i), \+c_i)$ but additional conditions are needed to conclude that
\[
	\sum_{t} p'_1(\+a_i\cond \doo(\+b_i, T=t), \+c_i) p^g_1(T=t) = \sum_{t} p'_2(\+a_i\cond \doo(\+b_i, T=t), \+c_i) p^g_2(T=t).
\]
The lines \ref{line:emptybc}--\ref{line:anotherinput} of \checkid{} go through three separate cases that allow us to conclude this equality. 

First, assume that conditions on lines~\ref{line:emptybc} and \ref{line:docalcrule2} hold, i.e., there exists $\sI_j' \in \sI'$ such that $T \subseteq \+A_j'$, $\+B_j' = \emptyset$, and $\+C_j' = \emptyset$, and $(\+A_i' \independent T \cond \+B_i', \+C_i')_{\sG[\overline{\+B_i'}, \underline{T}]}$, $(T \independent \+C_i' \cond \+B_i')_{\sG[ \overline{\+B_i'}]}$ and $(T \independent \+B_i')_{\sG[\overline{\+B_i'}]}$. We can then write
\begin{align*}
	\sum_{t}p'_1(\+a_i \cond \doo(\+b_i, t), \+c_i)p^g_1(t) &= \sum_{t}p'_1(\+a_i \cond \doo(\+b_i), \+c_i, t)p'_1(t\cond \doo(\+b_i), \+c_i) \\ 
  &= p'_1(\+a_i\cond\doo(\+b_i), \+c_i).
\end{align*}
where the equality $p'_1(\+a_i \cond \doo(\+b_i), \+c_i, T=t) = p'_1(\+a_i \cond \doo(\+b_i, T=t), \+c_i)$ follows from the second rule of do-calculus, and the equality $p'_1(t\cond\doo(\+b_i), \+c_i) = p'_1(t\cond\doo(\+b_i)) = p'_1(t) = p^g_1(t)$ follows from the first and the third rule of do-calculus, and from line~\ref{line:emptybc}, which ensures that $\sI'_j = p'(\+ a'_j)$ exists such that $T \in \+ A'_j$. As $p'_1(\+a_i\cond\doo(\+b_i), \+c_i)=p'_2(\+a_i\cond\doo(\+b_i), \+c_i)$, we have 
\[
  \sum_{t}p'_1(\+a_i \cond \doo(\+b_i, T=t), \+c_i)p^g_1(T=t) = \sum_{t}p'_2(\+a_i \cond \doo(\+b_i, T=t), \+c_i)p^g_2(T=t).
\]
Then, assume that the condition of line~\ref{line:docalcrule3} holds, i.e., $(\+A_i' \independent T \cond \+B_i', \+C_i')_{\sG[ \overline{\+B_i'}, \overline{T(\+C_i')}]}$.

By the third rule of do-calculus, we can write
\begin{align*}
  \sum_{t}p'_1(\+a_i\cond \doo(\+b_i, T=t), \+c_i) p^g_1(T=t) &= \sum_{t}p'_1(\+a_i\cond \doo(\+b_i), \+c_i) p^g_1(T=t) \\ 
  &= p'_1(\+a_i\cond \doo(\+b_i), \+c_i)\sum_{t}p^g_1(T=t) \\ 
  &= p'_1(\+a_i\cond \doo(\+b_i), \+c_i).
\end{align*}
As $p'_1(\+a_i\cond \doo(\+b_i), \+c_i)=p'_2(\+a_i\cond \doo(\+b_i), \+c_i)$, we conclude
\[
  \sum_{t}p'_1(\+a_i \cond \doo(\+b_i, T=t), \+c_i)p^g_1(T=t) = \sum_{t}p'_2(\+a_i \cond \doo(\+b_i, T=t), \+c_i)p^g_2(T=t).
\] 
Finally, if the iteration continues, the condition of line \ref{line:anotherinput} is checked. When this condition holds, there exists another input distribution $\sI'_k = p'(\+ a_i \cond \doo(\+ b_i, t))$ which ensures that $p'_1(\+a_i\cond \doo(\+b_i, t), \+c_i) = p'_2(\+a_i\cond \doo(\+b_i, t), \+c_i)$. As $p^g_1(t) = p^g_2(t)$ by the construction, we then obtain directly 
\[
  \sum_{t} p'_1(\+a_i \cond \doo(\+b_i, T=t), \+c_i)p^g_1(T=t) = \sum_{t}p'_2(\+a_i \cond \doo(\+b_i, T=t), \+c_i)p^g_2(T=t).
\] 

Combining the results of parts (I), (II), (III), and (IV), we have shown that $p_1(\+a_i\cond \doo(\+b_i), \+c_i) = p_2(\+a_i\cond \doo(\+b_i), \+c_i)$.
  
\emph{Difference of Causal Effects:} Finally, we show that if $p_1'(\+y\cond \doo(\+x)) \neq p_2'(\+y\cond \doo(\+x))$ for $\sM_1'$ and $\sM_2'$ then the same also holds for $\sM_1$ and $\sM_2$. First, consider the case where members of $\+T \cup \Ch_{\sG}(\+ T)$ are not ancestors of $\+Y$ in $\sG[\overline{\+X}]$. The model construction ensures that $f_i=f_i'$ and $p(\+ U[V_i]) = p'(\+ U[V_i'])$ for all variables $V_i \in \+V \setminus (\+T \cup \Ch_\sG(\+T))$. Therefore, we have that $p_1(\+y\cond\doo(\+x))= p'_1(\+y\cond\doo(\+x))$ and $p_2(\+y\cond\doo(\+x))= p_2'(\+y\cond\doo(\+x))$ and thus $p_1(\+y\cond\doo(\+x)) \neq p_2(\+y\cond\doo(\+x))$. Next, suppose that members of $\+ T \cup \Ch_{\sG}(\+ T)$ are ancestors of $\+Y$ in $\sG[\overline{\+X}]$. Analogously to part (IV) of the proof, we can write
\[
\begin{aligned}
    p_j(\+ y \cond \doo(\+ x)) &= \gamma^{k+1} p'_j(\+ y\cond \doo(\+ x)) +  (1 - \gamma^{k+1})\sum_{t}p'_j(\+ y \cond \doo(\+ x,t)) p^g_j(t),
  \end{aligned}
\]
for $j = 1,2$. We can consider the difference
\[
  \begin{aligned}
    &p_1(\+ y \cond \doo(\+ x)) - p_2(\+ y \cond \doo(\+ x)) \\
    &= \gamma^{k+1}\left[p'_1(\+ y\cond \doo(\+ x)) - p'_2(\+ y\cond \doo(\+ x))\right] \\
    &\quad + (1-\gamma^{k+1})\left[\sum_{t}p'_1(\+ y \cond \doo(\+ x, t)) p^g_1(t) - \sum_{t}p'_2(\+ y \cond \doo(\+ x, t)) p^g_2(t),\right] \\
    &= \gamma^{k+1}\alpha(\+x,\+y) + (1-\gamma^{k+1})\beta(\+x,\+y)
  \end{aligned}
\]
where $\alpha(\+x,\+y)$ and $\beta(\+x,\+y)$ denote the differences under value assignments $\+ x$ to $\+ X$ and $\+ y$ to $\+ Y$, respectively. As the causal effects differ between $\sM'_1$ and $\sM'_2$, we know that $\alpha(\+x,\+y) \neq 0$ for some $\+x$ and $\+y$. Thus we can always choose the value of $\gamma$, regardless the value of $\beta(\+x,\+y)$, so that $\gamma^{k+1}\alpha(\+x,\+y) + (1-\gamma^{k+1})\beta(\+x,\+y) \neq 0$ for at least one pair of value assignments $\+y$ and $\+ x$ because the value of $\gamma$ has no bearing on the other parts of the proof. Thus $p_1(\+y\cond\doo(\+x)) \neq p_2(\+y\cond\doo(\+x))$.
\end{proof}

The leading idea of \checkid{} is to validate properties related to the clustered causal graph and the clustered input distributions that ensure identification invariance with respect to any unclustered graph. Some of these properties are conditional independence constraints derived via d-separation, while others require the existence of another clustered input distribution with specific features. The condition on line~\ref{line:noparents} considers a special case where the cluster has no external parents in the causal graph, from which we can directly conclude identification invariance without any other requirements.

Violations of the conditions checked by \checkid{} may lead to clustering not being identification invariant for the causal effect of interest, as we demonstrate via examples in Appendix~\ref{apx:counterexamples}. Theorem~\ref{thm:invariance_nonid} implies that \checkid{} is sound, that is, if the algorithm returns \textbf{TRUE}, then the clustering operation is identification invariant. The completeness of the algorithm remains unknown. Furthermore, even if the algorithm returns \textbf{FALSE}, verification may still succeed for a subset of the original inputs. We will describe this point in greater detail in Section~\ref{sec:functionals}. The simulation study in Section~\ref{sec:simulation} provides information on how often \checkid{} returns \textbf{TRUE} for random input distributions.

We highlight that Theorems~\ref{thm:invariance_id} and \ref{thm:invariance_nonid} encompass as a special case the identification invariance result for a single data source by \citet{tikka2023clustering}, who showed that if the original input distribution contains the set of emitters and the parents of the transit cluster, then the transit cluster can be replaced by a single vertex without affecting the identifiability of the causal effect. \checkid{} also determines such scenarios as identification invariant, because when $\Em(\+T)\cup(\Pa(\+T)\setminus\+T)\subseteq\+A_i$, all backdoor paths from any vertex of $\+ T$ are blocked by the parents of the cluster. By Lemmas~\ref{thm:transitcluster_edgeremoval} and \ref{thm:transitcluster_dseparation}, the same d-separation also holds for $T$ in the clustered input distributions, and thus the condition of line~\ref{line:aparents} is satisfied. In addition, because $\+C_i'=\emptyset$ for a single observation data source, the condition of line~\ref{line:cdesc1} is satisfied, guaranteeing identification invariance. We note that our results are more general even in the case of a single observational data source than the previous result, as the parents of the cluster need not all be observed.

Theorems~\ref{thm:invariance_id} and \ref{thm:invariance_nonid} and \checkid{} make no assumptions about the internal structure of the transit cluster $\+ T$. However, there are instances where additional information about the transit cluster guarantees identification invariance while only requiring the compatibility of the input distributions. One such scenario is characterized by the following theorem.

\begin{thmrep} \label{thm:singlelayerid}
Let $\sG(\+V\cup\Us)$ be a causal graph where $\+T \subset \+V\cup \Us$ is a transit cluster and let $\sI$ be a set of input distributions compatible with $\+ T$. If $\rec[\sG](\+ T) \cap \emi[\sG](\+ T) \neq \emptyset$, then clustering $(\sG, \sI) \to (\sG', \sI')$, where $\sI' = \sI[\+ T \to T]$ in $\sG' = \sG[\+ T \to T]$, is identification invariant for any causal effect $p(\+y \cond \doo(\+x))$, where $\+X \cup \+Y \subset \+V$, $\+Y \cap \+X = \emptyset$, $\+Y \cap \+T = \emptyset$, $\+X \cap \+T = \emptyset$.
\end{thmrep}
\begin{proof}
Assume first that $p(\+y \cond \doo(\+x))$ is not identifiable from $\sI$ in $\sG$. Thus, there exist two models $\sM_1$ and $\sM_2$ for which the input distributions $\sI_1$ and $\sI_2$ agree but the causal effects differ. We construct $\sM'_1$ and $\sM'_2$ based on these two models. The structural equation for $T$ is defined as $f_T = (f_{T_1},\ldots,f_{T_m})$ in both models, where $\+ T \cap \+ V = \{T_1,\ldots,T_m\}$. In other words, $T$ is defined as a random vector whose components are the members of $\+ T$. This is possible because $\sM_1$ and $\sM_2$ are recursive, and the construction is described in detail in the proof of Theorem~\ref{thm:invariance_nonid}. Likewise, for each variable in $\Ch_\sG(\+ T) \setminus \+ T$, we can simply copy the structural equations from $\sM_1$ and $\sM_2$ such that whenever any member of $\+ T$ appears in the structural equation, the corresponding functions in $\sM_1$ and $\sM_2$ simply select the corresponding component of $T$. The structural equations for the remaining variables in $\+ V \setminus (\+ T \cup \Ch_\sG(\+ T))$ are directly copied from $\sM_1$ and $\sM_2$ into $\sM'_1$ and $\sM'_2$, respectively. This ensures that $\sI_1' = \sI_2'$ as $\sI$ is assumed to be compatible with $\+ T$, but the causal effects differ between $\sM'_1$ and $\sM'_2$.

Assume then that $p'(\+y \cond \doo(\+x))$ is not identifiable from $\sI'$ in $\sG'$. Thus there exist two models $\sM_1'$ and $\sM_2'$ for which the input distributions $\sI'_1$ and $\sI'_2$ agree but the causal effects differ. We construct $\sM_1$ and $\sM_2$ by selecting one variable $T_1$ from the set $\rec[\sG](\+ T) \cap \emi[\sG](\+ T)$ to be defined exactly as $T$ is for $\sM_1'$ and $\sM_2'$. In other words $f_{T_1} = f_{T}$ and $\+U'[T_1] = \+U[T]$. This is possible, because $T_1$ is a receiver of $\+ T$. The other variables in $\+ T \setminus T_1$ are defined as Bernoulli distributed with the same probability parameter and as mutually independent. Finally, every other variable in $\sM_1$ and $\sM_2$ is defined as in $\sM_1'$ and $\sM_2'$, apart from $\Ch_\sG(\+T) \setminus \+ T$, for which $T$ is replaced by $T_1$ in their corresponding structural equations. This is possible, because $T_1$ is an emitter of $\+ T$. Therefore, $\+U = \+U'\cup \+U[\+T\setminus T_1]$, where $\+U'$ denotes the error terms for the clustered model and $\+U[\+T\setminus T_1]$ are the Bernoulli-distributed error terms. Now $\sM_1$ and $\sM'_1$ are equivalent, as are $\sM_2$ and $\sM'_2$. As variables $\+T \setminus T_1$ and $\+U[\+T \setminus T_1]$ are defined similarly between $\sM_1$ and $\sM_2$, and they have no effect on the remaining variables, it follows that $\sI_1 = \sI_2$ but $p_1(\+y \cond \doo(\+x)) \neq p_2(\+y \cond \doo(\+x))$.
\end{proof}

As an example, the transit cluster $\+T=\{T_1, T_2, T_3\}$ of Figure~\ref{fig:flat1} satisfies the conditions of Theorem~\ref{thm:singlelayerid}, because the vertex $T_1$ serves both as a receiver and as an emitter. In this case, clustering $\+T$ is identification invariant for the causal effect $p(y\cond\doo(x))$, and its identifiability can be assessed based on the clustered graph of Figure~\ref{fig:flat2} for all combinations of possible input distributions that are compatible with $\+T$.  

\begin{figure}[!ht]
  \begin{center}
    \begin{subfigure}{0.475\textwidth}
      \centering
      \begin{tikzpicture}[scale=2.5]
        \node [dot5 = {0}{0}{Y}{right}{Y}] at (1,0.5) {};
        \node [dot5 = {0}{0}{T_2}{above}{T_2}] at (-0.33,0.5) {};
        \node [dot5 = {0}{0}{T_3}{above}{T_3}] at (0.33,0.5) {};
        \node [dot5 = {0}{0}{T_1}{below}{T_1}] at (0, 0.15) {};
        \node [dot5 = {0}{0}{X}{left}{X}] at (-1,0.5) {};
        \draw [->] (X) -- (T_1);
        \draw [->] (X) -- (T_2);
        \draw [->] (T_1) -- (T_3);
        \draw [->] (T_2) -- (T_3);
        \draw [->] (T_1) -- (Y);
        \draw [->] (T_3) -- (Y);
        \draw [<->, dashed] (X) to [bend left=45] (Y); 
      \end{tikzpicture}
      \caption{}
      \label{fig:flat1}
    \end{subfigure}
    \begin{subfigure}{0.475\textwidth}
      \centering
      \begin{tikzpicture}[scale=2.5]
        \node [dot5 = {0}{0}{Y}{right}{Y}] at (1,0.5) {};
        \node [dot5 = {0}{0}{T}{below}{T}] at (0,0.5) {};
        \node [dot5 = {0}{0}{X}{left}{X}] at (-1,0.5) {};
        \node at (-1, 0) {\vphantom{$T_1$}};
        \draw [->] (X) -- (T);
        \draw [->] (T) -- (Y);
        \draw [<->, dashed] (X) to [bend left=45] (Y); 
      \end{tikzpicture}
      \caption{}
      \label{fig:flat2}
    \end{subfigure}
  \end{center}
  \caption{An example where the transit cluster $\{T_1, T_2, T_3\}$ in panel~(a) can be clustered to a single vertex $T$ as depicted in panel~(b) without any additional information on the input distributions when considering the identification invariance of clustering regarding }the causal effect $p(y\cond\doo(x))$.
  \label{fig:flatexample}
\end{figure}

Next, we present an example on the use of Theorems~\ref{thm:invariance_id} and \ref{thm:invariance_nonid}. Consider again the graph of Figure~\ref{fig:clustg}, for which we detected the transit cluster $\+S = \{R, S_1, S_2, E_1, E_2\}$ and the transit cluster $\+T = \{T_1, T_2\}$. We consider four different combinations of input distributions to identify the causal effect $p(y\cond\doo(x))$:
\begin{enumerate}[label=(\roman*)]
  \item $p(x, e_1, e_2, s_1, r)$ and $p(y, e_1, e_2, t_1, t_2)$,
  \item $p(x, e_1, e_2, r, w_1)$ and $p(y, w_2\cond \doo(w_1))$,
  \item $p(y \cond \doo(t_1, t_2), e_1, e_2, s_2)$ and $p(x, t_1, t_2)$,
  \item $p(y, t_1, t_2, r, e_1, e_2)$, $p(x, w_1 \cond \doo(e_1, e_2))$ and $p(x, w_1)$.
\end{enumerate}
First, we can notice that all sets of input distributions are compatible with respect to $\+S$ and $\+T$. This applies because $\Em_\sG(\+S)=\{E_1, E_2\}$ and $\Em_\sG(\+T)=\{T_1, T_2\}$ are always entirely present in an input distribution if any vertex within the cluster is present in the input distribution. The corresponding clustered input distributions for each case are:
\begin{enumerate}[label=(\roman*)]
  \item $p(x, s)$ and $p(y, s, t)$,
  \item $p(x, s, w_1)$ and $p(y, w_2\cond \doo(w_1))$,
  \item $p(y \cond \doo(t), s)$ and $p(x, t)$,
  \item $p(y, t, s)$, $p(x, w_1 \cond \doo(s))$ and $p(x, w_1)$.
\end{enumerate}

In case (i) the causal effect $p(y\cond\doo(x))$ is identifiable in the clustered graph. This means that according to Theorem~\ref{thm:invariance_id}, the causal effect $p(y\cond\doo(x))$ is also identifiable in the original graph. We can therefore use the clustered graph and input distributions to obtain the identifying functional
\[
p(y\cond \doo(x))=\sum_{s,t}p(s|x)p(t)p(y|s,t) ,
\]
which also provides us with an identifying functional for the causal effect in $\sG$ as we will show in Section~\ref{sec:functionals}. 

In cases (ii)--(iv), the causal effect $p(y\cond\doo(x))$ is not identifiable in the clustered graph. To assess whether the identification invariance holds in these cases, by Theorem~\ref{thm:invariance_nonid}, the conditions regarding the input distributions must be checked by \checkid{}. 

We can first see that $T$ does not have any incoming edges in the clustered graph, which means that \checkid{} returns TRUE in all cases per line~\ref{line:noparents}. Clustering $\+T$ is therefore identification invariant for the causal effect $p(y \cond \doo(x))$. Next, we will show that the clustering is also identification invariant with respect to the transit cluster $\+S$ by going through the remaining conditions of \checkid{} for the remaining clustered input distributions. 

In case (ii), for the input distribution $p(x, s, w_1)$, vertex $X$ blocks all backdoor paths between $S$ and its descendants, which satisfies the condition of line~\ref{line:aparents}. In addition, $\+C_1'=\emptyset$ and therefore the condition for FALSE on line~\ref{line:cdesc1} is not triggered. In the first input distribution $S \in \+A_1'$, and $\+B_1'$ and $\+C_1'$ are empty, allowing us to proceed to the line~\ref{line:docalcrule2}. Next, the input distribution $p(y, w_2 \cond \doo(w_1))$ does not contain any $S \in\+S$. We can see that $(S \independent Y, W_2 \cond W_1)_{\sG'[\overline{W_1}, \underline{S}]}$. In addition $(S\independent W_1)_{\sG'[\overline{W_1}]}$, and thus all conditions of line~\ref{line:docalcrule2} are satisfied. We can conclude that in case (ii) clustering $\+S$ is identification invariant for $p(y\cond\doo(x))$.

In case (iii), for the input distribution $p(y \cond \doo(t), s)$, it applies that $S \in \+C_2'$. Now $T \in \+B_2'$ blocks all backdoor paths between $S$ and $\+A_2'=\{Y\}$, satisfying the condition of line~\ref{line:cparents}. In addition, $\+C_2'$ does not contain any descendants of $S$, and therefore all conditions for this input distribution are satisfied. Finally, we can see that no member of $\+ S$ is present in the clustered input distribution $p(x, t)$. The condition on line~\ref{line:docalcrule2} is not satisfied but as $\+A_3'$ contains only non-descendants of $S$ the condition $(\+A_3' \independent S)_{\sG'[\overline{S}]}$ on line~\ref{line:docalcrule3} is fulfilled. All input distributions satisfy the conditions of \checkid{}, and therefore clustering $\+S$ is identification invariant for the causal effect $p(y\cond\doo(x))$. 

In case (iv), $T$ blocks the backdoor paths between the cluster vertices and $Y$, and $\+C_1'=\emptyset$, meaning that lines~\ref{line:cdesc1} and \ref{line:aparents} are satisfied for the first clustered input distribution $p(y, t, s)$. For the second input distribution $p(x, w_1 \cond \doo(s))$, it applies that $T \in \+B_2'$. In this case the condition of line~\ref{line:bparents} applies directly. Finally, we can see that no cluster vertex is present in $p(x, w_1)$. Now, we see that the second distribution, $p(x, w_1 \cond \doo(t))$, has the same variables, but with $T$ inside the do-operator. This satisfies the condition of line~\ref{line:anotherinput} for the third input distribution. Therefore, also in case (iv), clustering is identification invariant with respect to $\+S$. 

In all cases (ii)--(iv) we were able to establish the identification invariance. Using Theorem~\ref{thm:invariance_nonid}, we can conclude that the causal effect $p(y\cond\doo(x))$ is not identifiable in the graph where $\+S$ is not clustered either.

\section{Obtaining Identifying Functionals} \label{sec:functionals}

\begin{toappendix}
  \label{apx:functionalproof}
\end{toappendix}

Theorems~\ref{thm:Yancestors}, \ref{thm:Xancestors}, \ref{thm:isolated}, \ref{thm:invariance_id}, \ref{thm:invariance_nonid} and \ref{thm:singlelayerid} show that identifiability is neither lost nor gained by pruning or clustering under specific assumptions. Except for Theorem~\ref{thm:Yancestors}, these results do not provide a way to obtain an identifying functional for the causal effect in the original graph. Intuitively, pruning should not alter the identifying functional, i.e., an identifying functional obtained in the pruned graph should be applicable in the original. Similarly for clustering, we should simply be able to replace the clustered input distributions by their original counterparts in the identifying functional. Fortunately, both of these intuitive notions apply to a large class of identifying functionals, as we will demonstrate with the following two theorems for pruning and clustering, respectively. The proofs of this section are presented in Appendix~\ref{apx:functionalproof}.

\begin{thmrep}[Identifying functional after pruning] \label{thm:pruningfunctional}
Let $\sG(\+ V \cup \+ U)$ be a causal graph and let $\sI$ be a set of input distributions. Let $\sG'$ and $\sI'$ be the causal graph and the set of input distributions obtained respectively by applying Theorem~\ref{thm:Yancestors}, \ref{thm:Xancestors}, or \ref{thm:isolated} to $\sG$ and $\sI$. If $p'(\+ y \cond \doo(\+ x))$ is identifiable from $\sI'$ in $\sG'$ by do-calculus with the identifying functional $g(\sI')$, then $p(\+ y \cond \doo(\+ x))$ is identifiable from $\sI$ in $\sG$ by do-calculus with the identifying functional $f(\sI)$ where $f(\sI) = g(\sI')$.
\end{thmrep}
\begin{proof}
For Theorem~\ref{thm:Yancestors}, the claim follows directly from its proof. For Theorems~\ref{thm:Xancestors} and \ref{thm:isolated}, we investigate all possible paths not only in $\sG$ and  $\sG'$ but also in all possible edge subgraphs. This ensures that all d-separation conditions needed in any do-calculus derivation are valid. Let $\sG_*$ be a graph obtained from $\sG$ by removing edges $\+E_*$ between vertices in $\+V$ and let $\sG'_*$ be a graph obtained from $\sG'$ by removing all edges in $\+E_*$ that are in $\sG'$. The set $\+E_*$ may be empty. Let $V_1,V_2 \in \+V'$ be arbitrarily chosen variables, and let $\+W \subset \+V$ be an arbitrarily chosen set of variables for which $V_1,V_2 \notin \+W$. The set $\+W$ may be empty. We aim to show that $V_1$ and $V_2$ are d-separated in $\sG_*$ by $\+W$ if and only if they are d-separated in $\sG'_*$ by $\+W \cap \+V'$. This result would allow us to conclude that operations licensed by the rules of do-calculus are equivalent for $p(\+y \cond \doo(\+x))$ and $p'(\+y \cond \doo(\+x))$ in $\sG$ and $\sG'$, respectively.

It is assumed in Theorems~\ref{thm:Xancestors} and \ref{thm:isolated} that the graph does not contain non-ancestors of $\+ Y$. Let $\+Z$ be the pruned set of variables. We choose arbitrary $V_1,V_2 \in \+ V'$ and consider a path between $V_1$ and $V_2$ in $\sG_*$. If the path does not contain members of $\+Z$, the vertices of the path remain unchanged after pruning $\+Z$. The definitions of $\+Z$ in Theorems~\ref{thm:Xancestors} and \ref{thm:isolated} guarantee that members of $\+Z$ cannot be descendants of the vertices on the path, and thus conditioning on $\+Z$ or its subset does not affect whether the path is open or not. In the case of Theorem~\ref{thm:isolated}, the path cannot contain a member of $\+Z$ because the isolating vertex (denoted by $W$ in Theorem~\ref{thm:isolated}) can be on the path only once, and $V_1$ and $V_2$ do not belong to $\+Z$. We conclude that in the case of Theorem~\ref{thm:isolated}, the d-separation of $V_1$ and $V_2$ does not depend whether members of $\+Z$ are conditioned on or not.

In the case of Theorem~\ref{thm:Xancestors}, the path may contain a member of $\+Z$, which implies that the path must also contain two members of $\+X$. This follows from the definition of $\+ Z$, condition (a) of Theorem~\ref{thm:Xancestors}, and the fact that $V_1$ and $V_2$ are ancestors of $\+Y$. Let $X_1,X_2 \in \+X$ be such vertices on the path such that there is an incoming edge from $\+Z$ to $X_1$ and from $\+Z$ to $X_2$. The definition of $\+ Z$ and the fact that $\+Z$ is an ancestor of $\+Y$ guarantees that such $X_1$ and $X_2$ exist. Further, for the same reason, there must be a member of $\+Z$ that is an ancestor of $X_1$ and $X_2$. Now by the condition (c) of Theorem~\ref{thm:Xancestors}, there exists $U \in \+U$ such that the structure $X_1 \leftarrow U \rightarrow X_2$ exists in the graph. This means that there is another path between $V_1$ and $V_2$ that does not contain members of $\+Z$. Thus, in the case of Theorem~\ref{thm:Xancestors}, the d-separation of $V_1$ and $V_2$ does not depend on whether $V_1$ and $V_2$ are conditioned on members of $\+Z$ or not. 

As $V_1$, $V_2$, and $\+W$ were chosen arbitrarily, the same conclusions hold for all choices. Further, the identifying functional $f(\sI)$ can be derived by the same sequence of do-calculus rules that was used to derive the identifying functional $g(\sI')$. As the distributions of the variables that are present in the input distributions in $\sI'$ remain unchanged after pruning, it holds that $f(\sI) = g(\sI')$. 
\end{proof}

\begin{thmrep}[Identifying functional after clustering] \label{thm:clusteringfunctional}
Let $\+ T$ be a transit cluster in a causal graph $\sG(\+V\cup\Us)$ and let $\sI$ be a set of input distributions compatible with $\+ T$. If $p'(\+ y \cond \doo(\+ x))$ is identifiable from $\sI' = \sI[\+ T \to T]$ in $\sG' = \sG[\+ T \to T]$ by do-calculus with the identifying functional $g(\sI')$, then $p(\+ y \cond \doo(\+ x))$ is identifiable from $\sI$ in $\sG$ by do-calculus with the identifying functional $f(\sI)$ obtained from $g(\sI')$ by replacing every instance of $T$ by the subset of $\+T$ that is present in the corresponding input distribution in $\sI$.
\end{thmrep}
\begin{proof}
When $p'(\+ y \cond \doo(\+ x))$ is identifiable by do-calculus, there exists a sequence of operations licensed by the rules of do-calculus and probability calculus that derive the identifying functional $g(\sI')$ from the input distributions $\sI'$ in $\sG'$. The main idea of the proof is to show that the same sequence of operations can be used to derive identifying functional $f(\sI)$ for $p(\+ y \cond \doo(\+ x))$ in $\sG$. To do this, we have to show that the d-separation conditions required by the rules of do-calculus in $\sG'$ are fulfilled also when the sequence of operations is applied to derive $f(\sI)$ in $\sG$.

The rules of do-calculus operate with graphs where the incoming edges of a set of vertices and the outgoing edges of another set of vertices may have been removed. We investigate all possible paths not only in $\sG$ and $\sG'$ but also in all possible subgraphs that may occur in a do-calculus derivation. This ensures that all d-separation conditions needed in a do-calculus derivation are valid. Let $\+Z'_I,\+Z'_O \subset \+V'$ be disjoint, possibly empty, sets of vertices in $\sG'$ and let $\+Z_I$ and $\+Z_O$ be sets obtained from $\+Z'_I$ and $\+Z'_O$, respectively, by replacing $T$ by $\+T$. We define $\sG'_*=\sG'[\overline{\+ Z'_I},\underline{\+ Z'_O}]$ and $\sG_*=\sG[\overline{\+ Z_I},\underline{\+ Z_O}]$.

By Lemma~\ref{thm:transitcluster_edgeremoval}, a transit cluster $\+T$ in $\sG$ is also a transit cluster in $\sG_*$. Applying Lemma~\ref{thm:transitcluster_dseparation} we conclude that if two sets are d-separated in $\sG'_*$, then they are d-separated also in $\sG_*$. It follows that the sequence of do-calculus operations used to derive identifying functional $g(\sI')$ is applicable in $\sG$ when $T$ is replaced by $\emi_\sG(\+T)$. The identifying functional $f(\sI)$ is derived from $g(\sI')$ by replacing all input distributions in $\sI'$ by the corresponding input distributions in $\sI$ and replacing for all $i=1,\ldots,n$ the instance of $T$ in the input distribution $i$ by the subset of $\+T$ that is present in the input distribution $i$ in $\sI$. Due to compatibility of $\sI$, all these subsets include $\emi_\sG(\+T)$. We conclude that $p(\+ y \cond \doo(\+ x))$ is identifiable from $\sI$ with identifying functional $f(\sI)$.
\end{proof}

In the case of a single observational data source, Theorems~\ref{thm:pruningfunctional} and \ref{thm:clusteringfunctional} apply to all identifying functionals due to the completeness of do-calculus \citep{shpitser2006, huang2006}. Do-calculus has also been shown to be complete for specific constrained instances of the general causal effect identifiability problem, such as $g$-identifiability, where the set of input distributions $\sI$ is of the form $\{ p(\+ v \cond \doo(\+ z_i)) \}_{i=1}^n$ for a collection of subsets $\{\+ Z_i\}_{i=1}^n$ of $\+ V$ \citep{kivva2022revisiting}. Note that in the $g$-identifiability problem, the set of inputs can also include $p(\+ v)$ if $\+ Z_i = \emptyset$ for some $i$, but excludes several other types of input distributions such as conditional distributions, conditional interventional distributions, and joint distributions of subsets of variables. It is not known whether do-calculus is complete for the unconstrained case of the general causal effect identifiability problem.

We note that it is sometimes beneficial to exclude specific inputs from consideration if only a subset of inputs is necessary for identification. Suppose that we have a set of input distributions $\sI$ and a subset $\sJ \subset \sI$ such that Theorem~\ref{thm:pruningfunctional} or Theorem~\ref{thm:clusteringfunctional} applies to $\sJ$ but not $\sI$. If a causal effect of interest is identifiable in the pruned or clustered graph from $\sJ'$, where $\sJ'$ is the corresponding set of restricted input distributions or clustered input distributions, we can readily apply the relevant theorem to $\sJ$ instead to obtain the identifying functional. For instance, consider a simple graph $X \rightarrow Y \rightarrow Z$, with input distributions $\{p(x, y), p(x\cond z)\}$. As the causal effect $p(y\cond \doo(x))$ is identifiable from $p(x, y)$ alone in this graph, we can remove $Z$ by Theorem~\ref{thm:Yancestors}, even though the input $p(x\cond z)$ does not satisfy the conditions by applying the theorem with $p(x, y)$ as the sole input distribution.
We note that the number of possible subsets $\sJ$ is $2^{|\sI|}$, meaning that finding a suitable subset may be computationally challenging when the number of input distributions is large. However, it is often not necessary to check all subsets for the following reasons. In the case of pruning, the conditions in Theorems~\ref{thm:Yancestors}--\ref{thm:isolated} can be checked independently for each input distribution. Similarly in the case of clustering, the validity of a specific clustered input distribution $p(\+ a'_i \cond \doo(\+ b'_i), \+ c'_i)$ is independent of other clustered inputs whenever $T \in \+ A'_i \cup \+ B'_i \cup \+ C'_i$ as can be observed from the conditions checked by Algorithm~\ref{alg:idinvariant} in the if-block starting on line~\ref{line:abc}.

\section{Simulations} \label{sec:simulation}

We performed a simulation study to assess the practical computational impact of the presented graph reduction operations for causal effect identification. We studied the impact of pruning and clustering both together and separately. In this section, our main focus is on the simulation performed for pruning and clustering together; if we discuss the simulations for pruning and clustering separately, we mention it explicitly. The details and the results regarding the simulations for pruning and clustering separately are presented in Appendix~\ref{apx:algorithm}.

We started by generating random causal graphs and random sets of input distributions where each input distribution is of the form $p(\+a_i \cond \doo(\+b_i), \+c_i)$ with sets $\+ B_i$ and $\+ C_i$ potentially being empty. In addition, the graphs were generated such that each graph contained at least one transit cluster and one non-ancestor of the response. To our knowledge, the only general-purpose algorithm able to handle such inputs in causal effect identification is Do-search, implemented in the R package \texttt{dosearch} \citep{tikka2021dosearch, dosearch_package}. We compare two strategies

\begin{itemize}
  \item \textbf{Direct strategy:} Apply Do-search directly in the original graph and record the time elapsed.
  \item \textbf{Reduction strategy:} 
  \begin{enumerate}
    \item Prune the original graph according to the Theorems~\ref{thm:Yancestors}--\ref{thm:isolated} if the respective conditions of each theorem are satisfied. For the sake of the comparison, pruning is omitted if the set to be pruned intersects with the transit cluster. 
  	\item Find the transit clusters of the (possibly) pruned graph and determine the identifiability of the causal effect in the pruned and clustered graph by applying Do-search.
  	\item If the causal effect is identifiable in the reduced graph, it is also identifiable in the original graph by Theorem~\ref{thm:invariance_id}. Record the time elapsed so far and stop (Setting~A).
  	\item Otherwise, use \checkid{} (Algorithm~\ref{alg:idinvariant}) to determine the identification invariance using the pruned and clustered graph, and pruned and clustered input distributions. 
  	\item If the identification invariance holds, the causal effect is not identifiable in the original graph by Theorem~\ref{thm:invariance_nonid}. Record the time elapsed so far and stop (Setting~B). 
    \item If the graph size was reduced by pruning in step 1, apply Do-search in the pruned graph, record the time elapsed so far and stop (Setting~D; see Appendix~\ref{apx:algorithm} for the results on this setting).
  	\item Otherwise, apply Do-search in the original graph, record the time elapsed so far and stop (Setting~C).
  \end{enumerate}
\end{itemize}

Ideally, we expect benefits from the reduction strategy in the cases of settings A (the causal effect is identifiable in the clustered graph) and B (the causal effect is not identifiable in the clustered graph and clustering is identification invariant), compared to applying Do-search directly in the original graph. Additionally, in the worst case of setting C, where the graph reduction does not help in identification, we expect that assessing the graphical conditions to verify the identification invariance does not induce a major overhead. To avoid extensive running times, we imposed a time limit of 15 minutes for Do-search. The generation of one simulation instance is defined in Algorithm~\ref{alg:times} in Appendix~\ref{apx:algorithm}. 

\begin{figure}[!ht]
  \centering
  \begin{subfigure}{0.40\textwidth}
    \centering
    \includegraphics[width=\linewidth]{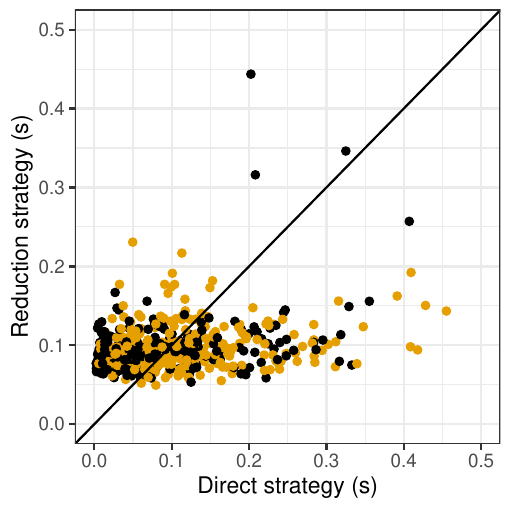}
    \caption{}
    \label{fig:scatter6}
  \end{subfigure}
  \begin{subfigure}{0.40\textwidth}
    \centering
    \includegraphics[width=\linewidth]{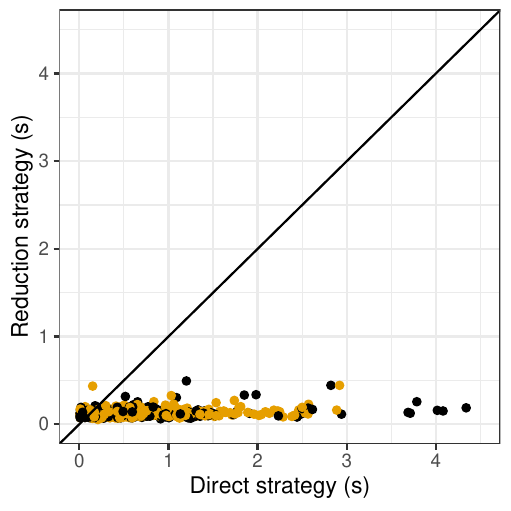}
    \caption{}
    \label{fig:scatter8}
  \end{subfigure}
  \begin{subfigure}{0.4\textwidth}
    \centering
    \includegraphics[width=\linewidth]{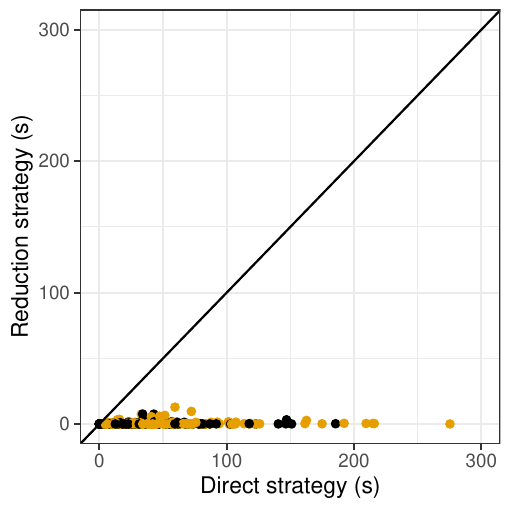}
    \caption{}
    \label{fig:scatter10}
  \end{subfigure}
  \begin{subfigure}{0.4\textwidth}
    \centering
    \includegraphics[width=\linewidth]{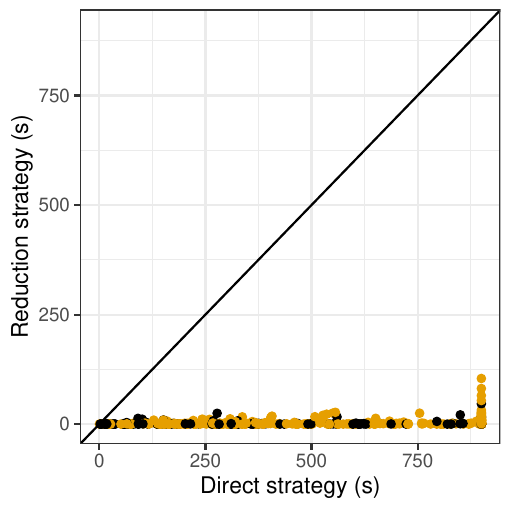}
    \caption{}
    \label{fig:scatter12}
  \end{subfigure}
  \caption{Comparison of the direct and the reduction strategies when reductions are possible. The scatter plots show the running times for  both strategies in settings A (black) and B (orange). The panels depict different original graph sizes: 7 \subref{fig:scatter6}, 8 \subref{fig:scatter8}, 10 \subref{fig:scatter10} and 12 \subref{fig:scatter12}. A random sample of 500 instances is used for each panel. In Panel (d), the heap of points at 900 seconds is due to the time limit of 15 minutes.}
  \label{fig:scatter}
\end{figure}

\begin{figure}
    \centering
    \includegraphics[width=0.95\linewidth]{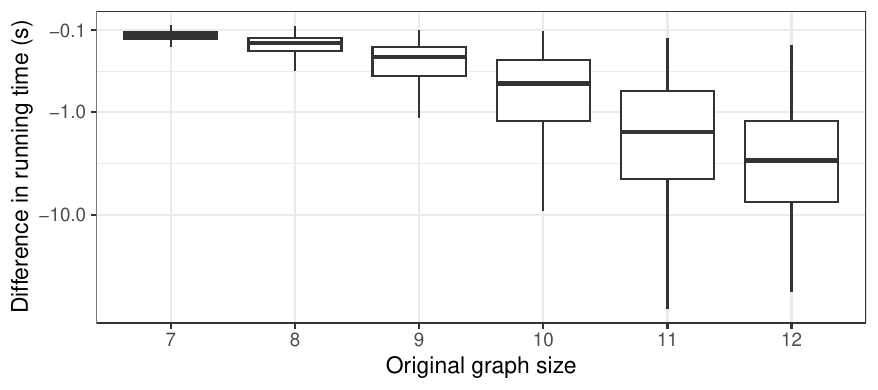}
    \caption{Overhead of the reduction strategy when no reductions are possible. The box plots show the differences in running times between the direct strategy and the reduction strategy for setting C by the size of the original graph. Negative differences indicate that more time was spent with the reduction. The vertical axis uses logarithmic scaling.}
    \label{fig:box}
\end{figure}

In total, we obtained 108~933 instances. 6.6\% of the instances exceeded the time limit of 15 minutes when using Do-search in the original graph. Table~\ref{tab:simresults} reports the differences and ratios of the running times between the direct strategy and the reduction strategy for settings~A--C. For running times that exceeded 15 minutes, we applied a conservative approach and fixed these times to 15 minutes. Additionally, we visualized the results for settings A--B with scatter plots in Figure~\ref{fig:scatter}, and for setting C with a box plot in Figure~\ref{fig:box}.

When identification with clustering was applicable (Settings A--B), the benefits of the graph reduction increased considerably as the original graph sizes increased. For instance, when the original graph size was 12, the median difference was over 12 minutes to the benefit of the reduction strategy in the case of setting B. In contrast, for small graphs with seven vertices, the reduction strategy performed worse for setting A. However, in this case the median time lost was less than a tenth of a second. In the case of setting C, where the original graph must be used for causal effect identification after checking if reduction is applicable, it is naturally faster to use the original graph in identification directly. Reassuringly, the median time lost for checking whether graph reduction is applicable was less than five seconds for all graph sizes, with the ratio approaching one as the graph size increases.

\checkid{} returned \textbf{TRUE} for 52.0\% of the instances that reached step 4 of the reduction strategy, which indicates that the conditions of the algorithm are not too restrictive for practical use. Additionally, it appears that the benefits of the clustering operations were higher for setting B than for setting A. This is likely explained by the search-based nature of Do-search. When the causal effect is identifiable (as is the case for setting A), Do-search does not have to explore its entire search space, unlike in setting B. Therefore, reduction of the search space by clustering benefits setting B more than setting A.

\begin{table}[ht]
\centering
\begin{tabular}{c c r l r l r}
\toprule
Graph size & Setting & \multicolumn{2}{c}{Median difference (IQR)} & \multicolumn{2}{c}{Median ratio (IQR)} & $N$ \\
\midrule
7 & A & $-0.03$ & $(-0.07,\ 0.03)$ & $0.66$ & $(0.26,\ 1.28)$ & 918 \\
7 & B & $0.00$ & $(-0.04,\ 0.06)$ & $1.00$ & $(0.56,\ 1.62)$ & 875 \\
7 & C & $-0.12$ & $(-0.14,\ -0.11)$ & $0.40$ & $(0.30,\ 0.54)$ & 122 \\
8 & A & $0.20$ & $(-0.02,\ 0.60)$ & $2.57$ & $(0.85,\ 5.63)$ & 3 751 \\
8 & B & $0.50$ & $( 0.19,\ 1.00)$ & $4.73$ & $(2.50,\ 8.80)$ & 3 830 \\
8 & C & $-0.17$ & $(-0.22,\ -0.14)$ & $0.72$ & $(0.61,\ 0.81)$ & 674 \\
9 & A & $1.52$ & $( 0.34,\ 3.78)$ & $9.72$ & $(3.15,\ 23.46)$ & 6 875 \\
9 & B & $3.54$ & $( 1.57,\ 7.00)$ & $20.98$ & $(9.45,\ 41.81)$ & 8 059 \\
9 & C & $-0.26$ & $(-0.42,\ -0.19)$ & $0.89$ & $(0.82,\ 0.95)$ & 1 895 \\
10 & A & $9.54$ & $( 2.68,\ 24.11)$ & $36.60$ & $(11.94,\ 99.00)$ & 7 782 \\
10 & B & $21.65$ & $( 9.89,\ 45.57)$ & $75.46$ & $(32.12,\ 184.04)$ & 10 950 \\
10 & C & $-0.51$ & $(-1.21,\ -0.28)$ & $0.96$ & $(0.91,\ 0.98)$ & 3 262 \\
11 & A & $49.07$ & $( 14.72,\ 134.37)$ & $109.78$ & $(30.70,\ 338.38)$ & 5 912 \\
11 & B & $122.88$ & $( 54.19,\ 268.86)$ & $241.27$ & $(83.99,\ 603.91)$ & 10 592 \\
11 & C & $-1.57$ & $(-4.54,\ -0.61)$ & $0.98$ & $(0.95,\ 0.99)$ & 3 852 \\
12 & A & $278.47$ & $( 72.58,\ 736.50)$ & $272.05$ & $(69.12,\ 888.89)$ & 3 453 \\
12 & B & $725.57$ & $( 302.57,\ 898.71)$ & $584.86$ & $(150.68,\ 1517.28)$ & 7 418 \\
12 & C & $-4.28$ & $(-11.42,\ -1.64)$ & $0.99$ & $(0.97,\ 0.99)$ & 3 505 \\
\bottomrule
\end{tabular}
\caption{Comparison of the direct and the reduction strategies. Median differences (in seconds) of running times between the strategies, and median ratios (the running time for the direct strategy divided by the running time of the reduction strategy) are reported. The interquartile ranges (IQR) are included in parentheses. The column $N$ tells the number of instances for each graph size and setting (A: identifiable in the clustered graph, B: non-identifiable in the clustered graph but identification invariant, C: no graph reduction operation was identification invariant, original graph had to be used).}
\label{tab:simresults}
\end{table}

Overall, the results show that when graph reduction is applicable, it can offer significant performance benefits in general causal effect identification. Additionally, in the cases where the graph reduction does not offer improved running time performance, the disadvantage of using the operations is minimal in practice. Across all instances, the median time spent finding transit clusters was 0.13 seconds (IQR 0.10 to 0.17), median time for checking the identification invariance was 0.004 seconds (IQR 0.001 to 0.010) and the median time for performing the pruning operations was 0.024 seconds (IQR 0.017 to 0.038). We also highlight that graph reduction does not have to be automatically applied if the researcher views it as redundant---as can be the case with small graphs or small cluster sizes---for example.

\section{Demonstrations}\label{sec:applications}

To show how the results of this paper can be used in practice, we consider two example cases that are selected from the review by \citet{tennant2021use}. For the first example, we use the causal graph of the study on infant mortality by \citet{tobacco}. As the second example, we consider the ELSA-Brasil study  \citep{elsabrazil} on the effect of life-course socioeconomic position on a marker of atherosclerosis. The focus in the first example is on pruning while the second example focuses on clustering. 

\subsection{Causal Model from an Infant Mortality Study} \label{subsec:pruningdemo}

\citet{tobacco} studied the associations and potential causal pathways of cigarette prices and infant mortality in 23 countries of the European Union. To select an appropriate set of covariates for the analyses, \cite{tobacco} used the causal graph depicted in Figure~\ref{fig:prune1}. To demonstrate the uses of pruning in the case of multiple data sources, we focus on various potential causal effects that can be assessed based on Figure~\ref{fig:prune1}. The variables represented by the causal graph are presented in Table~\ref{tab:pruningvariables}. For more detailed definitions of these variables, see \citep{tobacco}.

\begin{table}[h]
    \centering
    \begin{tabular}{lc}
        \toprule
        Variable & Symbol \\
        \midrule
        GDP unemployment rate & $W$ \\
        Health expenditure & $H$ \\
        Access to healthcare & $A$ \\
        Tobacco control policies & $O$ \\
        Smoking during pregnancy & $D$ \\
        Congenital anomalies & $N$ \\
        Tobacco price & $R$ \\
        Cigarette consumption & $C$ \\
        Second-hand smoking in pregnancy & $S$ \\
        Pre-term birth & $B$ \\
        Education & $E$ \\
        Maternal age & $M$ \\
        Small-for-gestational-age infant & $G$ \\
        Social and cultural factors & $F$ \\
        Ethnicity & $Q$ \\
        Second-hand smoking in infancy & $J$ \\
        Infant mortality & $I$ \\
        \bottomrule
    \end{tabular}
    \caption{Variables in the DAG of \citep{tobacco} and their corresponding symbols in Figure~\ref{fig:prune1}.}
    \label{tab:pruningvariables}
\end{table}

\begin{figure}[!ht]
  \begin{center}
    \begin{subfigure}{0.575\textwidth}
      \centering
      \begin{tikzpicture}[scale=1.5]
        \node [dot5 = {0}{0}{D}{below}{D}] at (0,-1) {};
        \node [dot5 = {0}{0}{H}{below}{H}] at (0,-2.5) {};
        \node [dot5 = {0}{0}{A}{below}{A}] at (2,-2.5) {};
        \node [dot5 = {0}{0}{J}{above}{J}] at (0,1) {};
        \node [dot5 = {0}{0}{O}{above}{O}] at (-1,1) {};
        \node [dot5 = {0}{0}{M}{above left}{M}] at (-1,-1) {};
        \node [dot5 = {0}{0}{G}{below right}{G}] at (1,-1) {};
        \node [dot5 = {0}{0}{R}{above left}{R}] at (-2,0) {};
        \node [dot5 = {0}{0}{C}{above left}{C}] at (-1,0) {};
        \node [dot5 = {0}{0}{S}{above right}{S}] at (0,0) {};
        \node [dot5 = {0}{0}{B}{above}{B}] at (1,0) {};
        \node [dot5 = {0.05}{0}{N}{above left}{N}] at (1,1) {};
        \node [dot5 = {0}{0}{E}{above}{E}] at (-2, -1) {};
        \node [dot5 = {0}{0}{W}{below}{W}] at (-2.5,-2.5) {};
        \node [dot5 = {0}{0}{F}{below}{F}] at (-2,-2) {};
        \node [dot5 = {0}{0}{Q}{below}{Q}] at (-1,-2) {};
        \node [dot5 = {0}{0}{I}{above right}{I}] at (2,0) {};
        \draw [->] (O) -- (R);
        \draw [->] (O) -- (C);
        \draw [->] (O) -- (S);
        \draw [->] (R) -- (C);
        \draw [->] (C) -- (S);
        \draw [->] (S) -- (B);
        \draw [->] (B) -- (I);
        \draw [->] (W) -- (R);
        \draw [->] (W) -- (H);
        \draw [->] (E) -- (C);
        \draw [->] (M) -- (C);
        \draw [->] (C) -- (J);
        \draw [->] (O) -- (J);
        \draw [->] (J) -- (I);
        \draw [->] (D) -- (N);
        \draw [->] (M) -- (B);
        \draw [->] (Q) -- (F);
        \draw [->] (Q) -- (M);
        \draw [->] (M) -- (B);
        \draw [->] (F) -- (M);
        \draw [->] (F) -- (E);
        \draw [->] (H) -- (A);
        \draw [->] (A) -- (I);
        \draw [->] (C) -- (D);
        \draw [->] (D) -- (B);
        \draw [->] (D) -- (G);
        \draw [->] (E) -- (A);
        \draw [->] (N) -- (I);
        \draw [->] (B) -- (I);
        \draw [->] (G) -- (I);
        \draw [->] (D) -- (I);
        \draw [->] (Q) -- (E);
        \draw [->] (S) -- (G);
        \draw [->] (M) to [bend right = 60] (G);
        \draw [->] (M) to [bend left = 30] (N);
        \draw [->] (F) to [bend left = 60] (R);
      \end{tikzpicture}
      \caption{}
      \label{fig:prune1}
    \end{subfigure}
    \begin{subfigure}{0.4\textwidth}
      \centering
      \begin{tikzpicture}[scale=1.5]
        \node [dot5 = {0}{0}{O}{above}{O}] at (-1,1) {};
        \node [dot5 = {0}{0}{T}{right}{T}] at (-1.5,-1) {};
        \node [dot5 = {0}{0}{R}{above left}{R}] at (-2,0) {};
        \node [dot5 = {0}{0}{C}{above left}{C}] at (-1,0) {};
        \node [dot5 = {0}{0}{S}{above right}{S}] at (0,0) {};
        \node [dot5 = {0}{0}{F}{below}{F}] at (-2,-2) {};
        \node [dot5 = {0}{0}{Q}{below}{Q}] at (-1,-2) {};
        \draw [->] (O) -- (R);
        \draw [->] (O) -- (C);
        \draw [->] (O) -- (S);
        \draw [->] (R) -- (C);
        \draw [->] (C) -- (S);
        \draw [->] (T) -- (C);
        \draw [->] (Q) -- (F);
        \draw [->] (Q) -- (T);
        \draw [->] (F) -- (T);
        \draw [->] (F) to [bend left = 60] (R);
      \end{tikzpicture}
      \caption{}
      \label{fig:prune2}
    \end{subfigure}

    \begin{subfigure}{0.575\textwidth}
      \centering
      \begin{tikzpicture}[scale=1.5]
        \node [dot5 = {0}{0}{D}{below}{D}] at (0,-1) {};
        \node [dot5 = {0}{0}{O}{above}{O}] at (-1,1) {};
        \node [dot5 = {0}{0}{M}{above left}{M}] at (-1,-1) {};
        \node [dot5 = {0}{0}{R}{above left}{R}] at (-2,0) {};
        \node [dot5 = {0}{0}{C}{above left}{C}] at (-1,0) {};
        \node [dot5 = {0}{0}{S}{above right}{S}] at (0,0) {};
        \node [dot5 = {0}{0}{E}{above}{E}] at (-2, -1) {};
        \node [dot5 = {0}{0}{F}{below}{F}] at (-2,-2) {};
        \node [dot5 = {0}{0}{Q}{below}{Q}] at (-1,-2) {};
        \node [dot5 = {0}{0}{B}{above}{B}] at (1,0) {};
        \draw [->] (O) -- (R);
        \draw [->] (O) -- (C);
        \draw [->] (O) -- (S);
        \draw [->] (R) -- (C);
        \draw [->] (C) -- (S);
        \draw [->] (E) -- (C);
        \draw [->] (M) -- (C);
        \draw [->] (Q) -- (F);
        \draw [->] (Q) -- (M);
        \draw [->] (F) -- (M);
        \draw [->] (F) -- (E);
        \draw [->] (Q) -- (E);
        \draw [->] (D) -- (B);
        \draw [->] (C) -- (D);
        \draw [->] (M) -- (B);
        \draw [->] (S) -- (B);
        \draw [->] (F) to [bend left = 60] (R);
      \end{tikzpicture}
      \caption{}
      \label{fig:prune3}
    \end{subfigure}
    \begin{subfigure}{0.4\textwidth}
      \centering
      \begin{tikzpicture}[scale=1.5]
        \node [dot5 = {0}{0}{D}{below}{D}] at (0,-1) {};
        \node [dot5 = {0}{0}{O}{above}{O}] at (-1,1) {};
        \node [dot5 = {0}{0}{C}{above left}{C}] at (-1,0) {};
        \node [dot5 = {0}{0}{S}{above right}{S}] at (0,0) {};
        \node [dot5 = {0}{0}{G}{below right}{G}] at (1,-1) {};
        \node [dot5 = {0}{0}{M}{above left}{M}] at (-1,-1) {};
        \node [dot5 = {0}{0}{R}{above left}{R}] at (-2,0) {};
        \node [dot5 = {0}{0}{E}{above}{E}] at (-2, -1) {};
        \node [dot5 = {0}{0}{F}{below}{F}] at (-2,-2) {};
        \node [dot5 = {0}{0}{Q}{below}{Q}] at (-1,-2) {};
        \draw [->] (O) -- (C);
        \draw [->] (O) -- (R);
        \draw [->] (R) -- (C);
        \draw [->] (M) -- (C);
        \draw [->] (O) -- (S);
        \draw [->] (C) -- (S);
        \draw [->] (C) -- (D);
        \draw [->] (D) -- (G);
        \draw [->] (S) -- (G);
        \draw [->] (E) -- (C);
        \draw [->] (F) -- (E);
        \draw [->] (F) -- (M);
        \draw [->] (Q) -- (F);
        \draw [->] (Q) -- (M);
        \draw [->] (Q) -- (E);
        \draw [->] (M) to [bend right = 60] (G);
        \draw [->] (F) to [bend left = 60] (R);
      \end{tikzpicture}
      \caption{}
      \label{fig:prune4}
    \end{subfigure}
  \end{center}
  \caption{Causal graphs related to \citep{tobacco}: \subref{fig:prune1} the original graph, \subref{fig:prune2} the pruned graph for the causal effect $p(s \cond \doo(r))$, \subref{fig:prune3} the pruned graph for the causal effect $p(b \cond \doo(r))$, and \subref{fig:prune4} the pruned graph for the causal effect $p(g \cond \doo(c))$.}
  \label{fig:pruningapp}
\end{figure}

Assume we have two sources of data available. The first one provides the observational distribution $p(c, o, r, e, m)$ of cigarette consumption, tobacco control policies, tobacco price, education level, and maternal age. The second one provides the experimental distribution $p(c, s, g, d, b \cond\doo (o))$ on the effect of tobacco control policies on cigarette consumption, second-hand smoking in pregnancy, smoking during pregnancy, pre-term birth, and small-for-gestational-age infants. 

We consider three different causal effects: tobacco price on second-hand smoking in pregnancy, tobacco price on pre-term birth, and cigarette consumption on gestational age, denoted as $p(s\cond \doo(r))$, $p(b\cond \doo(r))$ and $p(g\cond \doo(c))$, respectively. In each setting, we can prune the graph differently. 

The pruned graph for the causal effect $p(s \cond \doo(r))$ is presented in Figure~\ref{fig:prune2}. Using Theorem~\ref{thm:Yancestors}, we pruned $J$, $N$, $B$, $I$, $D$, $G$, $H$ and $A$ which are non-ancestors of $S$. Variable $W$ was pruned according to Theorem~\ref{thm:Xancestors}, as it is connected to $S$ only through $R$ after pruning the non-ancestors. In addition, having pruned $A$, $G$, $J$ and $N$, the vertices $E$ and $M$ have the same parents and children, meaning that $\{E, M\}$ is a transit cluster that satisfies the conditions of Theorem~\ref{thm:singlelayerid}. This cluster is depicted as vertex $T$ in Figure~\ref{fig:prune2}. Now, using the pruned and clustered graph and the input distributions $p(c, o, r, t)$ and $p(c, s \cond \doo(o))$, we can obtain an identifying functional
\[
p(s \cond \doo(r))=\sum_{c,t,o}p(s \cond \doo(o),c)p(c \cond r,t,o)p(t,o),
\]
where the first term is derived from $p(c, s \cond \doo(o))$ and the other terms from $p(c, o, r, t)$.

Regarding the causal effect $p(b \cond \doo(r))$, we can perform the same pruning operations as in the case of the causal effect $p(s \cond \doo(r))$, with the exception that $D$ and $B$ cannot be pruned this time. This also means that we are not able to cluster $E$ and $M$ as they no longer share the same children. The pruned graph is presented in Figure~\ref{fig:prune3}. However, with the given data sources, the causal effect $p(b\cond \doo(r))$ is not identifiable. Relying on Theorems~\ref{thm:Yancestors} and \ref{thm:Xancestors}, we conclude that the causal effect is also not identifiable using a graph where none of the variables are removed.

Lastly, the pruned graph for the causal effect $p(g \cond \doo(c))$ is presented in Figure~\ref{fig:prune4}. Again, variables $J$, $N$, $B$, $I$, $A$ and $H$ can be pruned by Theorem~\ref{thm:Yancestors}. Now, $W$ is still connected to $G$ even after an intervention to $C$. However, after the first pruning, $W$ is connected to $G$ only through $R$, and can therefore be pruned using Theorem~\ref{thm:isolated}. Using the pruned graph and the restricted input distributions $p(c, o, r, e, m)$ and $p(c, s, g, d \cond \doo (o))$, we can identify the causal effect as
\[
p(g\cond \doo(c)) = \sum_{s,d,o}p(o)p(s,d \cond \doo(o),c)\sum_{c'}p(c' \cond \doo(o))p(g \cond \doo(o),c',s,d),
\]
where the first term is obtained from $p(c, o, r, e, m)$ and the other terms are derived from $p(c, s, g, d \cond \doo (o))$. All three settings are summarized in Table~\ref{tab:pruningapp}.

\begin{table}[!ht]
    \centering
    \begin{tabular}{ccccc}
        \toprule
        Causal effect & Fig. & Pruned vertices & Input distributions & ID \\
        \midrule
        $p(s \cond \doo(r))$  & \ref{fig:prune2} & $J, N, B, I, D, G, H, A, W$  & $p(c,o,r,t), p(c,s\cond \doo(o))$ & Yes \\
        $p(b \cond \doo(r))$  & \ref{fig:prune3} & $J, N, I, G, H, A, W$        & $p(c,o,r,e,m), p(c,s,d,b\cond \doo(o))$ & No \\
        $p(g \cond \doo(c))$  & \ref{fig:prune4} & $J, N, B, I, H, A, W$        & $p(c,o,r,e,m), p(c,s,d,g\cond \doo(o))$ & Yes \\
        \bottomrule
    \end{tabular}
    \caption{Summary of the results of Section~\ref{subsec:pruningdemo}. The table presents the causal effects, the graphs obtained by pruning, vertices that were pruned, the pruned input distributions, and whether the causal effect is identifiable.}
    \label{tab:pruningapp}
\end{table}

\subsection{Causal Model from an Atherosclerosis Study} \label{subsec:clusteringdemo}

We consider the ELSA-Brasil study \citep{elsabrazil} which investigated the effect of life-course socioeconomic position on carotid intima-media thickness, a marker of atherosclerosis \citep{atherosclerosis}. The variables of interest are presented in Table~\ref{tab:clusteringvariables}.

\begin{table}[h]
    \centering
    \begin{tabular}{lc}
        \toprule
        Variable & Symbol \\
        \midrule
        Baseline variables & $\+ B$ \\
        Life-course socioeconomic position & $L$ \\
        Job stress & $S$ \\
        Health-related behavior & $\+ H$ \\
        Metabolic, endocrine, and immune dysregulation & $\+ M$ \\
        Intima-media thickness & $Y$ \\
        \bottomrule
    \end{tabular}
    \caption{Variables in the causal graph of \citep{elsabrazil} and their corresponding symbols in Figure~\ref{fig:cluster1}.}
    \label{tab:clusteringvariables}
\end{table}

\citet{elsabrazil} suggested the causal graph of Figure~\ref{fig:cluster1} for the study setting. For this demonstration, we assume that the causal graph has been constructed in accordance with top-down causal modeling and consider variables $\+B$, $\+H$, and $\+M$ as transit clusters, with otherwise unspecified internal structures. Hence, we show the identification invariance for all relevant clusters in the examples. 

\begin{figure}[!ht]
  \begin{center}
    \begin{subfigure}{0.475\textwidth}
      \centering
      \begin{tikzpicture}[scale=2.5]
        \node [dot5 = {-0.05}{0}{L}{below left}{L}] at (0,0) {};
        \node [dot5 = {0}{0}{H}{below}{H}] at (0.75,0) {};
        \node [dot5 = {0}{0}{M}{below}{M}] at (1.25,0) {};
        \node [dot5 = {0.05}{0}{Y}{below right}{Y}] at (2,0) {};
        \node [dot5 = {0}{0}{S}{above right}{S}] at (1,0.618) {};
        \node [dot5 = {0}{0}{B}{above left}{B}] at (0.25,0.618) {};
        \draw [->] (L) -- (H);
        \draw [->] (H) -- (M);
        \draw [->] (M) -- (Y);
        \draw [->] (B) -- (S);
        \draw [->] (B) -- (L);
        \draw [->] (B) -- (M);
        \draw [->] (L) -- (S);
        \draw [->] (S) -- (H);
        \draw [->] (S) -- (M);
        \draw [->] (L) to [bend right=35] (Y); 
      \end{tikzpicture}
      \caption{}
      \label{fig:cluster1}
    \end{subfigure}
    \begin{subfigure}{0.475\textwidth}
      \centering
      \begin{tikzpicture}[scale=2.5]
        \node [dot5 = {-0.05}{0}{L}{below left}{L}] at (0,0) {};
        \node [dot5 = {0}{0}{H}{below}{H}] at (0.75,0) {};
        \node [dot5 = {0}{0}{M_1}{below}{M_1}] at (1.25,0) {};
        \node [dot5 = {0}{0}{M_2}{above}{M_2}] at (1.75,0) {};
        \node [dot5 = {0.05}{0}{Y}{below right}{Y}] at (2.25,0) {};
        \node [dot5 = {0}{0}{S}{above right}{S}] at (1,0.618) {};
        \node [dot5 = {0}{0}{B}{above left}{B}] at (0.25,0.618) {};
        \draw [->] (L) -- (H);
        \draw [->] (H) -- (M_1);
        \draw [->] (M_1) -- (M_2);
        \draw [->] (M_2) -- (Y);
        \draw [->] (B) -- (S);
        \draw [->] (B) -- (L);
        \draw [->] (B) -- (M_1);
        \draw [->] (L) -- (S);
        \draw [->] (S) -- (H);
        \draw [->] (S) -- (M_1);
        \draw [->] (L) to [bend right=30.5] (Y); 
      \end{tikzpicture}
      \caption{}
      \label{fig:cluster2}
    \end{subfigure}
  \end{center}
  \caption{Causal graphs related to \citep{elsabrazil}: \subref{fig:cluster1} the original graph and \subref{fig:cluster2} a graph where we assume that the cluster $\+M$ consists of two vertices $M_1$ and $M_2$.}
  \label{fig:clusterapp}
\end{figure}

\cite{elsabrazil} assumed that the data have been observed conditionally on the baseline variables, i.e., we have the input distribution of $p(y, l, \+h, s, \+m \cond \+b)$ available. With this data alone, the causal effect $p(y\cond\doo(l))$ is not identifiable. Given the input distribution, identification invariance holds when $\+M$, $\+H$ and $\+B$ are clustered in any order. This follows because in the input distribution $L$ blocks all backdoor paths between $\+M$ and its children, while $L$ and $S$ block all backdoor paths between $\+H$ and its children. In addition, the data source is not conditioned by a descendant of any cluster. Finally, transit cluster $\+B$ does not have any paths entering the cluster, satisfying the condition of line~\ref{line:noparents}. Therefore, the necessary conditions of \checkid{} apply to each cluster. Using Theorem~\ref{thm:invariance_nonid}, we can conclude that the causal effect is non-identifiable regardless of the internal structures of the clusters $\+B$, $\+H$ and $\+M$. 

Now, suppose we have additional data on the baseline variables in the Brazilian population. We can denote the combination of this information as a single input distribution $p(y, l, \+h, s, \+m, \+b)$. By Theorem~\ref{thm:invariance_id}, we obtain the identification formula $p(y \cond \doo(l))=\sum_{b}p(b)p(y \cond l,b)$ and by Theorem~\ref{thm:clusteringfunctional} we may write 
\[
p(y \cond \doo(l)) = \sum_{\+ b}p(\+ b)p(y \cond l,\+ b).
\]

Finally, we consider the effect of all health-related variables on intima-media thickness, i.e., $p(y\cond \doo(\+h))$, with input distributions of $p(\+b, \+m, s, y)$ and $p(\+b, \+h, \+m, s)$. Now, we cannot identify the causal effect in the clustered graph. In addition, Theorem~\ref{thm:invariance_nonid} does not apply to this setting because for the first data source there exists an open backdoor path between $\+M$ and $Y$ that is $Y \leftarrow L \rightarrow \+H \rightarrow \+M$. Therefore, it is possible that  $p(y\cond \doo(\+h))$ could be identifiable in a graph where the variables in $\+M$ are not clustered. To show that this indeed is the case, we consider the graph of Figure~\ref{fig:cluster2} where the transit cluster $\+M$ consists of two vertices $M_1$ and $M_2$. Now, using the same input distributions $p(\+b, m_1, m_2, s, y)$ and $p(\+b, \+h, m_1, m_2, s)$, where $\+M$ is not clustered, the causal effect can be identified as
\[
p(y\cond \doo(\+h)) = \sum_{s,m_2,\+b}\left(\sum_{m_1}p(s,\+b)p(m_1,m_2 \cond \+h,s,\+b)\right)\left(\sum_{m_1}p(m_1 \cond s,\+b)p(y \cond s,m_1,m_2,\+b)\right).
\]
This also demonstrates the necessity of line \ref{line:aparents} in \checkid{}. All four settings are summarized in Table~\ref{tab:clusterapp}.

\begin{table}[!ht]
    \centering
    \begin{tabular}{c c c c c c}
        \toprule
        Causal effect & Fig. & Input distributions & Clusters & Invariance & ID \\
        \midrule
        $p(y \cond \doo(l))$ & \ref{fig:cluster1} & $p(y, l, \+h, s, \+m \cond \+b)$ & $\+B$, $\+H$, $\+M$ & $\+B$, $\+H$, $\+M$ & No \\
        $p(y\cond \doo(l))$ & \ref{fig:cluster1} & $p(y, l, \+h, s, \+m, \+b)$ & $\+B$, $\+H$, $\+M$ & $\+B$, $\+H$, $\+M$ & Yes \\
        $p(y\cond \doo(\+h))$ & \ref{fig:cluster1} & $p(\+b,\+m,s,y), p(\+b,\+h,\+m,s)$ & $\+B$, $\+M$& $\+B$ & No \\
        $p(y\cond \doo(\+h))$ & \ref{fig:cluster2} & $p(\+b,m_1, m_2,s,y), p(\+b,\+h,m_1, m_2,s)$ & $\+B$ & $\+B$ & Yes \\
        \bottomrule
    \end{tabular}
    \caption{Summary of the results of Section~\ref{subsec:clusteringdemo}. The table presents the causal effects, the graphs used in the identification, the available input distributions, the clusters considered and for which clusters the identification invariance holds according to Theorems~\ref{thm:invariance_id} and \ref{thm:invariance_nonid}, and whether the causal effect is identifiable in the clustered graph.} 
    \label{tab:clusterapp}
\end{table}

\section{Conclusion} \label{sec:conclusion}

We have proposed pruning and clustering as preprocessing operations for causal data fusion and derived sufficient conditions for identification invariance under these operations. Preprocessing enables the presentation of causal graphs in a more concise manner and leads to more efficient application of identification methods and simplified identifying functionals. In addition, pruning and clustering can serve as valuable tools in the planning of data collection as it is not necessary to measure variables that can be pruned, and for transit clusters it suffices to measure only the emitters of the cluster.

The practical benefits of pruning and clustering were demonstrated in Section~\ref{sec:applications} with graphs that have been previously used in applied works. The benefits in identification are the most obvious when the variables in different data sources are partially overlapping because the current identification algorithms with polynomial complexity are not applicable in such settings. The simulation study presented in Section~\ref{sec:simulation} indicates that in moderately sized graphs, causal effect identification with clustering and pruning can be substantially faster, and even a small reduction in the graph size can decrease the running time considerably. The study also showed that when the graph could not be reduced, i.e., identification invariance could not be established for any graph reduction operation, the overhead for assessing the graphical conditions for identification invariance is insignificant compared to running the Do-search algorithm for the full graph.

We focused solely on marginal causal effects, but similar pruning and clustering results could likely be derived for conditional causal effects as well. Given a conditional causal effect of interest $p(\+ y \cond \doo(\+ x), \+ z)$, we can still apply the presented results to $p(\+ y,\+ z \cond \doo(\+ x))$ as sufficient criteria. In addition to transit clusters, there may also exist other types of clusters for which identification invariance can be established, but we emphasize that transit clusters can be found efficiently \citep{tikka2023clustering} and they have an intuitive interpretation. 

Apart from Theorem~\ref{thm:Yancestors} being the first pruning operation, we do not impose any particular order for performing the rest of the pruning operations or clustering. However, it is intuitive to think that the vertices that are completely irrelevant regarding the causal effect identification are removed first, after which we aim to find the transit clusters. Performing pruning first also removes the possibility that we end up pruning the clusters, which would render the clustering redundant. 

We emphasize that applying pruning and clustering is a decision made by the researcher and there might be reasons to avoid these operations even when they are technically possible. We also acknowledge that there may be  multiple ways to cluster a graph. In such situations, one approach could be to start with the largest cluster. Should the identification invariance be established, this approach would offer the largest benefits with respect to identification. Another approach could be to incorporate domain knowledge into clustering so that the cluster has a straightforward interpretation.

The theoretical results were derived without concepts such as c-components, hedges, or the latent projection which are essential when operating with a single observational data source but are not well-defined for the general causal effect identification problem where the variables may be observed in one data source but unobserved in another. Instead, to prove identification invariance between two causal graphs, we relied directly on the definition of identifiability and constructed the models that demonstrate non-identifiability in one graph starting from models with the same property that exist for the other graph by assumption. \citet{kivva2022revisiting} employed a similar proof strategy for g-identifiability by establishing a connection between the original graph and a modified graph.

Identifying functionals obtainable via do-calculus in the pruned or clustered causal graphs are directly applicable to the original graphs as demonstrated by Theorem~\ref{thm:pruningfunctional} and Theorem~\ref{thm:clusteringfunctional}. We note that the limitation to functionals obtainable specifically via do-calculus is merely technical because in practice no other methods to identify causal effects are known for the general non-parametric setting. Nonetheless, it is of theoretical interest whether do-calculus is complete for the general causal effect identifiability problem.

\acks{The authors wish to thank Jouni Helske and the anonymous referees for their comments that helped to improve the paper. The authors wish to acknowledge CSC -- IT Center for Science, Finland, for computational resources. This work was supported by the Finnish Doctoral Program Network in Artificial Intelligence (AI-DOC), Decision VN/3137/2024-OKM-6 and by the Research Council of Finland under grant number 368935.} 

\nosectionappendix

\begin{toappendix}
\section{Examples on Identification Non-invariance} \label{apx:counterexamples}

We show that the d-separation conditions of lines~\ref{line:aparents} and \ref{line:cparents} of \checkid{} cannot be relaxed. If either condition is violated, then it is possible that a causal effect is identifiable in the original graph but non-identifiable in the clustered graph.

First, we consider the case where $T \in \+ A'_i$, but the condition of line~\ref{line:aparents} does not hold, i.e., $(\+D \not\independent T \cond \+B_i, \+C_i, \+A_i\setminus(\+D\cup T))_{\sG'[\overline{\+B_i}, \underline{T}]}$, where $\+D= \De_{\sG'}(T) \cap \+ A'_i$. Consider the graph of Figure~\ref{fig:counter1a} and suppose our input distributions are $p(x, z_1, z_2)$ and $p(y, z_1, z_2)$. The causal effect $p(y \cond \doo(x))$ is identifiable and
\begin{equation} \label{eq:counter1}
  p(y \cond \doo(x)) = \sum_{z_2}p(z_2 \cond x)\sum_{z_1}p(z_1)p(y \cond z_1, z_2).
\end{equation}
However, if we cluster $\{Z_1, Z_2\}$ as $Z$ and consider the corresponding causal graph depicted in Figure~\ref{fig:counter1b}, the causal effect is no longer identifiable from the clustered input distributions $p(x, z)$ and $p(y, z)$. We can see that condition of line~\ref{line:aparents} is not satisfied for the second input distribution $p(y, z_1, z_2)$ because the input distribution does not contain a sufficient set to close the backdoor path $Y\leftarrow W \rightarrow X \rightarrow Z$. In contrast, if the second input contains $X$ or $W$, such as $p(y, x, z_1, z_2)$ or $p(y, z_1, z_2 \cond w)$, then the aforementioned backdoor path can be closed, and the causal effect is identifiable both before and after clustering.

\begin{figure}[!ht]
  \begin{center}
    \begin{subfigure}{0.475\textwidth}
      \centering
      \begin{tikzpicture}[xscale=1.5,yscale=2]
        \node [dot = {0}{0}{W}{above}] at (1.5,0.67) {};
        \node [dot = {0}{0}{X}{below}] at (0,0) {};
        \node [dot = {0}{0}{Z_1}{below}] at (1,0) {};
        \node [dot = {0}{0}{Z_2}{below}] at (2,0) {};
        \node [dot = {0}{0}{Y}{below}] at (3,0) {};
        \draw [->] (W) -- (X);
        \draw [->] (W) -- (Y);
        \draw [->] (X) -- (Z_1);
        \draw [->] (Z_1) -- (Z_2);
        \draw [->] (Z_2) -- (Y);
      \end{tikzpicture}
      \caption{}
      \label{fig:counter1a}
    \end{subfigure}
    \begin{subfigure}{0.475\textwidth}
      \centering
      \begin{tikzpicture}[scale=2]
        \node [dot = {0}{0}{W}{above}] at (1,0.67) {};
        \node [dot = {0}{0}{X}{below}] at (0,0) {};
        \node [dot = {0}{0}{Z}{below}] at (1,0) {};
        \node [dot = {0}{0}{Y}{below}] at (2,0) {};
        \draw [->] (W) -- (X);
        \draw [->] (W) -- (Y);
        \draw [->] (X) -- (Z);
        \draw [->] (Z) -- (Y);
      \end{tikzpicture}
      \caption{}
      \label{fig:counter1b}
    \end{subfigure}
  \end{center}
  \caption{Causal graphs demonstrating the necessity of lines~\ref{line:aparents} and \ref{line:cparents} in \checkid{}.}
  \label{fig:counter1}
\end{figure}

Next, we consider the case where $T \in \+ C'_i$, but condition of line~\ref{line:cparents} is violated, i.e., $(\+A'_i \not\independent T \cond \+B_i, \+C_i \setminus T)_{\sG'[\overline{\+B'_i}, \underline{T}]}$. We continue with the graphs of Figure~\ref{fig:counter1} and the transit cluster $\{Z_1,Z_2\}$. Suppose now that our input distributions are instead $p(x,z_1,z_2)$ and $p(y \cond z_1,z_2)$. The causal effect $p(y \cond \doo(x))$ is again identifiable in the graph of Figure~\ref{fig:counter1a} with the same formula as in Equation~\ref{eq:counter1}. Again, the causal effect is no longer identifiable if we cluster $\{Z_1,Z_2\}$ as $Z$ and consider the clustered input distributions $p(x,z)$ and $p(y \cond z)$. Similarly as before, line~\ref{line:cparents} is not satisfied for the second input distribution because the backdoor path between $Z$ and $Y$ is open. If the second input is replaced by $p(y \cond z_1,z_2,x)$, $p(y\cond \doo(x),z_1,z_2)$, $p(y \cond z_1,z_2,w)$, or $p(y\cond \doo(w),z_1,z_2)$, which would satisfy the condition, the causal effect is again identifiable in both graphs.
\end{toappendix}

\clearpage

\begin{toappendix}
\section{Further Details and Results on the Simulations} \label{apx:algorithm}

\subsection{Generation of a Simulation Instance}

In Algorithm~\ref{alg:times}, we present the process of generating one simulation instance when clustering and pruning are applied together. As a rough outline of the algorithm, we begin in lines~\ref{simulate:init}--\ref{simulate:nonancestors} by creating the clustered graph $\sG'$ which contains at least one non-ancestor of the response $Y$. We then select one vertex as the cluster vertex and construct the original graph $\sG$ by replacing the cluster vertex in $\sG'$ by the newly formed cluster set in lines~\ref{simulate:randomZ}--\ref{simulate:finalizeT}. Lines~\ref{simulate:ninputs}--\ref{simulate:finalizeinputs} generate two or three input distributions, while ensuring that the inputs are compatible by Definition~\ref{def:clusteredinput} and that no counterintuitive inputs are formed (such as inputs with no observed variables). Finally, lines~\ref{simulate:dosearch_og}--\ref{simulate:settingD} execute the direct and reduction strategies, as was explained in Section~\ref{sec:simulation}.

\begin{algorithm}[!t]
    \small
  \begin{algorithmic}[1]
\Function{\simulate{}}{}
\State Let $\+V'=\{X, Y, Z_1,\dots Z_n\}$, where $n \sim U(\{3,4,5,6\})$. \label{simulate:init}
\State Let $\pi$ be a topological order for $\+V'$ such that $X$ is not first and $Y$ is the last in $\pi$. \label{simulate:topological}
\State For all $V', W'\in \+V'$ such that $V' \prec W'$ in $\pi$, generate a directed edge from $V'$ to $W'$ with probability 0.35 if $V' \prec X$ in $\pi$, otherwise with probability 0.5.
\State If $V' \in \+V'$ is not an ancestor of $Y$, generate an edge to $W'$ such that $V' \prec W'$ in $\pi$ and $W'\in Y \cup \An(Y)$. \label{simulate:path_to_y}
\State Let $k \sim U(\{1, 2\})$. If $k=1$, let $\+V'=\+V'\cup\{W_1\}$, else $\+V'=\+V'\cup\{W_1, W_2\}$, and let there be directed edges from $Y$ to $W_1$ and $W_2$. \label{simulate:nonancestors} 
\State Let $\sG'[\+V']$ be the clustered causal graph that is the result of the lines~\ref{simulate:init}--\ref{simulate:nonancestors}. 
\State Select a random vertex $Z \in \{Z_1, Z_2, \dots Z_n\}$ as the cluster vertex. \label{simulate:randomZ}
\State Let $\+R=\emptyset$. If $\Pa(Z)\neq\emptyset$, let $\+R=\{R_1\}$ or $\+R=\{R_1, R_2\}$ with probabilities 0.5.
\State Let $\+E=\emptyset$. If $\Ch(Z)\neq\emptyset$, let $\+E=\{E_1\}$ or  $\+E=\{E_1, E_2\}$ with probabilities 0.5.
\State Let $\+S=\emptyset$ with probability 0.5, otherwise $\+S=\{S\}$. 
\State Let $\+T= \+R \cup \+S \cup \+E$. If $|\+T| < 2$, go to line 8.
\State Let $\phi$ be a topological order of $\+T$ such that $\+R \prec \+S \prec \+E$.
\State For all $T_1, T_2\in \+T$ such that $T_1 \prec T_2$ in $\phi$, generate a directed edge from $T_1$ to $T_2$ with probability 0.5. Repeat this step until all following conditions are met: 1) Each $R \in \+R$ is an ancestor of some $E \in \+E$, 2) Each $E \in \+E$ is a descendant of some $R \in \+R$, 3) If $\+S\neq\emptyset$, then $\Pa(S)\neq\emptyset$ and $\Ch(S) \neq \emptyset$. After this step $\+T$ is a transit cluster by Definition~\ref{def:transitcluster}. \label{simulate:finalizeT}
\State Let $\+V = (\+V' \cup \+T) \setminus Z$. Expand $\sG'[\+V']$ to $\sG[\+V]$ where $Z$ is replaced by $\+T$.
\State Choose the number of input distributions in $\sI'$ randomly as 2 or 3 with probabilities 0.5. \label{simulate:ninputs}
\State Each member $p(\+a_i\cond \doo(\+b_i), \+c_i)$ of $\sI'$ is generated as follows: every member of $\+V'$ is randomly assigned to the sets $\+ A_i$, $\+ B_i$, $\+ C_i$ and $\+ D_i$ with probabilities 0.35, 0.125, 0.125, and 0.40, respectively. Repeat this step until all following conditions are met: 1) $\+A_i\neq\emptyset$ for all $i$, 2) if $Y\in\+A_i$ then $X \notin \+B_i$ for all $i$, 3) $Y\in\+A_i$ for some $i$, 4) $X, Z\in \+A_i\cup\+B_i\cup\+C_i$ for some $i$. 
\State To generate original inputs, let $\sI=\sI'$. For all $\sI_i \in \sI$ such that $Z\in \+A_i \cup \+B_i \cup \+C_i$, $\textbf{do}$
    \State \quad \quad Let $\+Q=\+E$. For all $T\in\+T\setminus\+E$, append $\+Q$ by $T$ with probability 0.5.  \label{simulate:constructQ}
    \State \quad \quad If $\+Q = \emptyset$, go to line~\ref{simulate:constructQ}. Otherwise, replace $Z$ by $\+Q$ in $\sI_i$. \label{simulate:finalizeinputs}
\State Use Do-search to determine if $p(y\cond\doo(x))$ is identifiable from $\sI$ in $\sG$. Let $t_1$ be the running time for this operation. If more than 15 minutes have elapsed, terminate the operation. \label{simulate:dosearch_og}
\State Let $\sI_P=I$ and $\sG^P=\sG$. \label{simulate:initpruning} 
\State Check if Theorems~\ref{thm:Yancestors}--\ref{thm:isolated} can be applied. Let $\+P_7$, $\+P_8$ and $\+P_9$ be the corresponding pruned sets. If a pruned set intersects with $\+T$, replace it with the empty set. Now it holds $(\+P_7\cup\+P_8\cup\+P_9)\cap \+T = \emptyset$. Let $\+P=\+P_7\cup \+P_8 \cup \+P_9$ and let $\sI^P=\sI^P[\+V\setminus\+P]$, $\sG^P=\sG^P[\+V \setminus \+P]$ and $\sI'=\sI'[\+V'\setminus\+P]$, $\sG'=\sG'[\+V' \setminus \+P]$. Let $t_2$ be the running time for the pruning operation. \label{simulate:pruning}
\State Use \textsc{FindTrClust} \citep{tikka2021dosearch}, to find the transit clusters of $\sG$. Let $t_3$ be the running time for this operation. \label{simulate:findtrclust}
\State Use Do-search to determine if $p(y\cond\doo(x))$ is identifiable from $\sI'$ in  $\sG'$. Let $t_4$ be the running time for this operation.
\State If $p(y\cond\doo(x))$ is identifiable from $\sI'$ in $\sG'$\textbf{ return} $(t_1, t_2 + t_3 + t_4)$. 
\State Run \checkid{}($\sG'$, $\sI'$, $Z$). Let $t_5$ be the running time for this operation. If \mbox{\checkid{}} returns \textbf{TRUE}, \textbf{return}  $(t_1, t_2 + t_3 + t_4 +t_5)$. \label{simulate:verifyinputs}
\State If $\sI^P\neq\sI$, use Do-search to determine if $p(y\cond\doo(x))$ is identifiable from $\sI^P$ in $\sG^P$. Let $t_6$ be the running time for this operation and \textbf{return} $(t_1, t_2+t_3+t_4+t_5+t_6)$ \label{simulate:dosearch_pruned} 
\State Else \textbf{return} $(t_1, t_1 + t_2 + t_3 + t_4 + t_5 + t_6)$. \label{simulate:settingD}
\EndFunction
  \end{algorithmic}
  \caption{Generate a single simulation instance with a random unreduced and reduced causal graph and random input distributions. The function returns the running times for causal effect identification with direct and reduction strategies.}
  \label{alg:times}
\end{algorithm}

\subsection{Benefits of Pruning When Clustering Is Not Helpful}

In the simulation study of Section~\ref{sec:simulation}, we generated instances of four settings. However, in Table~\ref{tab:simresults} and Figure~\ref{fig:scatter} we concentrated on the most essential settings A, B and C. In setting D, the graph is pruned and clustered but it turns out that clustering is not helpful for identification. Nevertheless, the graph was reduced by pruning which decreases the running time of Do-search. The simulation results for setting D are summarized in Table~\ref{tab:settingD}.

\begin{table}[ht]
\centering
\begin{tabular}{c c r l r l r}
\toprule
Graph size & Setting & \multicolumn{2}{c}{Median difference (IQR)} & \multicolumn{2}{c}{Median ratio (IQR)} & $N$ \\
\midrule
7 & D & $-0.03$ & $(-0.07,\ 0.02)$ & $0.74$ & $(0.40,\ 1.18)$ & 397 \\
8 & D & $0.30$ & $( 0.09,\ 0.69)$ & $2.66$ & $(1.57,\ 4.20)$ & 1 692 \\
9 & D & $2.47$ & $( 1.12,\ 4.88)$ & $6.16$ & $(4.07,\ 11.42)$ & 4 045 \\
10 & D & $15.34$ & $( 6.97,\ 31.79)$ & $8.42$ & $(5.93,\ 32.37)$ & 6 277 \\
11 & D & $85.11$ & $( 39.36,\ 179.14)$ & $10.08$ & $(6.71,\ 46.17)$ & 7 219 \\
12 & D & $442.94$ & $( 198.37,\ 752.05)$ & $15.14$ & $(5.51,\ 45.22)$ & 5 578 \\
\bottomrule
\end{tabular}
\caption{Benefits of pruning when clustering is not helpful (setting D). Median differences (in seconds) of running times between the direct and reduction strategies, and median ratios (the running time for the direct strategy divided by the running time of the reduction strategy) are reported. The interquartile ranges (IQR) are included in parentheses. The number of instances for each graph size is denoted by $N$.}
\label{tab:settingD}
\end{table}

We can note that, as was the case with settings A and B, setting D also provides clear benefits as the graph size increases. Compared to settings A and B, it appears that setting D was the least beneficial. However, this observation can be explained by the design of the study. In setting D, Do-search is applied twice: first for the clustered and pruned graph, and then for the pruned-only graph. In addition, pruning is likely to remove fewer vertices from the graph, due to limiting pruning to vertices that are neither treatment nor response variables, and that do not belong to the cluster. The benefits of pruning alone are illustrated more prominently in Table~\ref{tab:pruning}.

\subsection{Simulations for Clustering and Pruning Separately}

In addition to performing pruning and clustering together in the same instance, we also study their impacts separately. In such cases, some streamlining can be made to the procedures described earlier. 
In the pruning strategy, the following changes are made to the reduction strategy of Section~\ref{sec:simulation}:

\begin{itemize}
    \item The steps~2--5 in the reduction strategy are dismissed, as they are related to clustering. This leaves us with two settings, E: apply pruning to the original graph and use Do-search in the pruned graph, and F: pruning operations are not identification invariant and Do-search is used for the original graph.
    \item On line~\ref{simulate:pruning}, $\+T=\emptyset$, i.e., we can also allow the pruning of the members in the generated cluster. 
    \item Lines~\ref{simulate:findtrclust}--\ref{simulate:verifyinputs} in \simulate{} are not executed, as they concern steps~2--5 of the reduction strategy. In this case $t_3 = t_4 = t_5 = 0$.
\end{itemize}

\noindent Correspondingly, the following changes are made in the clustering strategy: 

\begin{itemize}
    \item The steps~1 and 6 in the reduction strategy are dismissed, as they are related to pruning. In this case, we have three settings, G: the causal effect is identifiable in the clustered graph, H: the causal effect is non-identifiable, but the clustering operation is identification invariant and I: the clustering is not identification invariant, Do-search is used for the original graph.
    \item In \simulate{}, line~\ref{simulate:nonancestors} that generates non-ancestors, and lines~\ref{simulate:initpruning}, \ref{simulate:pruning} and \ref{simulate:dosearch_pruned} that assess the identification invariance of the pruned graph, are not executed. In this case $t_2 = t_6=0$.
\end{itemize}

The results regarding pruning only (N = 120~000) are summarized in Table~\ref{tab:pruning} and Figures~\ref{fig:scatterpr} and \ref{fig:box_pr}. The clustering-only results (N = 105~772) are summarized in Table~\ref{tab:clustering} and Figures~\ref{fig:scattercl} and \ref{fig:box_cl}. In brief, the results are in line with the simulation conducted in Section~\ref{sec:simulation}, demonstrating the benefits of both pruning and clustering.

\clearpage

\begin{table}[ht]
\centering
\begin{tabular}{c c r l r l r}
\toprule
Graph size & Setting & \multicolumn{2}{c}{Median difference (IQR)} & \multicolumn{2}{c}{Median ratio (IQR)} & $N$ \\
\midrule
7 & E & $0.05$ & $( 0.01,\ 0.11)$ & $2.68$ & $(1.44,\ 4.55)$ & 1 875 \\
7 & F & $-0.02$ & $(-0.03,\ -0.01)$ & $0.82$ & $(0.70,\ 0.89)$ & 254 \\
8 & E & $0.40$ & $( 0.15,\ 0.83)$ & $7.45$ & $(3.98,\ 18.14)$ & 8 336 \\
8 & F & $-0.02$ & $(-0.03,\ -0.01)$ & $0.96$ & $(0.91,\ 0.98)$ & 1 477 \\
9 & E & $2.51$ & $( 1.03,\ 5.33)$ & $18.65$ & $(6.82,\ 58.07)$ & 16 711 \\
9 & F & $-0.02$ & $(-0.03,\ -0.01)$ & $0.99$ & $(0.98,\ 1.00)$ & 3 629 \\
10 & E & $15.25$ & $( 6.19,\ 33.72)$ & $32.29$ & $(7.76,\ 116.89)$ & 21 619 \\
10 & F & $-0.02$ & $(-0.04,\ -0.02)$ & $1.00$ & $(1.00,\ 1.00)$ & 5 498 \\
11 & E & $87.43$ & $( 35.36,\ 196.99)$ & $34.72$ & $(7.59,\ 176.70)$ & 20 923 \\
11 & F & $-0.02$ & $(-0.04,\ -0.02)$ & $1.00$ & $(1.00,\ 1.00)$ & 6 099 \\
12 & E & $481.80$ & $( 194.75,\ 860.78)$ & $39.77$ & $(8.78,\ 265.42)$ & 14 267 \\
12 & F & $-0.03$ & $(-0.05,\ -0.02)$ & $1.00$ & $(1.00,\ 1.00)$ & 5 084 \\
\bottomrule
\end{tabular}
\caption{Comparison of the direct strategy and the pruning strategy. The settings are E: graph was reduced by pruning, F: pruning was not possible and the original graph was used. Median differences (in seconds) of running times between the direct and reduction strategies, and median ratios (the running time for the direct strategy divided by the running time of the reduction strategy) are reported. The interquartile ranges (IQR) are included in parentheses.  The column $N$ tells the number of instances for each setting and original graph size.}
\label{tab:pruning}
\end{table}

\begin{figure}[!ht]
  \centering
  \begin{subfigure}{0.40\textwidth}
    \centering
    \includegraphics[width=\linewidth]{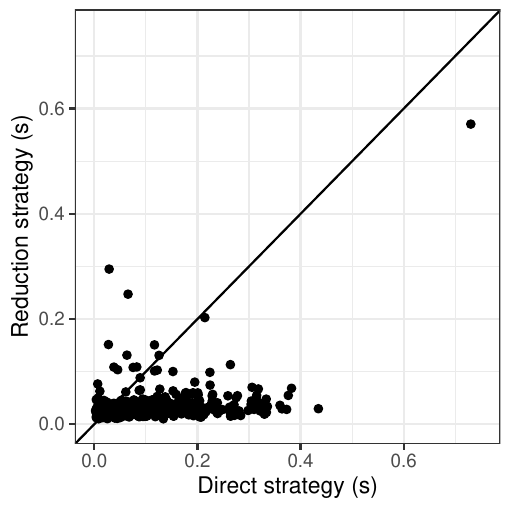}
    \caption{}
    \label{fig:scatter6pr}
  \end{subfigure}
  \begin{subfigure}{0.40\textwidth}
    \centering
    \includegraphics[width=\linewidth]{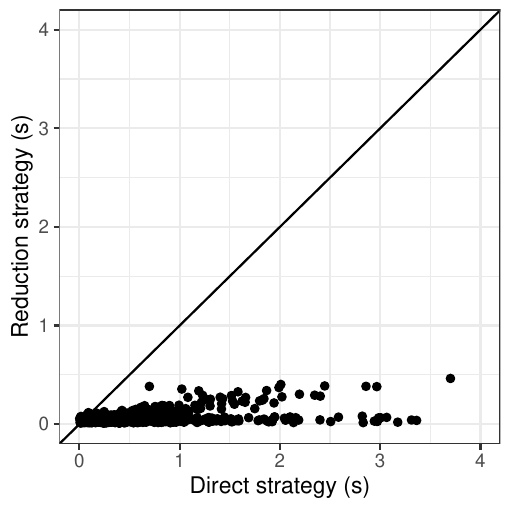}
    \caption{}
    \label{fig:scatter8pr}
  \end{subfigure}
  \begin{subfigure}{0.4\textwidth}
    \centering
    \includegraphics[width=\linewidth]{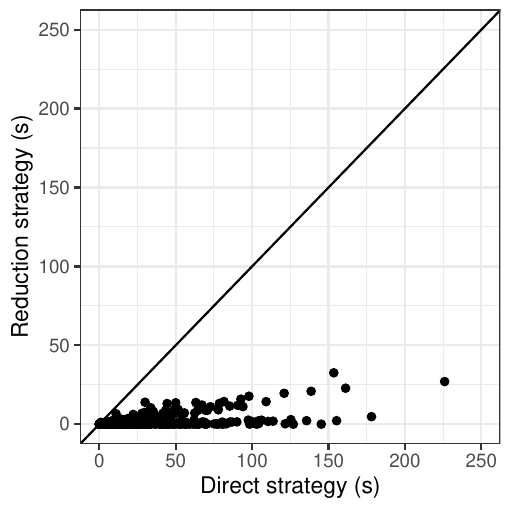}
    \caption{}
    \label{fig:scatter10pr}
  \end{subfigure}
  \begin{subfigure}{0.4\textwidth}
    \centering
    \includegraphics[width=\linewidth]{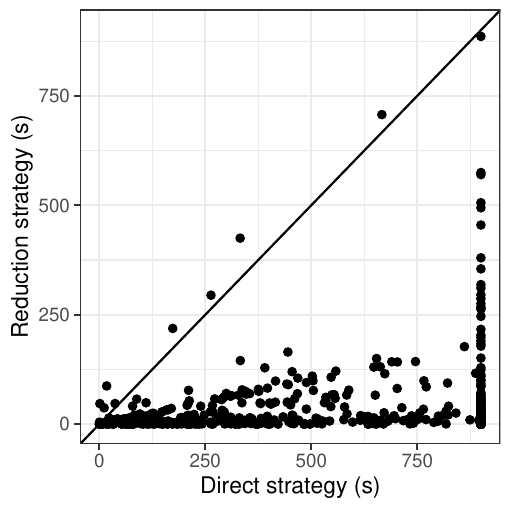}
    \caption{}
    \label{fig:scatter12pr}
  \end{subfigure}
  \caption{Comparison of the direct and the pruning strategies when pruning is possible. The scatter plots show the running times for the direct strategy and the pruning strategy for setting E. The panels depict different original graph sizes: 7 \subref{fig:scatter6pr}, 8 \subref{fig:scatter8pr}, 10 \subref{fig:scatter10pr} and 12 \subref{fig:scatter12pr}. A random sample of 500 instances is used for each panel. In Panel (d), the heap of points at 900 seconds is due to the time limit of 15 minutes.}
  \label{fig:scatterpr}
\end{figure}

\begin{figure}[!t]
    \centering
    \includegraphics[width=0.95\linewidth]{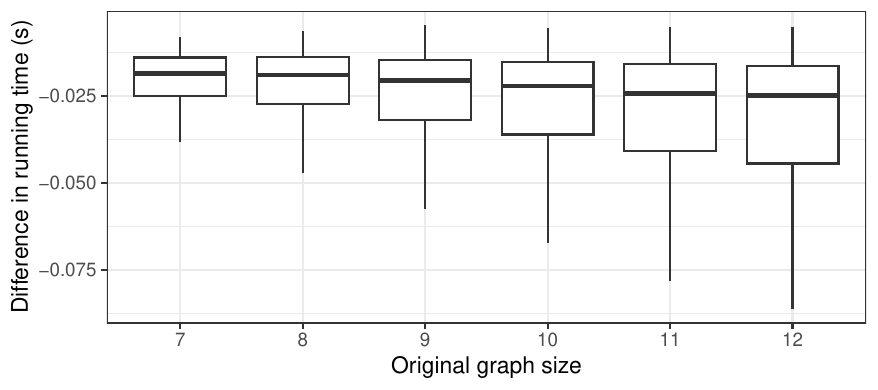}
    \caption{Overhead of the pruning strategy when pruning is not possible. The box plots show the differences in running times between the direct strategy and the pruning strategy for setting F by the size of the original graph. Negative differences indicate that more time was spent with the pruning strategy.}
    \label{fig:box_pr}
\end{figure}

\begin{table}[ht]
\centering
\begin{tabular}{c c r l r l r}
\toprule
Graph size & Setting & \multicolumn{2}{c}{Median difference (IQR)} & \multicolumn{2}{c}{Median ratio (IQR)} & $N$ \\
\midrule
6 & G & $-0.06$ & $(-0.07,\ -0.05)$ & $0.17$ & $(0.09,\ 0.30)$ & 1 872 \\
6 & H & $-0.05$ & $(-0.07,\ -0.04)$ & $0.23$ & $(0.14,\ 0.37)$ & 1 585 \\
6 & I & $-0.09$ & $(-0.11,\ -0.08)$ & $0.13$ & $(0.09,\ 0.20)$ & 944 \\
7 & G & $-0.04$ & $(-0.08,\ 0.02)$ & $0.57$ & $(0.22,\ 1.17)$ & 6 156 \\
7 & H & $-0.01$ & $(-0.05,\ 0.07)$ & $0.90$ & $(0.47,\ 1.66)$ & 5 792 \\
7 & I & $-0.13$ & $(-0.15,\ -0.11)$ & $0.37$ & $(0.24,\ 0.52)$ & 3 718 \\
8 & G & $0.16$ & $(-0.04,\ 0.57)$ & $2.08$ & $(0.70,\ 4.70)$ & 8 726 \\
8 & H & $0.45$ & $( 0.14,\ 1.03)$ & $3.88$ & $(1.95,\ 7.78)$ & 9 954 \\
8 & I & $-0.17$ & $(-0.21,\ -0.14)$ & $0.71$ & $(0.56,\ 0.83)$ & 7 635 \\
9 & G & $1.48$ & $( 0.32,\ 3.85)$ & $6.95$ & $(2.62,\ 17.08)$ & 7 691 \\
9 & H & $3.36$ & $( 1.39,\ 7.27)$ & $13.37$ & $(6.08,\ 30.41)$ & 11 191 \\
9 & I & $-0.25$ & $(-0.36,\ -0.20)$ & $0.89$ & $(0.81,\ 0.95)$ & 10 735 \\
10 & G & $9.23$ & $( 2.59,\ 23.30)$ & $21.92$ & $(9.22,\ 52.05)$ & 5 059 \\
10 & H & $21.55$ & $( 9.01,\ 45.88)$ & $44.62$ & $(21.05,\ 90.50)$ & 9 877 \\
10 & I & $-0.40$ & $(-0.71,\ -0.28)$ & $0.97$ & $(0.94,\ 0.98)$ & 10 917 \\
11 & G & $49.07$ & $( 16.28,\ 129.81)$ & $80.32$ & $(33.69,\ 159.16)$ & 2 347 \\
11 & H & $122.58$ & $( 49.48,\ 289.03)$ & $163.69$ & $(88.25,\ 361.43)$ & 5 312 \\
11 & I & $-0.60$ & $(-0.97,\ -0.39)$ & $0.99$ & $(0.99,\ 1.00)$ & 6 661 \\
12 & G & $243.13$ & $( 80.53,\ 590.55)$ & $285.31$ & $(129.83,\ 481.84)$ & 549 \\
12 & H & $681.84$ & $( 262.97,\ 898.71)$ & $528.05$ & $(351.77,\ 829.21)$ & 1 429 \\
12 & I & $-0.80$ & $(-1.13,\ -0.55)$ & $1.00$ & $(1.00,\ 1.00)$ & 1 848 \\
\bottomrule
\end{tabular}
\caption{Comparison of the direct strategy and the clustering strategy. The settings are G: identifiable in the clustered graph, H: non-identifiable in the clustered graph but identification invariant, I: non-identifiable in the clustered graph and not identification invariant. Median differences (in seconds) of running times between the direct and reduction strategies, and median ratios (the running time for the direct strategy divided by the running time of the reduction strategy) are reported. The interquartile ranges (IQR) are included in parentheses. The column $N$ tells the number of instances for each setting and original graph size.}
\label{tab:clustering}
\end{table}

\begin{figure}[!ht]
  \centering
  \begin{subfigure}{0.40\textwidth}
    \centering
    \includegraphics[width=\linewidth]{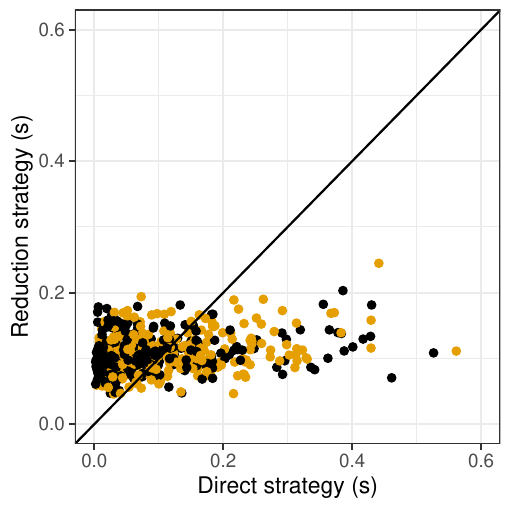}
    \caption{}
    \label{fig:scatter6cl}
  \end{subfigure}
  \begin{subfigure}{0.40\textwidth}
    \centering
    \includegraphics[width=\linewidth]{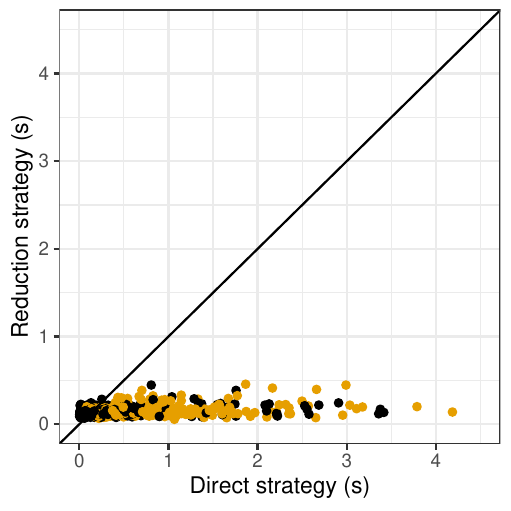}
    \caption{}
    \label{fig:scatter8cl}
  \end{subfigure}
  \begin{subfigure}{0.4\textwidth}
    \centering
    \includegraphics[width=\linewidth]{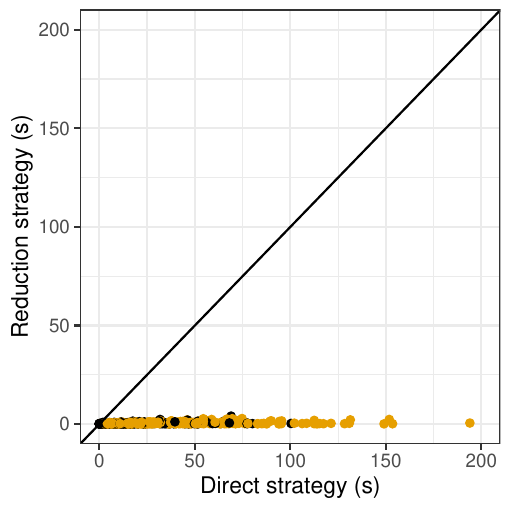}
    \caption{}
    \label{fig:scatter10cl}
  \end{subfigure}
  \begin{subfigure}{0.4\textwidth}
    \centering
    \includegraphics[width=\linewidth]{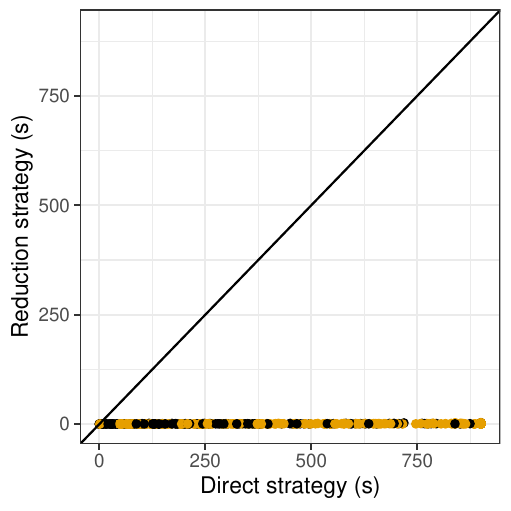}
    \caption{}
    \label{fig:scatter12cl}
  \end{subfigure}
  \caption{Comparison of the direct and the clustering strategies when clustering is helpful. The scatter plots show the running times for the direct strategy and the clustering strategy for settings G and H. The panels depict different original graph sizes: 7 \subref{fig:scatter6cl}, 8 \subref{fig:scatter8cl}, 10 \subref{fig:scatter10cl} and 12 \subref{fig:scatter12cl}. A random sample of 500 instances is used for each panel. The points are colored by the setting (black for setting G, and orange for setting H).}
  \label{fig:scattercl}
\end{figure}

\begin{figure}
    \centering
    \includegraphics[width=0.95\linewidth]{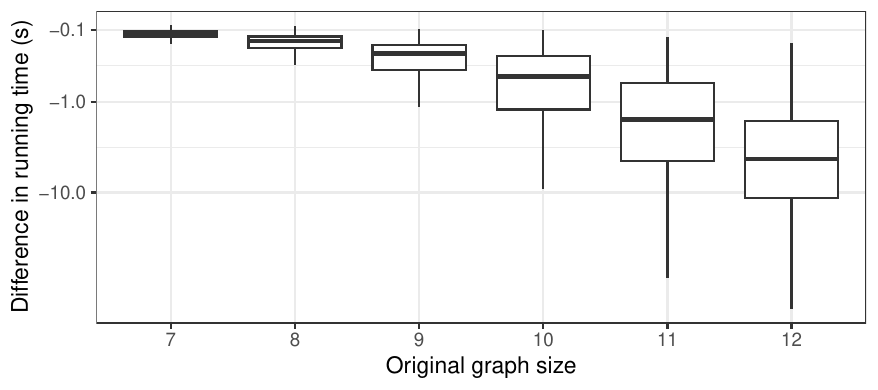}
    \caption{Overhead of the clustering strategy when clustering is not helpful for identification. The box plots show the differences in running times between the direct strategy and the clustering strategy for setting I. Negative differences indicate that more time was spent with the clustering strategy. The vertical axis uses logarithmic scaling.}
    \label{fig:box_cl}
\end{figure}

\end{toappendix}

\begin{toappendix}
\clearpage
\bibliography{references}
\label{applastpage} 
\end{toappendix}

\end{document}